\documentclass[sigconf]{acmart}

\usepackage{dirtytalk}
\usepackage{enumitem}
\usepackage{flushend}
\usepackage{changepage}

\AtBeginDocument{%
 \providecommand\BibTeX{{%
   \normalfont B\kern-0.5em{\scshape i\kern-0.25em b}\kern-0.8em\TeX}}}

\acmConference[Intelligence Prime (2026 Release)]{Intelligence Primer}{Seattle,Washington State}

\begin{document}

\title{Intelligence Primer: Natural and Artificial\\\normalsize
---+---\\
6th Edition (\today)}


\author{Andrew N. Sloss}
\affiliation{%
  \city{Seattle}
  \state{Washington State}
  \country{USA}
}
\email{andrew@sloss.net}

\author{Karl F. Fezer}
\affiliation{%
  \city{Seattle}
  \state{Washington State}
  \country{USA}
}
\email{kfezer@gmail.com}


\begin{abstract}

\vspace{1.0em}


We first published the \textit{Intelligence Primer} in 2020 to demystify a word that had become so ubiquitous and overused. We were struck by how little we truly understood \textit{what intelligence is?}, so we decided to explore the ideas and principles that surround the topic. This \textit{primer} represents a recording of our journey. For context, at the time, \textit{Large Language Models} were still in their infancy\textemdash landmark systems like GPT-3 were only just emerging. \textit{Convolutional Neural Networks} had already proven themselves as important tools for object recognition. Since 2020, we have continued to update the paper on an annual basis with new editions. For the authors, it is an exciting exploration of an important topic that is challenging and changing in real time as research uncovers new attributes. \par 

Intelligence interweaves our biological selves and artificial systems. Since we started writing, technology and its applications have advanced significantly, leading us to create the sixth edition. These rapid advancements potentially mark the start of a new human-triggered \textit{anthropocene} \footnote{Anthropocene is not an officially recognized unit of measure, but it is a useful tool to explain the magnitude of the repercussions \cite{bbcAnthropoceneUnit}.}. Similar to the \textit{Trinity Bomb} test that marked the beginning of the \textit{Atomic} age or the invention of the \textit{Newcomen Atmospheric Engine} that brought about the \textit{Industrial Revolution}, Artificial Intelligence appears to be on the cusp of introducing a new \textit{Knowledge} age\textemdash the prospect of a \textit{Cambrian like explosion} of ideas and human progress. \par

As excitement grows, so do concerns and realizations. As technology companies accelerate toward this new era, a new \textit{Pandora's box} \cite{theguardianGodfatherGeoffrey} opens. \textit{What problems does this newly opened box bring?} Normalization \textit{of knowledge} (making everything the same) and \textit{slopware} may be among the lesser concerns, while \textit{misinformation}, \textit{disinformation},  excessive datacenter expansion (consuming the worlds resources in the process), and employment disruptions \cite{ai20272027} are among the more serious ones. Our concerns may even extend to how we perceive money and wealth in the future \cite{BSB2025}. As Artificial Intelligence becomes more centralized, uniform, and potentially manipulative, it risks constraining human development by concentrating influence and control in a small number of powerful actors. These concerns continue to grow as Artificial Intelligence becomes more capable, affecting everything from jobs to education. Together, this excitement and these concerns bring us back to our original question from 2020, \textit{What is intelligence?} \par

Intelligence is a fundamental part of most living things and the foundation for Artificial Intelligence. In this primer, we explore the ideas associated with intelligence and, by doing so, understand the implications and constraints of future systems. Today, Artificial Intelligence, in the form of machine learning, has significantly impacted our lives and will continue to do so for the foreseeable future. Our demand for more data, intelligently handled, with ever-increasing computational efficiency, continues to act as the driving force for progress. \par
 
   As an exploration, we take a journey into the essential domains of intelligence. We hope that people find this helpful in determining the future. In addition, during the exploration, we hope to create new thought-provoking questions. Intelligence is not a single weighable quantity but a subject that spans Biology, Physics, Philosophy, Cognitive Science, Neuroscience, Psychology, and Computer Science. \par

The historian \textit{Yuval Noah Harari} has argued that engineers and scientists of the future will need to expand their knowledge beyond technical fields to include topics such as Psychology, Philosophy, and Ethics. Fiction writers have long portrayed engineers and scientists as lacking in these areas, but this is set to change. The rise of Artificial Intelligence, combined with new legal requirements, is pushing these topics from the fringes to the center of debate. \par 

We begin this journey with an introduction to intelligence and quickly move on to more profound thoughts and ideas. We call this a \textit{Life, the Universe, and Everything} \cite{DOUGLASADAMS} primer, after the famous science fiction author \textit{Douglas Adams}. Forty-two may be the correct answer, but what are the questions?

\end{abstract}


\settopmatter{printacmref=false}
\setcopyright{none}
\renewcommand\footnotetextcopyrightpermission[1]{}
\pagestyle{plain}


\begin{teaserfigure}
  \centering
  \includegraphics[width=\textwidth]{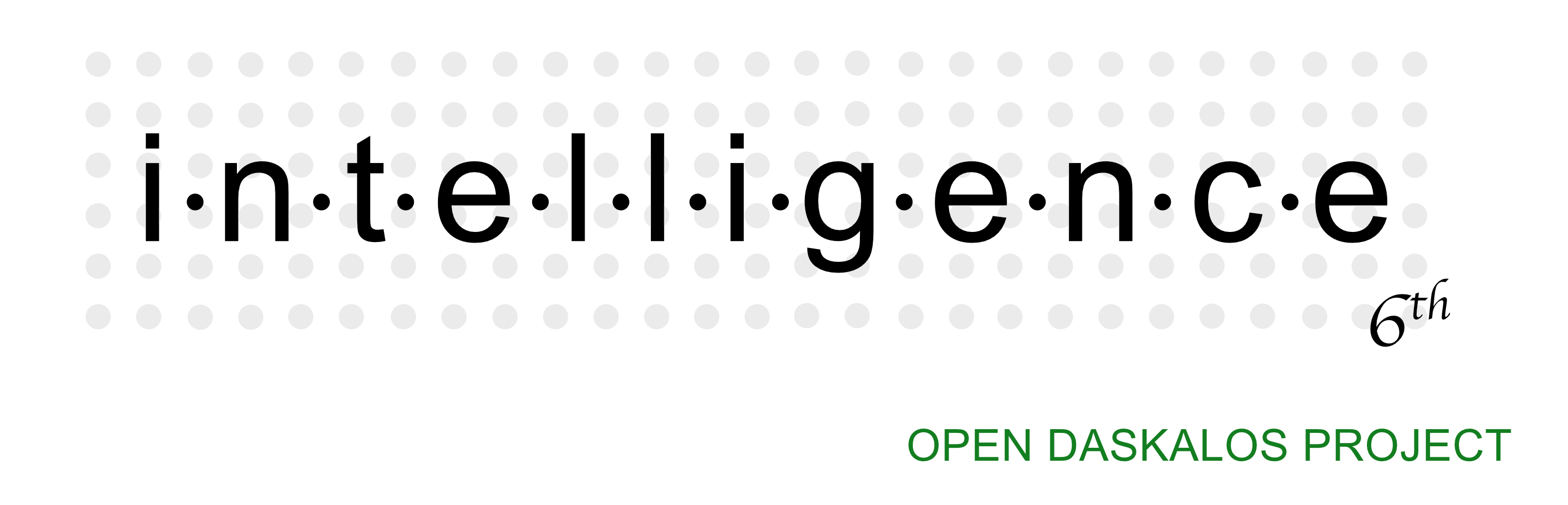}
  \label{fig:teaser}
\end{teaserfigure}

\maketitle

\section{Introduction}\label{sec:introduction}

\vspace{1.0em}

\begin{adjustwidth}{0.3cm}{}
\textbf{>> First, we examine the foundations of intelligence, exploring why it arises, why it exists, and which attributes make it difficult to define in any simple, unified way.}
\end{adjustwidth}

\hspace{0.5cm}

\begin{quote} 
\centering 
\vspace{1ex}
\textit{\say{Perhaps the unintelligent machines can do even better than the intelligent ones}}\\
\textbf{Richard Feynman, Lecture, 1985} \cite{Feynman85}
\end{quote}
\hspace{0.5cm}

We start with our main question: \textit{What is intelligence?} We and other organisms use intelligence to find food, avoid dangers, remember patterns, provide survival instincts, solve problems, and communicate with other entities. Intelligence simulates the past and the future. Simulations allow for past narratives and predictions of the future. With these simulations, we can plan successful actions and perceive possible known errors. We can place ourselves and others in the simulator. Intelligence creates opposites: kindness \& cruelty, love \& hatred, and reasoning \& madness. At its core, it cares about self-preservation, reproduction (copying), and self. It can learn and remember; It can improve and forget\textemdash all of which we view as biological traits. \par

By comparison, we view Artificial Intelligence as different in that it mimics our intelligence. It is not formed through biological processes that span hundreds of thousands, if not billions, of years, but through mathematics, algorithms, and computers over a relatively short period of time. It is our construction. We view our artificial cousins as \textit{soulless philosophical zombies} \cite{koch2012consciousness}, while biological entities have consciousness, understand consequences, provide emotions, and, most importantly, have a purpose. With these differences, we naturally place ourselves at the epicenter of the intelligent universe.\\

\begin{figure}[ht]
  \centering
  \includegraphics[width=0.8\linewidth]{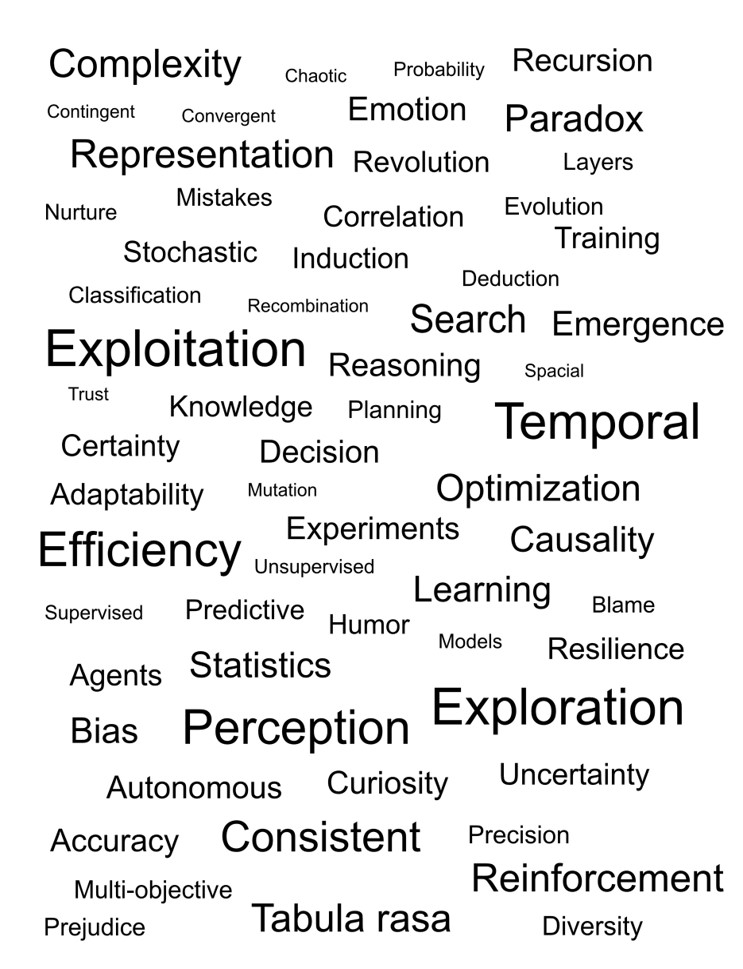}
  \caption{Words associated with intelligence (circa 2019)}
  \label{figure:aiwordz}
\end{figure}

\textsc{\textbf{Why is intelligence so difficult to describe?}} It comes down to whether a system under study can be entirely and self-consistently understood when we are the same, the research object and the researcher. Intelligence is complicated, and not in a small way. The biochemist \textit{Efraim Racker} eloquently put it: \say{\textit{anyone who is not thoroughly confused doesn't understand the subject}}. We suffer from introspection, where the system under study and the observer are the same (\say{\textit{us}}). This self-introspection causes natural hubris when it comes to explaining intelligence and makes it difficult for us to identify other forms of intelligence. \par

We start our journey with confusion, see Figure \ref{figure:aiwordz}, and end with the wrong numbers in Section \ref{sec:wrong numbers}. We hope to transfer some exciting thoughts, ideas, and challenges throughout the journey. \\

\begin{figure}[ht]
  \centering
  \includegraphics[width=0.7\linewidth]{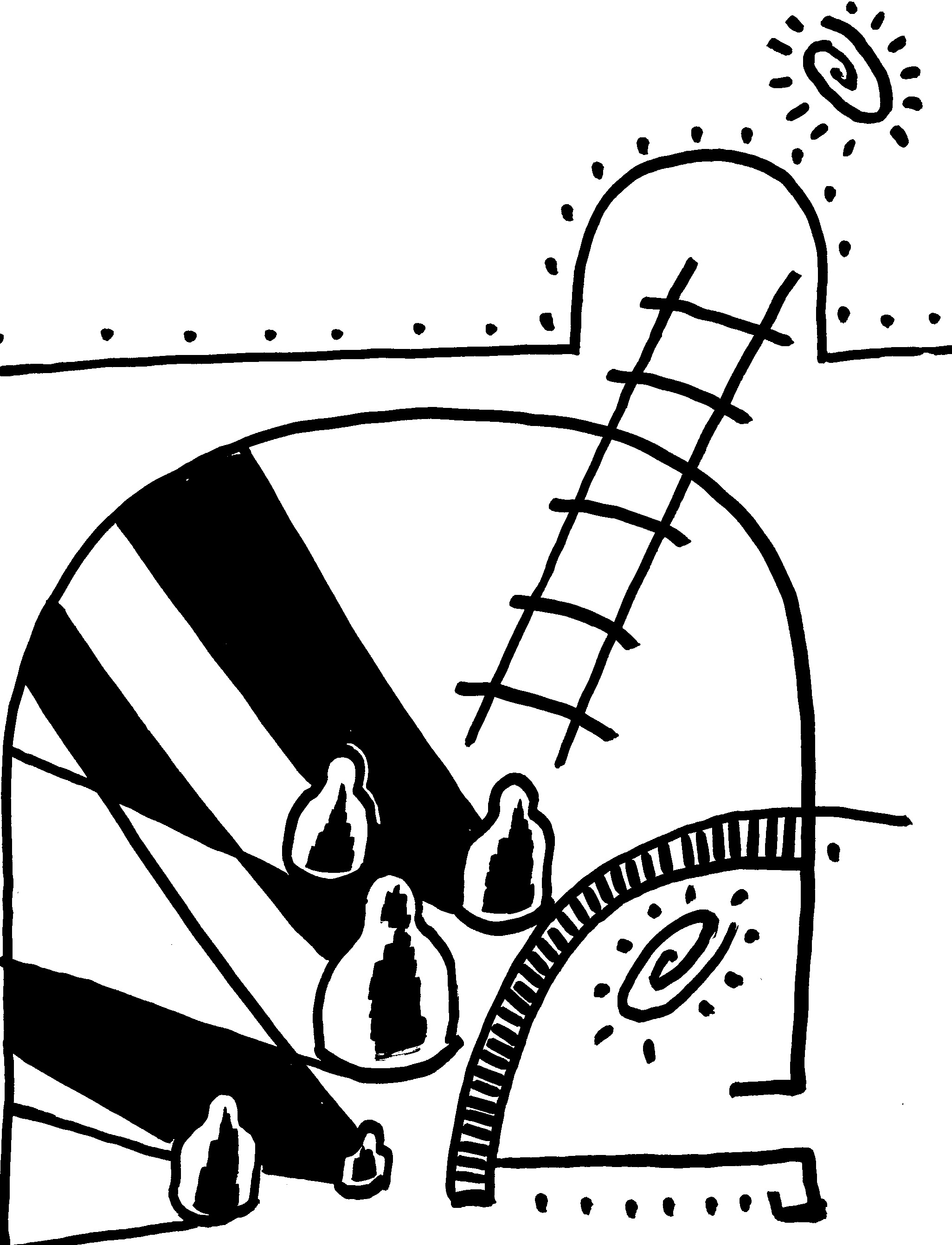}
  \caption{Plato's Cave by Rex St. John}
  \label{figure:platocave}
\end{figure}

\textsc{\textbf{Introspective:}} As discussed, intelligence is complex due to our deep involvement. To help put this in some form of perspective, we use a scenario introduced by \textit{Plato} in his work called \textit{The Republic} (514a$-$520a). The scenario is called \textit{Plato's Cave} \cite{respublica}, see Figure \ref{figure:platocave}. Plato depicts a cave with a set of prisoners. Each prisoner faces forward in one direction, separated, bound to one another, and without communication. Behind them shines a bright candle. From their perspective, their entire world consists of silhouettes of themselves. Their world and realization of self are narrow and distorted, as people outside the cave view them. Intelligence suffers from the same problem since we consider the subject from our perspective with our biases. We do not have the luxury of viewing intelligence objectively, which forces introspection\textemdash also known as the \textit{$n=1$ problem}, where the only example is ourselves \cite{green2023possibility}.  \par

When we discuss intelligence, what are we trying to understand? Is it to find the purpose of intelligence? What components make up intelligence? How can it be applied? Discover the limitations? Measure the capability? How to build it? Or even how to control it? Each question is essential and complicated. But the question we should probably start with is: \textit{Why should we care?}.\\

\textsc{\textbf{Why should we care?}} In addition to the fact that intelligence increases the likelihood of \textit{survival}, it also empowers humans to handle abstract thoughts. One such thought is, \say{\textit{why has something occurred?}}' explaining the cause of a situation. We will discuss the idea of \textit{cause-and-effect} throughout this primer, as it touches on many concepts associated with intelligence. Another reason to care is that artificial versions can augment or even replace our intelligence, e.g., extend our sensory capabilities such as eyesight, hearing, and data processing. We can use intelligence to push the boundaries of knowledge and exploration. Lastly, in theory, intelligence brings resilience, higher reliability, and, we hope, better decision making.\\

\textsc{\textbf{Creativity:}} We use creativity to solve problems by understanding the environment and problem domain to find a novel solution. We would have already created the most advanced intelligent system if we had viewed intelligence as carrying out tasks. Even with traditional rule-based systems, we can effectively repeat complicated tasks. The challenge is to handle new problems and new situations. The handling of new cases is why creativity comes into the equation. It is one of the significant attributes that separate animal species. It is also how we solve new problems that are either fully or partially visible. \par

Creativity can take many forms: playing football, painting, composing music, solving mathematical equations, or designing new User Interfaces. The list is endless. There are many cultural and social aspects associated with creativity. An essential factor is moving humans beyond primary survival objectives to other areas such as art, music, and esthetics. In other words, secondary reasons are sought, not just the primary goals of life. \par

It is arguable whether there is creativity in other animals. For example, is a bird song creative? Is it an expression of bird-self? Or is it simply a prescribed sequence of musical notes that a bird uses to maximize its chances of mating? Does a bird song purely satisfy the functional purpose of optimizing the effects of mating or mimicking? Are humans alone in creativity?\par

What is creativity? \textit{Margaret Boden}, a professor of cognitive science, divided creativity into three useful mechanisms \cite{Boden2013}: namely, \textit{exploration}\textemdash (playing within the rules), \textit{combination}\textemdash (applying one set of rules to another domain), and \textit{transformation}\textemdash (rewriting the rules by removing a critical constraint). For intelligence, exploration creativity is risk-averse and limited, while at the other end of the scale, transformation creativity has the highest risk of novelty.\par

Humans can achieve all three levels. In contrast, current Artificial Intelligence struggles with anything beyond exploratory; it never ventures beyond its original programming. If Artificial Intelligence is going to be part of our society, does it need to be creative? Do we want it to be creative? For now, creativity serves as a distinct difference between human-level intelligence and machine intelligence.\\

\textsc{\textbf{Are we living in a trusted simulator?}} \label{sec:trusted simulator} Are we? Maybe. René \textit{ Descartes} (circa 1637) struggled with this question and went to enormous lengths to create an answer, culminating in the famous line from \textit{Discourse on Method}: \say{\textit{Cogito, ergo sum}}, or translated to \say{\textit{I think, therefore I am}}. Regardless of the answer, intelligence plays the same role in real and simulated situations. A simulation can replace everything in the world, and the underlying decision processes of the individual would remain unchanged. Simulation is critical if we want to mimic intelligence. We will return to this topic in Section \ref{sec:consciousness} \& \ref{sec:wrong numbers} on \hyperref[sec:consciousness]{\textit{Consciousness}} and \hyperref[sec:wrong numbers]{\textit{Wrong Numbers}} respectively.\par

We can only trust our ability to think; everything else (i.e., input from our environment) is considered untrustworthy. Untrustworthy inputs place \textit{skepticism} at the forefront of intelligence and lead to a tendency towards \textit{solipsism} (only one mind exists). Skepticism of Artificial Intelligence is a valuable quality, and it encourages developers to reduce potential errors by having more input data of different types and from different sources to lessen concerns about possible errors. Simultaneous correlation of dispersed sensors is one of the best reasons for the \textit{Internet of Things}.\par

We may or may not live in a trusted simulator, but simulation may be essential for who we are and, more importantly, for intelligence in general. Our memory is episodic, which means that we build past or future simulations \cite{bennett2023brief} through experiences and emotions. We place ourselves in those simulations to help ascertain the past (i.e., causal history) or the future (i.e., prediction). We can introduce others into the simulator, such as criminals, friends, animals, or competitors. It can spark imaginative scenarios, allowing us to vary the distance from reality. We can play those simulations in our minds.\par

Certain brain abnormalities or injuries may cause a person to have the inability to place themselves in the simulator or visualize others in the simulator, removing empathy \cite{bennett2023brief}. Morality relies on empathy \cite{pollan2026world}. The lack of empathy in people is concerning, but it could become a major issue with Artificial Intelligence. \\  

\textsc{\textbf{Induction, correlation, causality, and inference:}} \textit{Computational functionalism} \footnote{orthogonal to \textit{humanism}, which deals with uniqueness and ethics.} relies on four fundamentally different ideas: induction, correlation, causality, and inference. As an important step, let us dive into these four ideas. The first idea is \textit{induction}. Induction takes what has previously occurred as a pattern and extends it into the future. Mathematics employs this technique to prove equations; for example, if we can prove $f(0)$ and $f(1)$, then in theory, we can prove $f(n)$ and $f(n+1)$. Using this proof mechanism, we can determine whether a recursive function converges to a solution or diverges from it. This method is called \textit{proof by induction}. We use this technique to justify future actions because the past has shown that future actions can be determined. This form of intelligence uses historical patterns to predict the future. It is worth noting that the philosopher \textit{David Hume} spent significant effort thinking about \textit{The Problems of Induction} \cite{sep-induction-problem}. We should also note that there are methods of \textit{contradiction} and \textit{contrapositive} related to induction, which we will not cover. \par

Next comes correlation. Humans spend a significant amount of time \textit{correlating} information. We take known inputs and attempt to match the patterns. The pattern may be an image, a sound, or a feeling. These correlations can be very sophisticated, such as face recognition or partial recognition, where only a small part of the information is visible. This ability to correlate comes with the added feature of learning new patterns. \par

We should also mention \textit{apophenia}, which is the ability to find meaning and connections in cases where there are no relationships\textemdash seen as things that are not there \cite{verywellmindWhatDoes}. Note that Artificial Intelligence easily suffers from the very same problem. Another, more narrow issue is \textit{patternicity}, which is the tendency to find patterns in random data, especially associated with visual recognition\textemdash imagining a face on toast or on the surface of Mars (known more specifically as \textit{pareidolia}) \cite{verywellmindWhatDoes,scientificamericanPatternicityFinding}. \par

Machine learning, particularly the concept of \textit{Deep Learning}, has progressed by leaps and bounds in recent years. After years of quietness, the field of artificial neural networks (ANN) suddenly became a subject of intense research. \textit{Geoffrey Hinton}, a computer scientist, coauthored a famous paper in 2012 titled \textit{ImageNet Classification with Deep Convolutional Neural Networks} \cite{NIPS2012_4824}. Hinton et al. showed that an algorithm for artificial neural networks could exceed human ability to correlate. Using a lot of training data, an artificial neural network could identify patterns like a human. This Deep Learning approach was superior to all other methods when \textit{AlexNet} won the ImageNet 2012 (LSVRC-2012) competition. \par

Causality is the next important concept. \textit{Causality} is discussed in \textit{The Book of Why: The New Science of Cause and Effect} (2018) \cite{Pearl} by \textit{Judea Pearl} and \textit{Dana Mackenzie}. Pearl and Mackenzie conjectured that humans are better at \textit{cause-and-effect} than at statistics-and-probability. They go on to point out that causality differs significantly from correlation. Correlation is the mapping of input data to output via a known pattern recognizer. In contrast, causality is about determining why. The mathematics around causality involves creating a unique model. The model connects data to a set of actions. Causal relationships predict the future by testing \textit{counterfactual} arguments. A counterfactual argument contradicts facts in a hypothetical future world or state. This is best described in a statement: \textit{if $X$ then $Y$}, where the conditional clause $X$ is false.  It allows intelligence to learn not just from what happened, but also from what might have happened.\par

Lastly, we introduce \textit{inference}. Inference is an umbrella concept. It is the step where a system uses the fundamental concepts described: correlation, induction, and causality. It adds prior knowledge and context to decide what could be true and what possible actions could occur.

\subsection*{\textbf{5-Thoughts}}
\begin{itemize}[leftmargin=*]
    \item We use intelligence to discover the unknown by applying some form of known knowledge or experience. It is easier to describe intelligence from a behavioral point of view (\textit{what does it do?}) than from the implementation (\textit{how is it accomplished?}). We are the system under investigation,  the introspection of ourselves. \textit{What could possibly go wrong?}
    \item There are indications that natural intelligence relies on internal simulation to recall the past and project into the future. To remain efficient in both time and space, these biological simulators simplify, omit information, and produce interpretations of the world and the entities within \cite{bennett2023brief}. Good enough to be useful. 
    \item Induction helps us predict the future from the past. Correlation is about matching an input to a known pattern. Lastly, causality is understanding the logical events that led to an event. 
    \item Does intelligence do \textit{more with less}, or \textit{more with more}? It would seem that intelligence should do more with less. Less input and more insightful speculation. More with more would suggest that the larger the knowledge base, the greater the intelligence, but nature appears to have gone in different directions, most notably by improving the learning process \cite{InterviewDavidKrakauer}. As \textit{David Krakauer}, biologist, put it, \say{\textit{intelligence should do a lot with very little}} \cite{krakauer2024complex}.
    \item The speed of adaptability is related to intelligence. Whether it is adapting to a new situation or a new experience. In essence, intelligence can be defined by the latency, process, and optimization of adaptation.
\end{itemize}

\subsection*{\textbf{Recommended reading}}

\begin{itemize}[leftmargin=*]
    \item \textit{The Republic}, by Plato 
    \item \textit{The Book of Why}, by Judea Pearl \& Dana Mackenzie
    \item \textit{The Creativity Code}, by Marcus Du Sautoy
    \item \textit{Introduction to Philosophy}, by John Perry et al.
    \item \textit{A Brief History of Intelligence}, by Max S. Bennett
    \item \textit{The Complex World}, by David C. Krakauer
\end{itemize}

\section{History}\label{sec:history}

\vspace{1.0em}

\begin{adjustwidth}{0.3cm}{}
\textbf{>> History allows us to understanding where we come from in order to see where we are, and even to hint at where we may go. Marking the progress made by nature and ourselves in building intelligent systems, reminding all of us that we never truly reach an endpoint.}
\end{adjustwidth}

\hspace{0.5cm}

\begin{quote} 
\centering 
\vspace{1ex}
\textit{\say{I have my hopes, \& very distinct ones, too, of one day getting cerebral phenomena such that I can put them into mathematical equations: in short, a law or laws for the mutual actions of the molecules of the brain (equivalent to the law of gravitation ...}}\\
\textbf{Ada Lovelace, Software Engineer, circa 1840} \cite{lovelace1843notes}
\end{quote}

\hspace{0.5cm}

This section is our attempt to create a rough history of intelligence; like all histories, it is a narrative to help explain some form of causal path. It is highly probable that we missed something important, so treat this as our disclaimer. In addition, many advances occur in parallel, but we describe them as linear sequences. \par

Determining where to start with the history of intelligence is difficult, so let us start with biological life first. \textit{Origin} and \textit{LUCA} are two important points in biology. \textit{Origin of life}, or \textit{abiogenesis}, is when life first appeared on Earth. Did it start at a single point or multiple points? Unknown? LUCA, the \textit{Last Universal Common Ancestor}, is the most recent ancestor of all living organisms. We recently estimated that the last ancestor may have occurred 4.2 billion years ago \cite{Moody2024}, far earlier than originally thought, with abiogenesis predating LUCA.\par

We can guess that there were probably many failed attempts at creating complex biological life before resting on a workable method, i.e., LUCA. These historical events can be used to guide our search for intelligence\textemdash; we can use our own historical path as a possible reference pattern for the search for intelligent life on other planets.

\subsection*{Biological transitions:}

We can guess that the first brain appeared in a small worm \cite{bennett2023brief}, around 600 million years ago; unfortunately, we do not have a name for this animal, which has the distinction of being the first intelligent life form on planet Earth. However, we can see significant transitions that culminate in the modern brains we see today. We summarize these transitions from the increase in \textit{cellular complexity} to the appearance of \textit{primates and hominids}. Each transition builds on the previous leap, eventually allowing humans to emerge as a dominant species, or so we believe. \\

\begin{enumerate}[label={B-\arabic*:},leftmargin=*]

\item \textsc{\textbf{Eukaryotic:}} The first significant increase in complexity occurred within cells; it involved the transition from \textit{prokaryotic} to \textit{eukaryotic} cells, which we estimate occurred approximately 1.9 billion years ago. Before this time, cells (prokaryotic) lacked subcellular or compartmental structures called \textit{organelles}. Organelles perform one or more specific functions within a cell. This increase in functionality represents \textit{endosymbiosis}, a symbiotic relationship where an organism lives within another. For mutual advantage. That led to  eukaryotic organelles like \textit{mitochondria} and \textit{chloroplasts} to appear. Eukaryotic organelles have this feature, which increases the biochemistry complexity of cells\footnote{In a 2025 paper titled, \textit{An Asgard archaeon from a modern analog of ancient microbial mats}, \textit{Stephanie-Jane Nobs} and team describe growing a rare deep-sea-relative microbe (an \textit{Asgard archaeon}) from a microbial mat. This was the first time we saw a microbe physically connect to a paired bacterium. Tiny tubes connected the microbe to the bacteria, allowing them to swap nutrients. The microbe was observed to have complex cell \textit{like} features normally associated with eukaryotic cells. The significance is that this is the clearest living potential ancestor of all complex life before life jumped to the next level \cite{Nobs2025AnAA}.}. \\

\item \textsc{\textbf{Neuron:}} This complexity allowed the first neuron to appear about 650 million years ago. We view neurons as specialized cells that transmit information through electrical and chemical signals. Six hundred million years ago, multicellular organisms started to appear. They cooperated and specialized, allowing organisms to take on different roles. This enhancement allowed biology to take on more complex structures and introduced the first nervous system. \\

\item \textsc{\textbf{Nervous system:}} The first nervous system is believed to have appeared during the \textit{Ediacaran period}, approximately 543 million years ago. \textit{Nerve} cells and simple \textit{neural networks} appeared, allowing organisms to process information quickly and respond to hunger or danger. The transition to a nervous system marked a leap in complexity and capability. The biological world now consisted of sophisticated cells that collaborated to handle specialized tasks and could form rudimentary systems to process information. \\

\item \textsc{\textbf{A brain:}} During the same period, more complex brains started to appear in animals. These animals were distinct in that they had a vertebral column and a cranium, or skull (known as \textit{vertebrates}). The central nervous system became significantly more complex and the brain expanded with more specialization. These changes lead to a brain that can handle more complicated processing, retain memories, and gain learning capabilities, as seen in many animals today. \\

\item \textsc{\textbf{Homo habilis:}} Following these progressions, \textit{primates} and, in turn, \textit{hominids} emerged. We estimate that this occurred about 2.4 million years ago with the appearance of \textit{Homo habilis} in East Africa. East Africa is an important location for many developments associated with intelligence. These primates had noticeable brain expansion, allowing for basic organization. This change led to higher levels of social interactions, such as shared tool usage and language communication, as well as thinking in the abstract\textemdash what we call modern intelligence. \\

\end{enumerate}

For simplicity, this movie script is heavily abridged. The movie is about the history of biological intelligence; a full film would show many more plot twists and characters. The transitions provide a rough summary of how we converged on what we have today \cite{calcott2011major,lane2005power}. A currently unanswerable but interesting question is: \textit{what happens if we can replay the movie of life?} \textit{Do we converge on the same transitions?} History always repeats. \textit{Do we converge on different transitions, or no transitions?} History never repeats. Or alternatively, \textit{does the movie diverge?} History is just a one-time set of events. We appear to be, through whatever process got us here, extremely rare in the universe. We are alone and remain alone in the universe. Unfortunately, we only have one movie to watch, and most of the interesting early plot twists have already occurred, which means that we have to piece together the missing starting scenes using archeological evidence. We do our best to guess the most likely scenarios; in other words, we \textit{speculate}.

\subsection*{Artificial transitions:} 

The history of Artificial Intelligence differs significantly from that of its natural counterpart because, in some respects, we are experiencing a compressed version of that development in real time. The 6th Edition of this paper reflects these ongoing changes. Convergence has yet to occur. Again, it is difficult to determine where to start, so let us pick a fake vision as the starting point. \par

The \textit{Mechanical Turk} was built in 1770 by the writer and inventor \textit{Wolfgang von Kempelen}. The claim made by von Kempelen was that the mechanical device could play chess and win games. People suspended their disbelief, hoping that this mechanical device could play chess. It was an illusion since someone was operating the machine from underneath the table. From the outside, it appeared to be autonomous, but in reality, it was a lie. It identified our appetite, even in 1770, for autonomous systems. It is also one of the first indicators that humans have a tendency to ascribe intelligence to objects that may or may not be intelligent. More on that topic later; see Section \ref{sec:consciousness}\par

\begin{figure}[ht]
  \centering
  \includegraphics[width=0.65\linewidth]{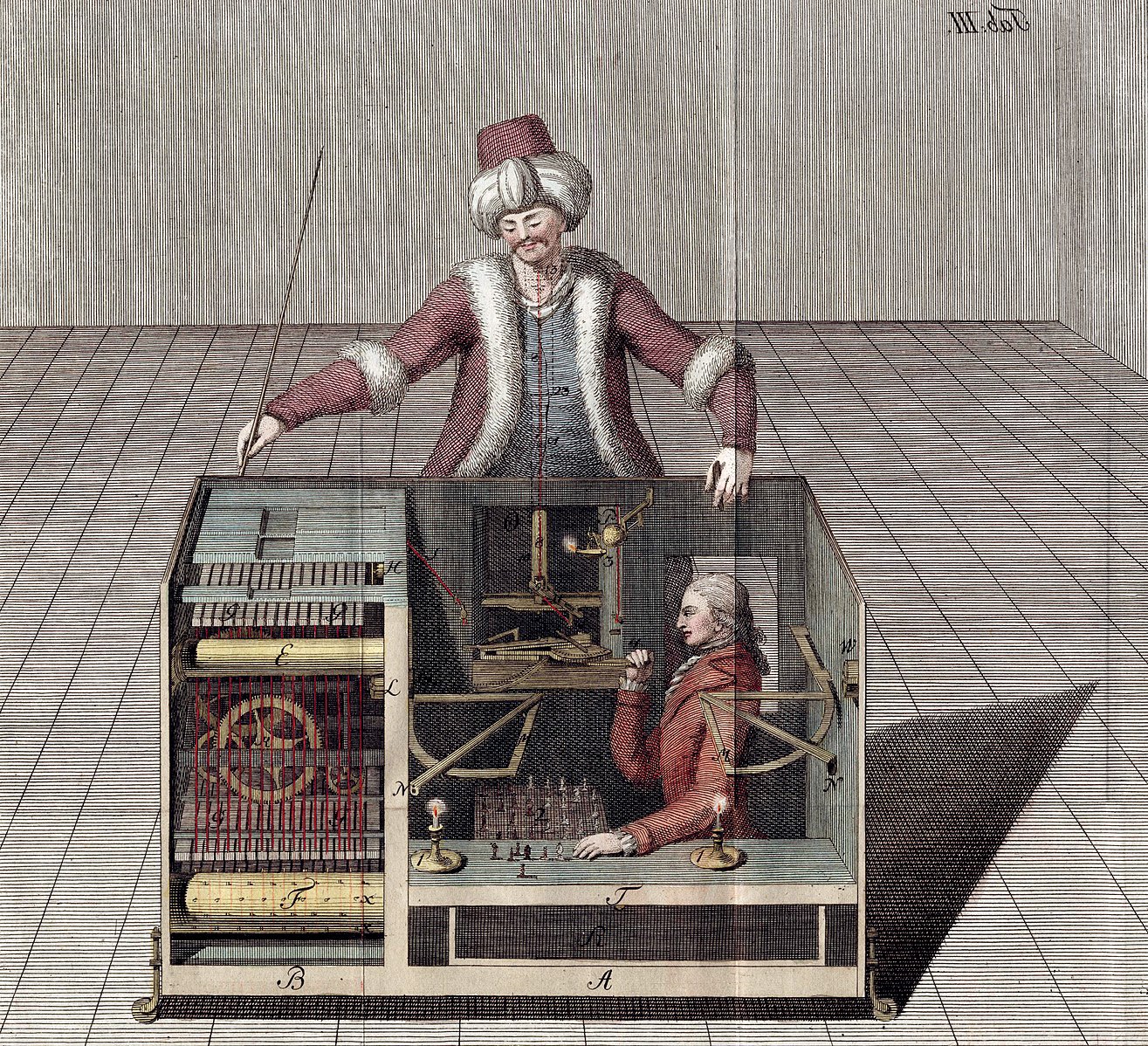}
  \caption{Mechanical Turk \cite{racknitz_turk_3} (Wikimedia Commons)}
  \label{figure:turk}
\end{figure}

Many parallel technological developments opened the door to modern machine learning. For the mainstream, the rise of electronic circuits, advanced algorithms, the Internet, and the development of Artificial Intelligence are all closely related. Other developments have the potential to improve Artificial Intelligence even further; these include a better understanding of biology, biochemistry, unconventional computing, and mathematics. Complete convergence still has to occur, especially since energy efficiency is a significant constraint with current solutions. \\

\begin{enumerate}[label={A-\arabic*:},leftmargin=*]

\item \textsc{\textbf{Binary:}} \textit{George Boole}, mathematician, published \textit{An Investigation of the Laws of Thought} in 1856, describing how to handle binary logic \cite{boole1854investigation}. \textit{Gottfried Wilhelm Leibniz} published in 1703 a method for binary code to convert logical statements into 1s and 0s \cite{strickland2022leibniz}. Both publications were important because they outlined how to manipulate a universal datatype. Boolean and digital logic are the current underlying language for Artificial Intelligence. With more unconventional forms of computation, this may change in the future, as the drive to more efficiency becomes paramount. \\

\item \textsc{\textbf{McCulloch-Pitt Neuron:}} To implement the mathematical language of logic more effectively, Claude Shannon, a mathematician, created the concept of \textit{logic gates} in 1937 that, when connected, form more complicated systems. These complicated digital systems are similar to biological cells that connect to form specialized functions. In 1943, the first mathematical model of a neuron was created by \textit{Warren McCulloch} and \textit{Walter Pitts} \cite{mcculloch1943logical,griffiths2026laws}. It was named the \textit{McCulloch-Pitt Neuron Model}. It provided the foundations for artificial neural networks. It used simple threshold logic. \\

\item \textsc{\textbf{Information theory:}} In 1945, \textit{John von Neumann} published \textit{First Draft of a Report on the EDVAC} describing a universal computer architecture that has become the core of all digital systems today \cite{danton2025computers}. These developments allowed for the construction of computational machines that could read instructions and modify data. \textit{Alan M. Turing}, whom we will discuss later, presented a lecture in London in 1947, describing a \textit{machine that could learn from experiences by altering its own code}. Then, in 1948, Claude Shannon made a significant step forward by publishing the famous paper about  \textit{information theory} that described a mathematical framework for quantifying, storing, and communicating information. Shannon later co-organized, 1956, the first Artificial Intelligence workshop at \textit{Dartmouth College}, called \textit{Summer Research Project on Artificial Intelligence}. \\

\item \textsc{\textbf{Perceptron:}} Artificial Intelligence comes from combining visionary ideas with important practical advancements in implementation. Each step forward was an increase in scale, complexity, and understanding. Some of those patterns are related to the history of biological intelligence, but one of the main differences is the time scales. In 1958, \textit{Frank Rosenblatt} proposed the \textit{perception model}. The perceptron was the first practical neural network capable of learning. We could place individual neurons into an interconnected network. It could accomplish binary classification using a neuron-like structure. \\

\item \textsc{\textbf{Backpropagation algorithm:}} After many disagreements (more commonly known as the \textit{first AI Winter}), the next significant milestone came in 1986, from Computer Scientists \textit{David Rumelhart}, \textit{Geoffrey Hinton}, and \textit{Ronald J. Williams}, who published the backpropagation algorithm. An algorithm that could train multilayered neural networks efficiently. Allow for more complexity in the form of deeper architectures. \\

\item \textsc{\textbf{Deep learning:}} Hardware changed significantly in 2010, i.e., \textit{Symmetric Multi-Processing} became standard. Allowing systems to do more complex parallel processing. With this new computing capability, in 2012, \textit{Alex Krizhevsky}, \textit{Ilya Sutskever}, and \textit{Geoffrey Hinton} created \textit{AlexNet}, which was a breakthrough in \textit{Deep Learning}, AlexNet, as previously discussed, outperformed previous methods for image recognition using \textit{Deep Convolutional Networks}. \\

\item \textsc{\textbf{Generative Adversarial Network:}} Following the success of Convolutional Networks, machine learning development accelerated with the formation of a \textit{Generative Adversarial Network} (GAN) appearing in 2014. \textit{Ian Goodfellow} et al., Computer Scientists, introduced GANs \cite{goodfellow2014generativeadversarialnetworks}; we will discuss GANs later; an important evolutionary step in machine learning where two neural networks argue for the purpose of generating realistic images, videos, and data. Opening the world to more creativity and potentially tools for disinformation through fake media. \\

\item \textsc{\textbf{Transformers:}} In 2017, transformers and natural language processing started to appear. \textit{Vawani et al.} published a paper called \textit{Attention is All You Need} \cite{vaswani2017attention}. Transformers brought a breakthrough in next-level text interaction and response generation. The most famous user interface being \textit{ChatGPT}.

\end{enumerate}

The rest of the paper covers these subjects and others in more detail, but first, we have a special call out to Alan Turing. \\

\textsc{\textbf{Alan M. Turing:}} \textit{Alan Turing} can be said to be one of the founders of modern Artificial Intelligence. He made significant contributions to our understanding of natural and Artificial Intelligence, earning him an important place in history. He covered the fundamental operations of computing, provided a measurement for intelligence, and described his ideas on how to build that intelligence. These achievements occurred in an era where the available hardware was significantly limited. \par

In the fundamental operations of computing, he devised a theoretical machine called the \textit{Turing Machine} (TM), where digital computation is a small set of essential functions. It is also worth mentioning that biological organisms use a variant of the Turing Machine called a \textit{Random Access Machine} (RAM). The significant difference between Turing machines and Random Access Machines is that a RAM-based system can access arbitrary tape locations. Turing Machines can convert to Random Access Machines, but this is not bidirectional; not all Random Access Machines convert to a Turing Machine. Also, it is worth noting that in biology, the hardware is numerous (there are many different types of neurons). The software is the same (DNA genome), whereas in digital computers, the hardware tends to be limited (only a few types of digital neurons) and the software provides uniqueness. Both biological hardware and mutation-crossover provide the variation in organisms. \par

In \textit{Computing machinery and intelligence} \cite{TURING1950}, released in 1950, Turing explored the idea of creating Artificial Intelligence. He determined that a system would require the ability to make errors or mistakes, i.e., allow for creativity, and must include some form of randomness as a critical component. \par

Finally, Turing envisaged a test that can measure whether Artificial Intelligence has been achieved, either by a system with high-performance mimicking or by an actual biological conscious intelligence. Today, this is commonly known as the \textit{Turing test}; see Section \ref{sec:measuring} for more details.\\

\subsection*{\textbf{8-Thoughts}}

\begin{itemize}[leftmargin=*]
    \item There are many differences and surprising similarities between biology and Artificial Intelligence. The difference lies in the time scale; for biology, timescales are significantly longer, but in terms of similarity, increasing complexity and experimentation have allowed both to become more capable.
    \item We have an inherent willingness to suspend disbelief in Artificial Intelligence by ignoring significant deficiencies in the technology. In a general sense, we have \textit{apophenia}, we tend to observe and connect things that are unrelated. We want Artificial Intelligence to exist and to have a conscious state. What it currently is, is a good pattern matcher. It certainly has great potential, but it is not a panacea. Biology is resilient and adaptable, while, for the most part, Artificial Intelligence is fixed, fragile, and inefficient.
    \item We see a symbiotic pattern between understanding the brain (our observations) and advances in Artificial Intelligence (implementation) \cite{bennett2023brief}. \textit{At what point do artificial systems completely diverge from biology and produce an entirely new form of conscious life?}
    \item The influence of science fiction on our view of Artificial Intelligence is worth mentioning, especially when it comes to providing a future vision. The \textit{HAL 9000} computer, from the 1968 movie \textit{2001: A Space Odyssey}, played a significant role in setting our expectations by envisioning an intelligence with cognitive skills (voice, video, physical interaction) and providing the ultimate warning that all intelligence is fallible \cite{stork1997hal}.
    \item Evolution may explain our ability to suspend disbelief when it comes to seeing consciousness or intelligence in inanimate objects. \textit{Hyperactive agency detection} (or HADD for short) is believed to be a cognitive bias that causes us to interpret ambiguous information as consciousness or intention, rather than as a random event \cite{pollan2026world}.    
    \item Vision alone must be supported by empirical evidence. The principles of how we demonstrate improvements in Artificial Intelligence are crucial for future advancements.
    \item There are many thoughts about why life became more intelligent. It could be due to environmental stress and the need for survival. Or through \textit{punctuated equilibrium}, which is the hypothesis that a short, rapid speciation occurred, producing increased intelligence. Or through \textit{physical graduality}, where speciation occurred through gradual slow transformations over a long period of time. Or a combination of both. We will return to this topic a number of times. 
    \item Living things have a universal goal to survive and reproduce; there are consequences to failure \cite{pollan2026world}. For artificial systems, the goals and consequences are less clear.
    
\end{itemize}

\subsection*{\textbf{Recommended reading}}

\begin{itemize}[leftmargin=*]
    \item \textit{The Major Transitions in Evolution Revisited}, by B. Calcott et al.
    \item \textit{Introduction to Philosophy}, by John Perry et al. 
    \item \textit{Human Compatible}, by Stuart Russell
    \item \textit{A Brief History of Intelligence}, by Max S. Bennett
    \item \textit{Power, Sex, Suicide}, by Nick Lane
    \item \textit{HAL's Legacy}, by D. G. Stork
    \item \textit{The Laws of Thoughts}, by Tom Griffiths
    \item \textit{the Computers that made the World}, by Tim Danton
    \item \textit{A World Appears}, by Michael Pollan
\end{itemize}

%
%

\section{Human intelligence} \label{sec:human}

\vspace{1.0em}

\begin{adjustwidth}{0.3cm}{}
\textbf{>> Time to explore the human dimension of intelligence and how we position ourselves within the broader landscape. This includes ideas and perspectives on the nature of our intelligence.}
\end{adjustwidth}

\hspace{0.5cm}

\begin{quote} 
\centering 
\vspace{1ex}
\textit{\say{I know that I am intelligent, because I know that I know nothing.}}\\
\textbf{Plato's Apology, around 399–387 BCE} \cite{plato_apology}
\end{quote}

\hspace{0.5cm}

Animals are multimodal in that they can simultaneously cope with multiple activities while handling multiple stimuli\textemdash controlling movement and analyzing vision, sound, taste, and smell. For many species, this activity is mundane and happens continuously without interruption from the day they are born to their final day. For the rest of the animal kingdom, it is more modal. \par

If intelligence were purely about brain size, then we would be ranked 4\textsuperscript{th} on our planet, see Figure \ref{figure:brainspec}. Whales, elephants, and dolphins would be in 1\textsuperscript{st}, 2\textsuperscript{nd}, and 3\textsuperscript{rd} place. Archeological evidence shows that our brains were significantly larger around 15,000 years ago \cite{BrainShrinkage,10.1159/000528710}, so our brains have shrunk during that time. One potential reason for this change is associated with society-based problem solving, i.e., moving from isolated individuals to communicating individuals. We do not require the same brain size to solve each problem. In other words, as humans, we have become a distributed system and solve problems in groups. See \hyperref[sec:systems]{Section \ref{sec:systems}}, titled \textit{Physical system design}, for more information on distributed intelligence. We could also have chosen a different metric, such as \textit{neural density}, in which case birds would be first and primates second \cite{kandel2013principles}. \par 

\begin{figure}[hbt!]
\begin{center}
\begin{tabular}{ |l|l| } 
\hline
\textbf{Brain Feature} & \textbf{Measurement} \\
\hline
Type & Massively parallel \\
\hline
Volume & 1400 $ {cm^{3}}$ (85 ${in^{3}}$)\\
\hline
Neurons & 86 Bn \cite{Herculano-Houzel2009-gz}\\
\hline
Synapses (network) & $>$ 100 Tn \\ 
\hline
Average adult weight & 1.5 kg (3.3 lb) \\ 
\hline
Processing capability & 1 exaFLOP \\ 
\hline
MIPS Performance & 100 Mn \\
\hline
Power Requirement & 15-20 Watts \\ 
\hline
Brain cell & 0.07 volts at 1 nanoamp \\
\hline
\end{tabular}
\end{center}
\caption{Human Brain Specification}
\label{figure:brainspec}
\end{figure}

\textsc{\textbf{Social-based problem solving:}} The first potential language, known as the \textit{proto language}, is believed to be the parent from which all languages are derived \cite{spinney2025proto}. It is speculated that this language appeared about 12,000 to 18,000 years ago in East Africa. The advancement of sophisticated language has contributed significantly to the transition toward solving social issues. We use language to coordinate complicated spatial-temporal landscapes: problem domains, creativity, courtship, causality, and abstraction. Language is a significant differentiator in social interactions between animals. This is also why the latest advances in machine learning have the potential to be so important. We can incorporate machine learning into our society because the traditional barriers of inter-species communication are removed\textemdash natural language is the common interface. Language has such a strong influence on our thoughts and experiences, i.e., \textit{Sapir-Whorf} hypothesis.\footnote{Is the idea that language determines thought or at the very least influences thought \cite{green2023possibility}.} \\

\textsc{\textbf{Plasticity:}} When a brain is damaged or learns a new skill, it adapts using \textit{neuroplasticity}. Neuroplasticity allows for modification, repair, and adaptation. The network of neurons changes, meaning that human brains do not remain fixed. Current research focuses on the frequency of adaptation. Artificially induced neuroplasticity can occur with carefully planned electrical stimulation at specific locations in the brain. When applied to stroke patients, the hope is to allow them to regain some of their lost dexterity. \\

\textsc{\textbf{Computational source:}} \textit{What type of computational system is the brain?} There has been much speculation among philosophers, neuroscientists, and computer scientists \cite{natureDecadeslongConsciousness}. Is the brain a conventional classical machine (computational functionalism), or is there something different? The \textit{University of Salzburg} concluded that there is insufficient energy to justify our capabilities \cite{e20080558,TIMP2022}. They speculate that the brain could be some form of a classical/quantum hybrid computer. This hybrid would explain why it is so difficult to pin down and why our brains are so energy efficient. We will return to quantum mechanics a few more times throughout this paper. \\

\textsc{\textbf{Automation:}} It is worth mentioning that many systems in the human body are automatic and rely on reflexes, e.g., touching something scorching (which produces an electrical signal) or something freezing (which produces a chemical signal). It is worth mentioning that the reaction speeds for touching something cold are significantly slower than when we touch something hot. Many of these reactive systems are not under the mental control of the host. The heart is one of those systems that are isolated and self-contained. \\

\textsc{\textbf{Organization:}} This discussion brings us to a broader question: \textit{How is human intelligence organized?} Because the answer is complex, we will approach it through three simplifying lenses: functional components, human physical system design, and knowledge representation. These filters are not mutually exclusive; there are probably many more, but we will limit them to three. We decided to separate the concept of consciousness as a unique and distinct subject; see Section \ref{sec:consciousness}. \\

\textsc{\textbf{1. Functional components:}} The human brain is divided into two primary functional components for decision making: the \textit{hippocampus} and the \textit{prefrontal lobe}. Both parts perform different operational tasks. The hippocampus stores memories, processes object recognition, and even makes simple decisions. In comparison, the prefrontal lobe is much more complicated. It can assess complex problems and perform advanced decision making. The main difference between the two functions is that the hippocampus cannot determine the ramifications of a decision. This limitation means that the hippocampus can easily include irrational prejudices and biases without understanding the broader implications. \par

Early machine learning systems exhibited this characteristic, producing some notably embarrassing results. Today, engineers mitigate this issue by carefully selecting the training data used to create the models to reduce bias in the system. For humans, the prefrontal lobe acts as an override for the hippocampus. The override is achieved by understanding the implications of a decision and stopping anything flagged as having an unpleasant outcome. The prefrontal lobe achieves this by abstracting the ideas and understanding the implications. \par

For machine learning systems, the prefrontal lobe function is one of the desired capabilities. Today, we rely on hippocampus-style processes with carefully selected training data, but tomorrow we want a higher-level capability that understands and perceives consequences.\\

\textsc{\textbf{2. Human physical system design}} Whether they control a bee or a human, brains exploit massive parallel processing to achieve intelligence based on a complicated network of biological neurons. These neurons and the associated structures differ in complexity between animals. A bee neuron is distinctly different from a human neuron \cite{bee2013,physBrainsChallenge}. In some respects, bees are more advanced since they can fly miles, store information, and interact with the world with a brain significantly smaller than our own \cite{physBrainsChallenge}. There are trade-offs with neuron complexity, scaling, and efficiency. \par

The human brain is divided into two physical hemispheres: the left and the right. Each hemisphere has distinct characteristics that give humans uniqueness, such as being left-handed or right-handed. There are famous controversies, such as the right hemisphere being artistic, creative, intuitive, and spatially aware \cite{TheTruthAboutLeftRight}. In comparison, the left hemisphere is verbal, analytical, linguistic, and linear \cite{TheTruthAboutLeftRight}. These ideas may be an inaccurate generalization, but we can say that the right hemisphere controls left-hand motor skills, and conversely, the left hemisphere controls the right-hand side and verbal responses. These physical roles, like handedness, can be swapped for various medical reasons, resulting in partial loss of dexterity, e.g., a left-handed person forced to write with their right hand. \par

\textit{Why do the left and right hemispheres matter to human intelligence?} It matters because they are two codependent intelligent systems \cite{koch2019feeling}. In \textit{Homo Deus: A Brief History of Tomorrow} \cite{harari}, Harari highlighted an important experiment carried out in the 1960s. People with life-threatening epilepsy had the option of hemisphere separation, which cuts the physical communication channels between the left and right sides of the brain. These unique individuals allowed researchers to gain valuable data. One experiment involved a patient and a simple question: \say{\textit{What do you want to be?}}. First, because the question is verbal, the left hemisphere replied, \say{\textit{a librarian}}. The researchers placed a pen in the patient's left hand and asked the same question as in a written note. The right hemisphere responded, saying, \say{\textit{a racing car driver}}. \par

The different answers reinforce the notion that the human brain is, in fact, not one brain, but two brains closely connected together. Each hemisphere has its own desires and assigned control tasks. These observations make a human more than a single intelligent system; they represent a combination of two distinctly different interconnected systems that require arbitration.\par

In addition to the left and right hemispheres, there are three critical physical components: namely, the \textit{cerebrum}, \textit{cerebellum}, and \textit{brain stem} \cite{Anatomy}. The most significant part of the brain is the cerebrum, which spans across the hemispheres; it handles an array of functions, including touch, vision, speech, reasoning, emotions, learning, hearing, and precise motor control. The cerebellum is much smaller and coordinates muscle movement and balance. Finally, the brain stem connects to the cerebrum, cerebellum, and spinal cord. It handles automatic reflex systems, from breathing to swallowing. \par

Our advantage in the animal kingdom could simply come from a more mundane physical property. There is a hypothesis that our unique advantage is related to connectivity (plus verbal skills). If you study any picture of the human brain, you will see grooves and folds; the valley features are called \textit{sulci}, and the peaks are called \textit{gyri} \cite{kkinene0726242025,sciencealertYourBrain}. Our brains have both large and small sulci. Humans are believed to have more sulcus compared to other animals, increasing our complexity, which could explain some of our abilities and differences. \textit{Kevin Weiner}, neuroscientist, stated, \say{\textit{The hypothesis is that the formation of sulci leads to shortened distances between connected brain regions, which could lead to increased neural efficiency and, in turn, individual differences in improved cognition with translational applications}} \cite{sciencealertYourBrain}. This has some similarities to observations in digital computers, in that the closer the computation is to the data and the higher the connectivity, the more efficient the system (a good reason for us to abandon the planar semiconductor designs and move to 3 dimensions). In addition, current digital systems move a lot of useless and uninteresting data, whereas the human brain attempts to optimize data movement with spiking neurons. \par

If sulci and gyri determine the level of intelligence, then in theory, we could use this knowledge to measure individual intelligence; for a more detailed discussion of measurement, see Section \ref{sec:measuring}. \\

\textsc{\textbf{3. Knowledge representation:}} \textit{Stuart Russell}, a computer scientist at the University of California, Berkeley,  wrote \say{\textit{that intelligence without knowledge is like an engine without fuel}} \cite{russell2019human}. Knowledge representation is at the heart of Artificial Intelligence and all forms of intelligence. Once the right representation is determined, a biological or digital algorithm can easily manipulate the system to create the desired result. \par

We pick one idea on human knowledge representation, which, if true, has significant implications for intelligence. A 2017 research paper titled \textit{Cliques of Neurons Bound into Cavities Provide a Missing Link between Structure and Function} conjectured that we might think in up to 11 dimensions \cite{Neuron}. They used a technique called \textit{Algebraic Topology} on neurons. The biological neurons appear to form groups called \textit{cliques} and spaces called \textit{cavities}. These elements, in turn, create high-dimensional geometric objects. Biological neurons differ significantly from artificial neurons in that each neuron has extreme connectivity, i.e., connections. An average human brain contains about 86 billion neurons, with more than 100 trillion connections, providing thought and consciousness. \par

Using this technique, the researchers observed that ideas appear in the brain as dynamic hills of shifting sand, rising up and then disappearing, with the potential of being up to 11 dimensions. Eleven dimensions are far greater than the four\textemdash three spatial plus one temporal, which we generally consider essential. Even today, if we look at the most complex Deep Learning systems, they operate at around six dimensions. \\

\textsc{\textbf{Humans versus dolphins:}} Humans  may have more folds than dolphins, but dolphins have more \textit{cortical gyrification}. Cortical gyrification is the process of forming folds. Gyrification (the peaks) is the process of creating the folds, while the folds are the resulting structures. Dolphins and whales possess a higher level of cortical gyrification and a larger cortical surface area than humans. \\

\textsc{\textbf{Memory:}} An intricate part of intelligence is memory. Memory is interwoven into how intelligence is structured and stored. Humans possess what is believed to be \textit{episodic memory}\textemdash a type of memory that records experiences and significant events (\textit{neocortex}), rather than capturing all available information. It is not a complete picture but a set of salient, linked details that include time and place. Episodic memory is finite and, over time, is subject to forgetting. A new event may only be stored if some other event can be forgotten. This process of forgetting has important implications for intelligence. \par

We retain significant events and the context of the present moment but forget some of the immediate and long-term details. Episodic memory also plays a role in the construction of imagined futures, as described by the \textit{constructive episodic simulation hypothesis} \cite{ADDIS20071363}. Proposed in 2007 by psychologists \textit{Daniel Schacter} and \textit{Donna Rose Addis}, the hypothesis suggests that our brains use stored memories to simulate and construct potential future experiences \cite{ADDIS20071363}. In this view, memory is not just a record of the past, but a foundation for imagination and foresight. It may potentially give us hallucinations (and related dreams) stemming from poorly constructed or damaged memories. \\

\textsc{\textbf{Human aging:}} \textit{Alexa Mousley} et al. wrote a paper on stages of human brain development, titled \textit{Topological turning points across the human lifespan} \cite{mousley2025topological}. The authors analyzed brain networks of about 4,000 people\textemdash from birth to 90 years of age. They found that the human brain changes in a non-linear manner, with four distinct turning points at approximately 9, 32, 66, and 83 years of age. They set five periods of different stages of intelligence. Each turning point marks a distinct phase of development. Rather than slowly changing over time, our brains undergo abrupt transitional changes \cite{mousley2025topological}. This possibly provides further evidence that intelligence emerged through punctuated equilibrium rather than slow changes. \par

The study saw 0-9 years as a time of reducing broad connectivity and increasing specialization. 9-32 years sees a rapid rise in efficiency and major reorganization for optimization. 32-66 years marked an era of relative stability, shifting slowly to local and modular processing. 66-83 years brought a decline in global integration, and our brains became more compartmentalized. 83+ years marked another period of further decline. Our intelligence grows, peaks, and declines. It is designed to function for a specific length of time, with a phase that is more focused on learning and a phase that is more focused on application.  \\

\textsc{\textbf{Bayesian brain:}} Causality could be a possible strength in humans. For example, our brains may not handle probability and statistics particularly well, but there is plenty of evidence that we think \textit{in a Bayesian} manner and form expectations that are constantly revised with more evidence \cite{spiegelhalter2025art}. That evidence comes from sensory input, such as vision, touch, and sound. These sensory inputs are used to update our perception of the world. We appear to have a mental model of the world around us that is constantly being updated, which could explain our ability to learn, reason, and predict. We can see this update occurring in real-time; when an object is just outside our focus range, our brains speculate as to what that object is. As more information becomes available, our brains correct the object classification as more details are provided. We will discuss Bayesian theory in Section \ref{sec:mathematic} covering \textit{Mathematically Modeling Intelligence}.\\

\textsc{\textbf{What are we?}} \textit{Christof Koch}, a neuroscientist at the \textit{Allen Institute for Brain Science} and others, pointed out that we may be a glorified machine \cite{koch2019feeling,hamming2003art}. Made from biological components, it is, in essence, a device incapable of understanding its \textit{own} programming \cite{koch2019feeling}. From this worldview, the mind is a program that runs on a \textit{wet} computer. This wet computer is unconventional compared to digital computers. It does not follow the standard \textit{von Neumann} irreversible design rules. The wet computer also does not have a system-wide clock or communication bus. It is built out of low-speed millisecond switches, combining compute with memory, and it uses mixed analog and digital signals. All of these create a very different kind of machine. He also notes that, unlike the artificial equivalent, the human brain hardly uses any feedforward for image processing. \par

The biological vision allows humans to learn from a single example, whereas today's Deep Learning systems require many samples to create a consistent visual model. The human vision system is more sophisticated and efficient than its artificial cousin.\par

Of course, biological systems follow billions of years of evolution; that is, the endless repetition of natural selection, crossover, and mutation. \\

\textsc{\textbf{Neuron view:}} As mentioned previously, the biological system has many different hardware devices. For vision, there are about 100 different types of biological neurons. Koch notes that the cortical network is fed backward \cite{koch2019feeling}. Approximately 1 in 10 neurons has connections from a previous stage. Most obtain information from nearby neurons or neurons at a higher abstract level, i.e., feedbackward. The theorists do not know precisely why this is the case, but it allows the brain to learn from a single example. Today, the most successful machine learning systems rely solely on feedforward. Each layer passes filtered data to the next. We do not find feed-forward mechanisms in nature. \par

A neuron can cope with different data types; for example, analog is handled by membrane voltage, binary is handled by action potentials (spiking), and time series is handled by spike-timing. There are also ways for a neuron to handle enumeration and mutable weights. We will dive more into the subject of neurons in Section \ref{sec:neurons}. \\

\textsc{\textbf{Future:}} For future machine learning, \textit{Recurrent Neural Networks} and \textit{Neuromorphic Computing} show promise in feedback connections. These technologies, particularly the new hardware appearing in the Neuromorphic Computing space, are beginning to excite neuroscientists as possible solutions for future artificial systems. In addition, we also mention the more general concept of moving artificial neural networks closer to biological neurons in both functional and system-level connections. \\

\subsection*{\textbf{10-Thoughts}}

\begin{itemize}[leftmargin=*]

\item We have identified at least 50 different functional areas in the human brain. These functions are known as \textit{Brodmann's} areas, after the German neurologist \textit{Korbinian Brodmann}. Each provides a specific purpose, culminating in the ability to think, communicate, and control our bodies. 

\item Animals, including humans, are multimodal. Multimodal allows us to handle multiple tasks using the same intelligence unit, that is, the brain.

\item Neuroplasticity allows the brain to reroute critical functions in response to injury or changing needs. In fact, many brain functions have been identified by studying individuals who suffered damage to specific regions of the brain.

\item We have left \& right hemispheres, prefrontal lobe \& hippocampus, and, lastly, the cerebrum, cerebellum, \& the brain-stem.

\item We are probably a wet computer with episodic memory; that is, we remember experiences, significant events, and a time and place stamp. We also forget as a result of our finite storage.

\item We learn while awake and continue processing during sleep.

\item The human advantage may come down to communication. The ability to share and exchange complex ideas and thoughts is a critical trick of human intelligence. Other animals may have larger brains, but articulation is challenging.

\item The ability to forget or compress past experiences into simpler representations is an aspect of sentience.

\item We have neuroplasticity, which can reshape how we think while using a Large Language Model, and Epigenetics may influence us over the longer term, so intelligence may never fully converge. Intelligence is not separate from us\textemdash we remain in the center.

\item  Humans undergo a long training process from childhood to adulthood, which is one of the features that makes our intelligence distinctive. We are slow. Learning takes years, with a lot of knowledge built gradually (and repeated) over time. Here, our best learning method is what some have called a \textit{productive struggle}.

\end{itemize}

\subsection*{\textbf{Recommended reading}}

\begin{itemize}[leftmargin=*]
    \item \textit{The Feeling Of Life Itself}, by Christof Koch 
    \item \textit{Homo  Deus:  A  Brief  History  of  Tomorrow}, by Yuval Noah Harari
    \item \textit{Human Compatible}, by Stuart Russell
    \item \textit{Novacene—The  Coming  Age  of  Hyperintelligence}, by J. Lovelock
    \item \textit{Deep Thinkers}, by Janet Mann (Editor) 
\end{itemize}

\section{Consciousness}
\label{sec:consciousness}

\vspace{1.0em}

\begin{adjustwidth}{0.3cm}{}
\textbf{>> If we can explore intelligence, then the question naturally follows: can consciousness be explored as well? Here we examine the complexity of consciousness and introduces some of the ideas surrounding it.}
\end{adjustwidth}

\hspace{0.5cm}

\begin{quote} 
\centering 
\vspace{1ex}
\textit{\say{To be a thing that knows it's a thing, and furthermore, knows that it knows it's a thing}}\\
\textbf{Robert Pantano, 2022} \cite{pantano2022art}
\end{quote}

\hspace{0.5cm}

There are many strongly held opinions and beliefs about consciousness, but most lack fundamental truths or core facts \cite{butlin2023consciousnessartificialintelligenceinsights}. The subject has seen many changes, from us humans being the center of consciousness (a geocentric view) to the realization that consciousness can appear in other animals (such as octopuses and birds) or even organisms such as plants \cite{pollan2026world}. It is a subject that generates a lot of speculation. And maybe we will never know for sure what consciousness is or how it can be replicated\textemdash we end up with a less than satisfactory view: \say{\textit{we only know it when we see it}}. Consciousness is the ability to experience existence, awareness of being, and understanding of reality \cite{green2023possibility}. Consciousness is the least understood and possibly the most intriguing aspect of intelligence. We can determine a level of intelligence, but consciousness is much more difficult. Consciousness is generally considered necessary for sentience because sentience requires the capacity to have felt subjective experiences. Due to the complexities, philosophers and scientists have constantly argued about its definition \cite{natureDecadeslongConsciousness}. \par

In this section, we will attempt to explain the important ideas. We start by determining how consciousness may differ from intelligence. One of the many complex questions is whether we can have unconscious intelligence. \textit{Can we separate true intelligence from consciousness?}. This question is essential for Artificial Intelligence and humans. Harari put the importance into perspective and stated that \say{\textit{humans are in danger of losing their value because intelligence is decoupling from consciousness}} \cite{harari}. \par

\begin{figure*}[htp]
\begin{center}
\includegraphics[width=0.85\linewidth]{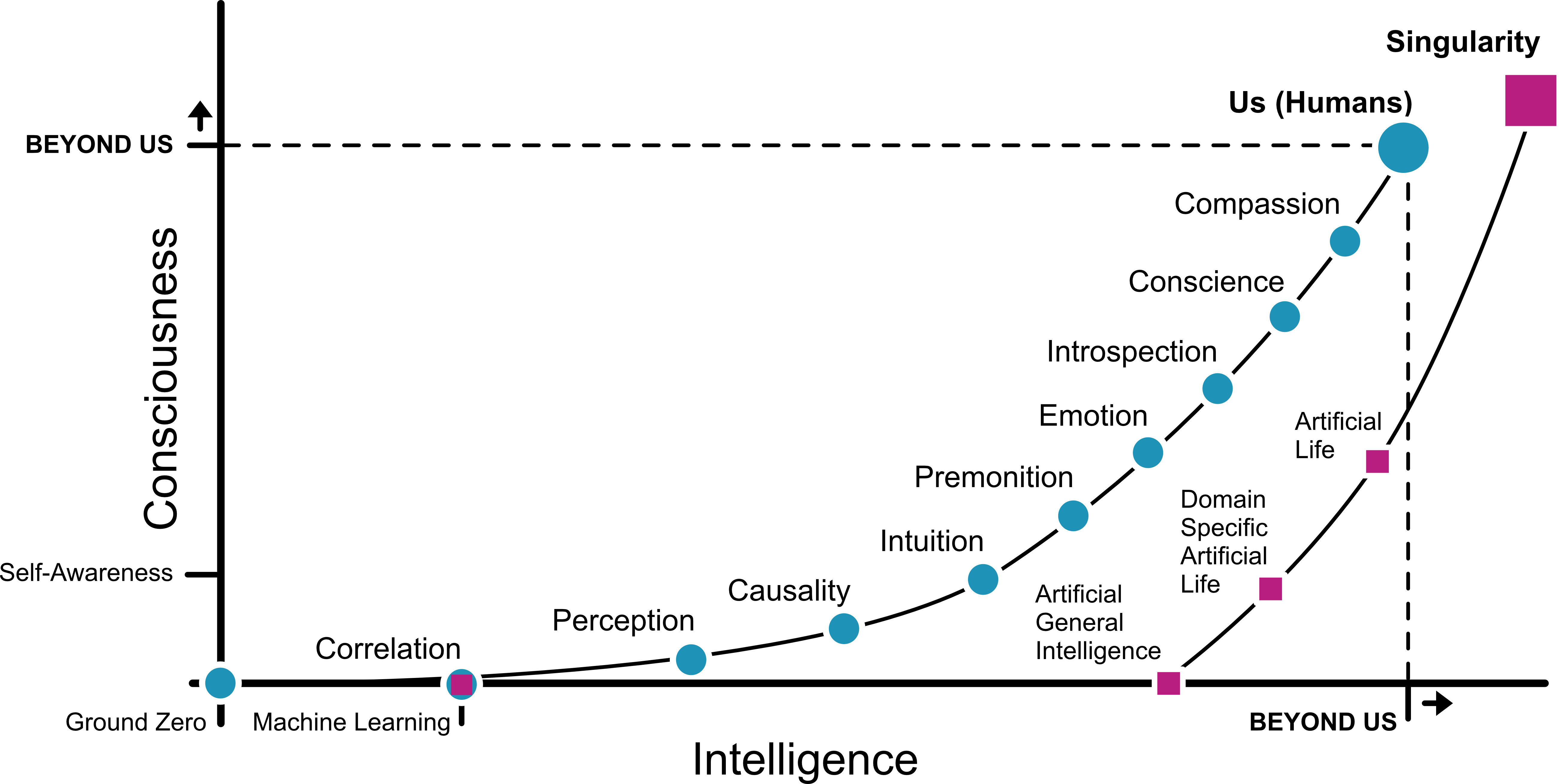}
\caption{A wrong diagram on consciousness}
\label{figure:consciousness}
\end{center}
\end{figure*}

\textit{What is the relationship between consciousness and intelligence?} Inspired by Christof Koch, Figure \ref{figure:consciousness} shows an attempt to visualize the relationship between intelligence and consciousness \cite{koch2019feeling}. The diagram shows the relationship between the two concepts, with intelligence along the x axis and consciousness represented by the y axis. The points represent levels of consciousness and intelligence that are best known. A lot of \textit{poetic license} was applied to show the progression required to achieve human intelligence, symbolized by the label \textit{Us}. This poetic license is why we, the authors, call this our wrong diagram on consciousness, since it is \textit{our} interpretation; e.g., it would be hard to prove whether compassion truly requires more consciousness than a premonition. All current Artificial Intelligence systems remain grounded on the x-axis. All animals, including humans, are on the x- and y-axes. The space beyond human intelligence is labeled \textit{Singularity}, see \hyperref[sec:exceeding]{Section \ref{sec:exceeding}} for more details. The figure shows the large technology gap between current artificial systems and humans. We hope it also highlights the importance and dangers of consciousness compared to intelligence. \\

\textsc{\textbf{Dualism:}} Traditionally, when discussing consciousness, philosophers have addressed the \textit{mind-body problem}. The mind-body problem defines the body as an objective physical system and the mind as a subjective system. This belief is called \textit{dualism}. The question revolves around how the mind and the body connect. This belief ultimately makes intelligence a combination of the physical and the mind (that is, \textit{embodiment}) and is inseparable \cite{Dennett1978-DENWAI}. These connections are why \textit{robots} (also known as \textit{Physical AI}) hold so much promise with the ability to emulate mind and body, providing a potential path to consciousness. \\
 
\textsc{\textbf{Example:}} \textit{Why is consciousness necessary?} Say we have two devices called A and B. Externally, they exhibit the same characteristics. They take the input data, process it, and finally output a result. These devices function the same way. Device \textit{A} is functionally correct, while device \textit{B} is self-aware. The difference is \textit{B} understands the data: it understands the meaning, cause, and implications for a particular conclusion. It is aware of its existence; that is, it is not mindless but conscious. Consciousness requires a measurement method. \textit{"A"} is a \textit{philosophical zombie}, in that it exhibits all the behavioral aspects of intelligence but lacks understanding. \\

\textsc{\textbf{Searle's Chinese room:}} To emphasize the differences between consciousness and intelligence, \textit{John Searle} created the notion of the \textit{Chinese room} \cite{ChineseRoomArgument,sep-chinese-room}. It poses a thought experiment as a counter-argument to the famous \textit{Turing test}  \cite{sep-turing-test}. The investigation requires you to imagine that you are in a room and that you receive three batches of information: a set of symbols ('a script'), 'a story', and 'questions.' You push a bunch of symbols out of the box that answer the questions. Also provided is a set of written rules in English, known as 'the program.' You do not understand Chinese. Chinese characters enter the room. You apply the rules in an unmindful way. The answers are returned in perfect Chinese. The people outside the room can only assume that the person inside speaks fluent Mandarin, but the person inside knows nothing about the language. \textit{Peter Watts}, a science fiction author, points out that even though the person inside the room does not understand Chinese, the system as a whole does because it provides symbols and pattern matching \cite{watts2020blindsight}.\\
 
\textsc{\textbf{Known:}} Professor Paul Li, a Cognitive Scientist from U.C. Berkeley, in a talk at Stanford University, asked the question: \say{\textit{what do we know for sure about consciousness?}}. Computational representation understanding of mind (CRUM) shows that it is most likely biological/neurological, electrical, and chemical. That is all we know, and it is not much.\\

\textsc{\textbf{Other minds:}} Finally, the concept of \textit{other minds}. If we understand our mind, can we understand other peoples? We can access our minds, but can we rationalize anything about anyone else's mind? \cite{scientificamericanConsciousnessPart}. These are important questions. Can we rationalize consciousness? \\

\noindent
Below is a subset of ideas about consciousness:\\ 

\begin{enumerate}[label={C-\arabic*:},leftmargin=*]

\item \textsc{\textbf{Computational functionalism basis:}} Consciousness occurs as a result of functional elements. By learning more about the brain, we will gain a better understanding. In this concept, consciousness is viewed simply as a byproduct of functional execution. The \textit{Global Workspace Theory} (GWT) \cite{GWT,doi:10.1073/pnas.2115934119} is one concept that falls under the functional category. This theory is closely related to an old idea in Artificial Intelligence. Global Workspace Theory has a centralized blackboard (workspace) where ideas live. These living ideas are placed or taken from the blackboard; some ideas appear briefly. Ideas can be combined, processed, or ignored. Subsystems are available to handle low-level ideas. Consciousness, therefore, comes about through functional processes. \par

The functional approach also includes the notion that animals are physical \say{wet} computers, processing complex data. The main task today is to understand how these complicated biological systems interact and function as a complex system.\par

High-level consciousness is a functional system, so the more we understand the functions, the closer we get to having the ability to understand and, in theory, create machines with consciousness. \\

\item \textsc{\textbf{Universe basis:}} Galileo believed certain things have repeatable physical characteristics that obey mathematical laws \cite{scientificamericanConsciousnessPart}, e.g., a ball rolling down a hill. Other concepts reside only in the conscious, e.g., smell, taste, and color. These qualities only exist in the mind. If people cease to exist, then these qualities evaporate. Meaning our physical laws are incapable of providing the complete story. \textit{Philip Goff} put forward the concept of Panpsychism \cite{goff2019galileo}. \textit{Panpsychism} proposes that consciousness is a fundamental aspect of reality \cite{scientificamericanConsciousnessPart}. Meaning it is equivalent to mass or charge. The belief is that consciousness is inherent in the fabric of the universe and not limited to a brain. This belief comes about if we separate the substrate and the concept. Christof Koch stated that if true, the \say{\textit{cosmos is suffused with sentience}} \cite{koch2012consciousness}. This idea leads to \textit{cosmopsychism} that the universe itself is conscious. \par

Panpsychism theory has difficulty explaining combination, i.e., how small consciousness combines to create a more significant form \cite{scientificamericanConsciousnessPart}. In other ideas such as \textit{Psychological ether theory}, brains do not produce consciousness but read consciousness from the universe, i.e., consciousness existed before our brains existed. \\

\item \textsc{\textbf{Biological basis:}} \textit{Biopsychism} is the view that all living things possess some form of consciousness. Here, life is defined as ranging from viruses and simple organisms to philosophers. The physicist Karl Friston calls this the evolutionary staircase of inference, adaptation, and self-maintenance  \cite{friston2012freeEnergyBiologicalSystems,pollan2026world}. \\

\item \textsc{\textbf{Experiences basis:}} \textit{Christof Koch} defines consciousness as a set of experiences. The historical experiences differentiate humans from each other and machines \cite{koch2019feeling}. Some humans are more disposed to experiences than others. Experiences are a form of causal action. Koch showed an exciting model, already introduced in Figure \ref{figure:consciousness}, to show the difference between consciousness and intelligence. In this figure, we have taken the liberty to show how we humans and Artificial Intelligence could map out. The driver for this figure is to provide two elements, the first is an abstract view of where we are in terms of intelligence, and the second is to highlight the importance of consciousness. Koch describes an experience in terms of the Integrated Information Theory (IIT) \cite{IIT}:-

\hspace{0.5cm}

\begin{quote} 
\textit{\say{consciousness is determined by the causal properties of any physical system acting upon itself. That is, consciousness is a fundamental property of any mechanism that has cause-effect power upon itself.}}\\
\textbf{Christof Koch} \cite{koch2019feeling,IIT}
\end{quote}

\hspace{0.5cm}

Integrated Information Theory can be said to be a form of Panpsychism. Koch and others proposed that we may never replicate biological consciousness if the replication method is digital simulation. Consciousness requires causal powers to make consciousness conscious. For engineers and scientists, this is probably best illustrated as a simple equation; see Figure \ref{figure:causal}. Using the equation, we would need both simulated rules and the ability to store and create causal effects to make a conscious machine.\\ 

\begin{figure}[h]
\begin{center}
\includegraphics[width=.6\linewidth]{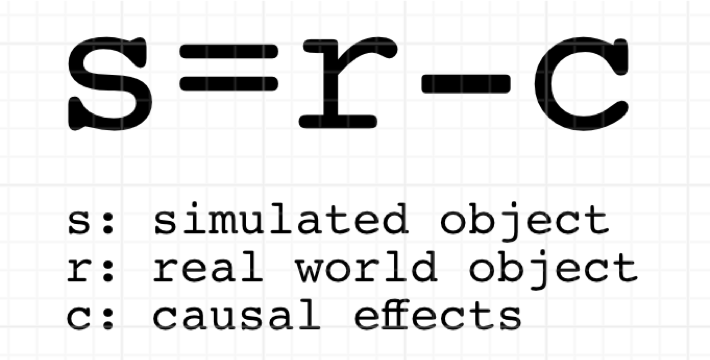}
\end{center}
\caption{Causal Effects Equation}
\label{figure:causal}
\end{figure}

\item \textsc{\textbf{Radical plasticity basis:}} It was reported in the medical journal \textit{The Lancet} in 2007 that a man had come in for medical treatment complaining of weakness in his left leg, but otherwise in good health. It was discovered the man was missing nearly 90\% of his brain \cite{qzCivilServant}. Most of his brain had been been replaced with fluid, caused by a condition known as \textit{hydrocephalus}, which started when he was an infant. The man was married with two children, and had a job as a civil servant. This discovery destroyed the notion that consciousness might live in a specific region in the brain. \textit{Axel Cleeremans}, a cognitive psychologist, came up with an hypothesis called \textit{Radical Plasticity Thesis} that puts forward the idea that the brain learns consciousness continuously. In other words, we are not born with consciousness. Consciousness can be flexible allowing it to reside in different regions of the brain \cite{sciencealertMeetLives}.\\ 

\begin{quote} 
\textit{\say{Cleeremans argues that in order to be aware, it's necessary not simply to know information, but to know that one knows information. In other words, unlike a thermostat that simply records temperature, conscious humans both know and care that they know $\cdots$Cleeremans claims that the brain is continually and unconsciously learning to re-describe its own activity to itself, and these descriptions form the basis of conscious experience.}}\\
\textbf{Olivia Goldhill summarizing Cleeremans thesis \cite{qzCivilServant}}\\
\end{quote}

\item \textsc{\textbf{Quantum basis:}} 
As an alternative thought, \textit{Roger Penrose}, a   mathematical physicist, believes that \say{\textit{whatever consciousness is, it's not computational}}. Penrose proposed the concept \textit{Orchestral Objective Reduction} (OrchOR). OrchOR \say{\textit{is a biological philosophy of mind that postulates the consciousness originates at the quantum level inside neurons, rather than the conventional view that it is a product of connections between neurons}} \cite{EmperorNewMind,Penrose}. Penrose believes microtubules, particularly symmetric ones, might be more quantum-driven. There is a chance these structures preserve the quantum state. Penrose believes consciousness surpasses the simple ability to compute neurons and the associated synapses. In other words, consciousness occurs at the quantum level and thus beyond the standard equations. Photosynthesis is just one of the proof points that quantum mechanics is involved in biology.\par

Penrose asks several questions. What turns consciousness off? How is consciousness constructed? When we anesthetize someone, how do the chemicals in the gas subdue or even turn off consciousness? These remain open questions. Is consciousness made with proto-consciousness elements? These proto-consciousness elements orchestrated as a group could form the entity's consciousness.\par 

These are theoretical questions that require substantial experimentation and proof. The mind involves understanding, intelligence, and awareness. As Penrose put it, intelligence needs interpretation, and interpretation requires awareness. And as of today, these terms have yet to be formally defined.\\

\item \textsc{\textbf{ALARM basis:}} 
The alarm theory of consciousness, proposed by \textit{Albert Newen} and \textit{Carlos Montemayor}, argues that the primary function of consciousness is warning and prioritization \cite{Newen2023-NEWTAT-8}. In this view, consciousness is an ancient and primitive survival system. It includes a concept called \textit{reflexive consciousness}, which supports both immediate responses and longer-term adaptation. And it is pain that causes adaptation and growth in animals \cite{pollan2026world}.

\end{enumerate}

\textsc{\textbf{Reflective self-awareness:}} As mentioned previously, many subareas of consciousness exist, including \textit{reflective self-awareness}. Reflective self-awareness is about recognizing the physical self and the ability to contemplate what the physical is thinking or feeling, that is, \textit{introspective}. This self-awareness extends to \textit{self-evaluation} and a sense of personal \textit{identification}. It allows an entity to assess how it interacts with the outside world. A related subject is \textit{mirror self-recognition}, where only a few animals recognize that they are looking at themselves when using a mirror. \\

\textsc{\textbf{Theory of Mind:}} Following on from Reflective self-awareness, we have \textit{Theory of Mind}. The theory of mind is the ability to understand others. This understanding includes empathy, feelings, perspectives, and thoughts that may differ from our own. This means understanding others and what drives their behavior. This level normally occurs through experiences and learning. This ability is not present in our early development stages; it is only as we mature that this skill materializes. \\

\textsc{\textbf{Artificial Consciousness:}} The last point on consciousness. There are ethical consequences to allowing artificial consciousness. These consequences concern existential, sentient, and cognitive capacities \cite{butlin2023consciousnessartificialintelligenceinsights}. In other words, when is a thing not a thing, but a sentient being with likes and dislikes? The question of sentience is difficult to answer but is essential if we are to go beyond basic intelligence. A necessary use for consciousness is to enable conscience, that is, to determine right from wrong. Does consciousness ultimately lead to emotion? What happens if a system gets angry or even happy? \\

With all these wonderful theories on consciousness, it remains an enigma, and that enigma is \textit{us}.

\subsection*{\textbf{11-Thoughts}}

\begin{itemize}[leftmargin=*]

\item Every year, new theories of consciousness emerge, and they often attract significant attention. From a Platonic perspective, \textit{consciousness is a great employer}. 

\item There are many ideas associated with consciousness. Does consciousness come from increasing complexity? Historical evidence suggests there is a correlation between intelligence and increased complexity. Does consciousness emerge from the same process as intelligence? In other words, as system complexity increases, does consciousness emerge for management purposes? Does intelligence converge into consciousness?

\item Consciousness provides an awareness of time (a fourth dimension).
\textit{James Gleick}, a science writer, wrote: \say{\textit{Our minds are time machines, we handle the present (e.g., walking), we can recall the past, and we can imagine the future}} \cite{gleick2017time}\textemdash \textit{Chronesthesia}.

\item What we call Artificial Intelligence today does not have consciousness. The debate is whether it should be? When is mimicking so good that it is indiscernible from biological consciousness? Is this indiscernibility the core worry with artificial systems?

\item Quantum mechanics may play an important role in consciousness. For example, animals might be able to navigate near the quantum limit of magnetic field detection \cite{physCertainAnimal}. 

\item As intelligence increases, so does the capacity for introspection, and deeper introspection may carry a greater risk of depression. This raises an important question: \textit{could artificial systems also exhibit forms of depression, or even more severe clinical-like states, or could they refuse a task and simply say no?}

\item \textit{Cambridge Declaration of Consciousness} \cite{low2012cambridge} was published in 2012. It was a landmark statement declaring that non-human animals, including mammals, birds, and even octopuses, possess the capacity to experience consciousness and emotions. We were gracious enough to allow other animals to enter our selective consciousness club. \textit{Are plants next} \cite{pollan2026world}?

\item Position can be understood as a form of consciousness: the ability to know where we are, where we need to go, and what we need to avoid \cite{pollan2026world}.

\item There are two broad views of consciousness. One view says consciousness comes from physical processes in the brain and could, in principle, be simulated by a computer \cite{pollan2026world}. The other view says this explanation is incomplete because it does not explain why physical processes create our experiences\textemdash we are missing something fundamental. 

\item We can think of consciousness as being panpsychism, idealism, illusionism, or quantum based, but with enough examples, can we mimic consciousness? Similar to how we \textit{think} we mimic intelligence? And \textit{what about unconsciousness?}

\item In an interview with physicist \textit{Karl Friston}, he stated that \say{\textit{all sorts of non-conscious systems make inferences about the causes of the information they gather ...}} \cite{pollan2026world}. Even viruses can be seen performing a minimal form of inference, altering their behavior in ways that support survival. \textit{Is consciousness just the ability to handle uncertainty and survive?}

\end{itemize}

\subsection*{Recommended reading}

\begin{itemize}[leftmargin=*]
    \item \textit{The Feeling Of Life Itself}, by  Christof Koch
    \item \textit{Galileo's Error}, by Philip Goff
    \item \textit{Homo  Deus:  A  Brief  History  of  Tomorrow}, by Yuval Noah Harari
    \item \textit{Novacene—The  Coming  Age  of  Hyperintelligence}, by J. Lovelock
    \item \textit{Consciousness as a Social Brain}, by Michael S. A. Graziano
    \item \textit{Where am I?}, by Daniel C. Dennett
    \item \textit{A World Appears}, by Michael Pollan
\end{itemize}
\section{Bias, prejudice, and individuality}

\vspace{1.0em}

\begin{adjustwidth}{0.3cm}{}
\textbf{>> Bias, prejudice, and individuality are connected because every individual develops a unique way of seeing the world, but those personal patterns can serve good and bad purposes.}
\end{adjustwidth}

\hspace{0.5cm}

\begin{quote} 
\centering 
\vspace{1ex}
\textit{\say{Fun is our brains way of giving us a reward when we do something challenging}}\\
\textbf{Paraphrasing Norman Doidge, 2007} \cite{doidge2007brain}
\end{quote}

\hspace{0.5cm}

Bias shows favoritism for or against an object compared to another. That object can be something, a person, or a group. Generally speaking, it is a learned mechanism. The human experience occurs through perception. A significant amount of perception filtering is performed to focus on specific tasks. For example, a very good demonstration of this is an easy one to perform on oneself: simultaneously touch your nose and toes with each of your hands. The touch sensation occurs simultaneously, but the neural distance between a human's nose and the brain is a fraction of the distance from their toes. Human biases exist to focus intelligence on specific goals and tasks. This is the way human brains define \textit{normal} and \textit{abnormal}. \par

Although human bias exhibits itself destructively, it does serve a significant function in day-to-day intelligence by determining what is relevant at any given moment. In a way, it is a filter on the human mind that shows what we care about most at that moment in time. Whether it is a walk to work in a city or our past selves gathering food in the forest, bias ignores mundane information and focuses on the novel. In this case, novelty is the task at hand or, more importantly, from a survival standpoint, danger. \par

We gain new knowledge if it is relevant and salient in epistemological terms. In this context, human bias determines what is relevant to the individual, and prior experience determines its relevance. Therefore, humans learn more items only if the information is understandable and satisfies their interests and worldviews. \par

System designers attempt to converge a machine learning model with a known norm, such as in producing solutions that lack inspiration or excitement \textemdash technical, perfect solutions without mistakes. \textit{Iris Murdoch}, a philosopher, pointed out that by allowing editing, a writer will move the text from the excitement of spontaneity to a more passive normative \cite{Murdoch1992-MURMAA}. \par

It is essential to note that in biology, bias defines individuals; there are no individuals without prejudice. \par

Good bias occurs when the system follows a specific path. By comparison, bad biasing is when a system exhibits unintended biases. There are many examples of bad biasing in machine learning. An excellent example is Microsoft's chatbot known as \textit{Tay} \cite{MSFTTAY}. The chatbot used Twitter information and interactions to learn. Unfortunately, the chatbot learned all the wrong things, that is, extreme human prejudice. The primary method to avoid poor bias is to select training data that reduces the possibilities of bad decision making. \par

Two other methods that could help identify and potentially avoid bad bias are \textit{causal modeling}, as outlined by Judea Pearl \cite{Pearl}, and understanding decisions with \textit{explainable Artificial Intelligence} \cite{russell2019human} (see \hyperref[sec:explainable]{Section \ref{sec:control}}).

\subsection*{\textbf{1-Thought}}

\begin{itemize}[leftmargin=*]
\item Biases make us who we are as individuals. The difficulty is to build an artificial system that demonstrates fairness better than we do as individuals. 
\end{itemize}

\subsection*{\textbf{Recommended reading}}

\begin{itemize}[leftmargin=*]
\item \textit{Introduction to Philosophy}, by John Perry et al. 
\end{itemize}

\section{Exceeding human intelligence}
\label{sec:exceeding}

\vspace{1.0em}

\begin{adjustwidth}{0.3cm}{}
\textbf{>> How do we extend intelligence beyond what we already know, not merely by increasing knowledge, but by enhancing intelligence itself? }
\end{adjustwidth}

\hspace{0.5cm}

\begin{quote} 
\centering 
\vspace{1ex}
\textit{\say{The hand of man is now too slow for the demands of his curiosity, but the power of steam comes to his assistance}}\\
\textbf{Charles Babbage, 1832} \cite{babbage1832economy}
\end{quote}
\hspace{0.5cm}

\begin{figure}[ht]
\begin{center}
\includegraphics[width=1.0\linewidth]{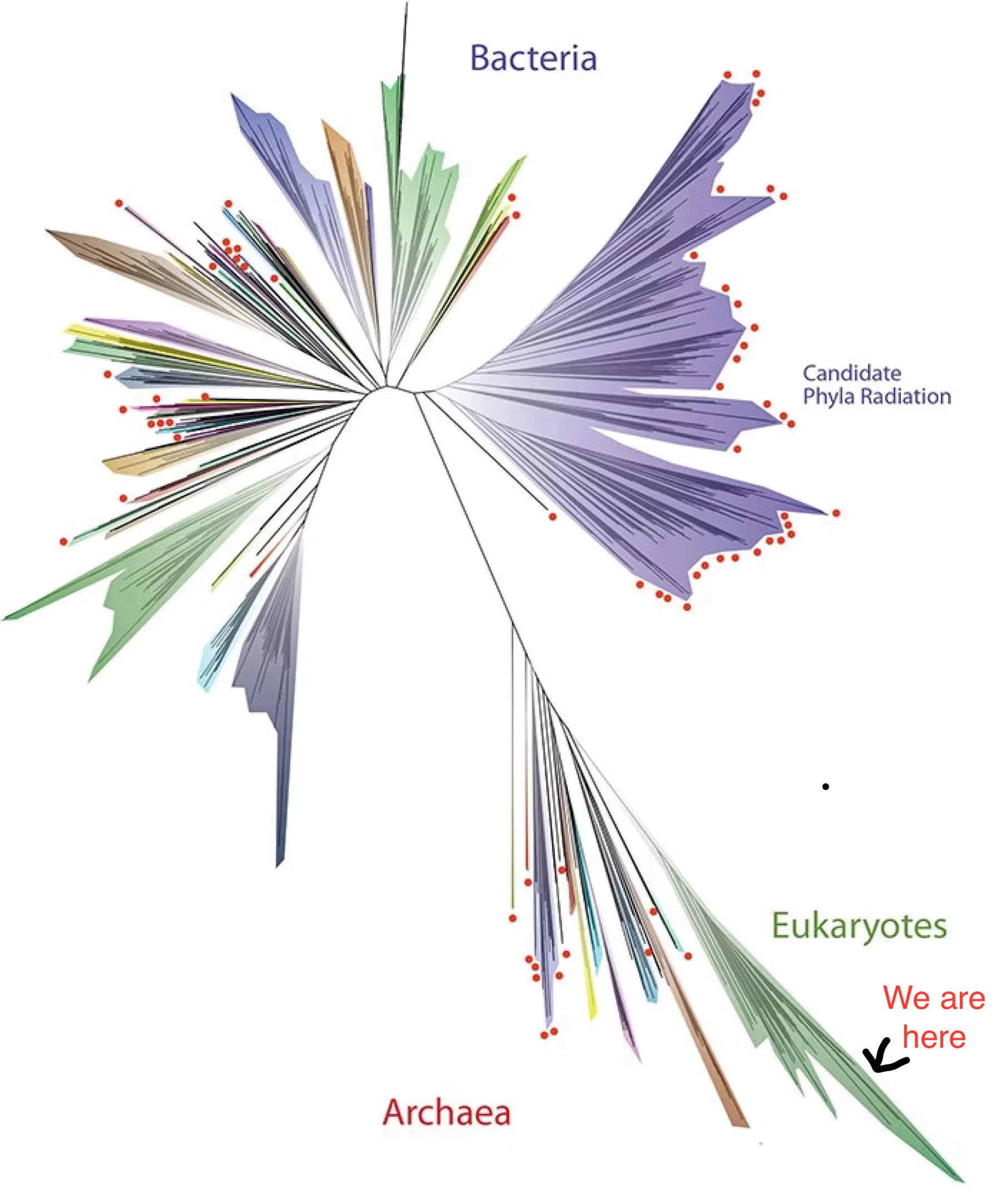}
\includegraphics[width=0.8\linewidth]{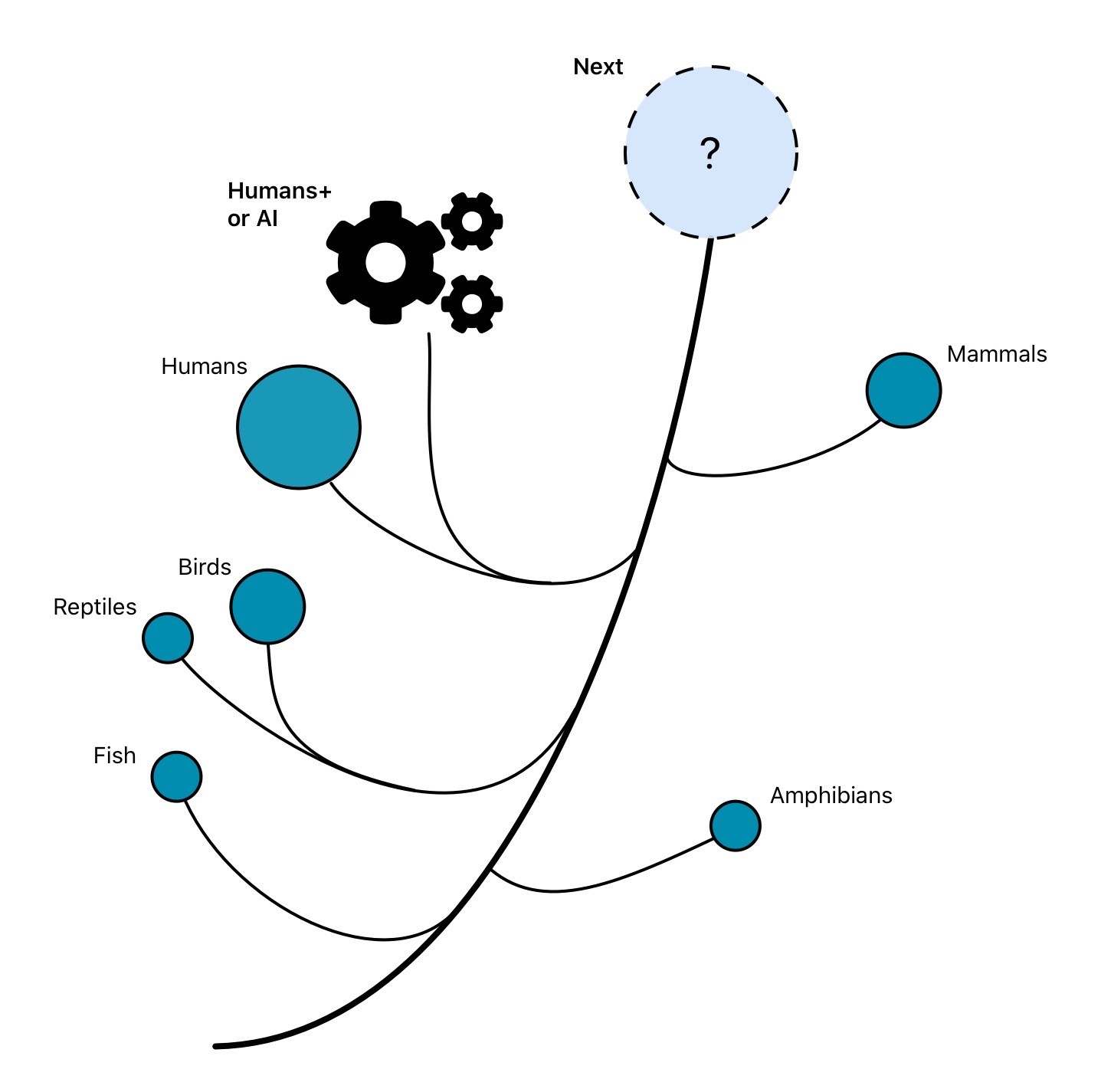}
\end{center}
\caption{[Top] A current view of the tree of life, encompassing the total diversity represented by sequenced genomes \cite{hug2016} [Bottom] Simplified Scala Nuturae "Ladder of Being" (plus some poetic license)}
\label{figure:tree}
\end{figure}

Up to this point, we have been discussing \textit{digital cloning} or \textit{digital mimicking}; in this section, we discuss how to exceed our intelligence. This area comes with both optimism and concern. Surprisingly, it was \textit{John von Neumann} who created the now infamous term \textit{Technological Singularity} \cite{TributetoJohnVonNeuman}. The belief is that once the singularity is obtained, technology will be at an irreversible point; see the lower diagram in Figure \ref{figure:tree}\textemdash Singularity is achieved by constructing an artificial system or via augmentation. Specifically for intelligence, this is known as \textit{superintelligence} and \textit{hyperintelligence}. Superintelligence is about exceeding human intelligence and being better at reasoning, creativity, objective identification, and problem solving. At the same time, hyperintelligence is more associated with augmentation and extensions. We are attempting to artificially extend ourselves to be better. \par

There are at least two other methods to exceed human intelligence, namely \textit{evolution} or an external influence. Evolution does not stop \cite{calcott2011major}; it continues. Humans are just an interlude in the process; assuming that we are an endpoint would be an error. The fourth possibility is an external influence, that is, \textit{extraterrestrials} \cite{green2023possibility}. Extraterrestrial influence is beyond the scope of this paper but raises some interesting questions about how we define life and, more precisely, how we define intelligent life, e.g., \textit{Assembly Theory} \cite{centauridreamsAssemblyTheory}. \par

There are always concerns. The primary problem with having superintelligence is that it possesses superintelligence without consciousness, i.e., no awareness of implications. Harari describes an artificial intelligent system that takes over the world (and beyond), and its only objective is computing $\pi$ \cite{harari}. It constantly seeks to obtain resources and removes all obstacles, without awareness of right or wrong; the system takes over the world by continuously consuming more and more resources to feed a pointless calculation. It has no evil intentions; it is too focused on its goal to consider other factors. It is unaware of itself and the implications of its actions. Humans are secondary in this scenario at best. \par

\textit{If some form of super Artificial Intelligence should emerge, will it occur by accident due to system complexity?} The system would attain sufficient complexity to allow conscious intelligence to emerge. In other words, emergence would not be designed or manufactured, but would form naturally through chaotic processes. This chaotic emergence would mean that intelligence would appear without the expected controls. Controls would have to be learned; see Section \ref{sec:emergence} for more discussion on emergence. \\

\textsc{\textbf{Superintelligence:}} There are at least two primary methods for building a system capable of superintelligence \cite{BOSTROM}. The first method is to create a system so complex in knowledge and sophistication that full conscious intelligence emerges. The other process involves transferring or copying an existing biological intelligence in the hope of jump-starting superintelligence. The jump start consists of reading and copying all neurons and synapses in a brain (a favorite of many science-fiction stories). \par

Koch noted that at the current rate of technological advancement, we should be able to simulate a complete rodent brain \cite{koch2019feeling} fairly soon. The \textit{OpenWorm} project is an open-source project dedicated to creating the first virtual life form (see \url{https://openworm.org}). \textit{Blue Brain} was a project developed to create a digital mouse \cite{Arachchige2023-vw}. \par

The \textit{University of Manchester} developed \textit{SpiNNaker}, a Spiking Neural Network Architecture large enough to simulate a brain. The project involved building a large-scale digital spiking neural network\textemdash a neuromorphic computer. The scale and function of spiking neurons allow for the potential simulation of a brain. SpiNNaker includes over a million processor cores, enabling up to a billion neurons ($10^{9}$) to be simulated in real time. Research continues with the second generation system called \textit{SpiNNaker 2}, which aims to have 10 million processor cores and simulate somewhere between 10 and 100 billion neurons \cite{gonzalez2024spinnaker2largescaleneuromorphiceventbased}. Compared to a human brain, SpiNNaker 1 can simulate about $1.2\%$ of the required neurons, and SpiNNaker 2 can simulate between $11$ and $116\%$. We are about to reach an interesting point where we have platforms capable of simulating all the neurons in a human brain. Noting that there are a number of projects chasing the same goal. \\

\textsc{\textbf{Hyperintelligence:}} Hyperintelligence is a concept in which humans are augmented with technology to enhance their intelligence. \textit{James Lovelock}, the originator of the \textit{Gaia hypothesis}\footnote{Argues that living organisms on the planet, along with inorganic materials, can self-regulate the climate and biochemical environment\textemdash allowing life to continue to exist within a balance.}. He believes that increasing human intelligence is the only solution to global warming and world problems. The only way to improve intelligence is to increase ourselves (through augmentation) \cite{Novacene2019}.\par

In recent years, we have seen a rapid increase in the sophistication of devices that link with the brain. These devices use a \textit{brain-computer interface} (BCI) to read and, in some cases, write to the brain. There are two types: noninvasive and invasive. The noninvasive ones are designed for read-only access. The invasive species can potentially do both. To give some perspective, we can use a brain-computer implant to transform people's thoughts into text or audible words. This achieves a precision of nearly 74\%. This allows people with severe paralysis to communicate verbally \cite{nield2025brainimplant}. \par

Noninvasive devices are used to measure brain states. There is a long history of trying to read our state of mind; one of the most famous techniques is the \textit{polygraph test}, often called the \textit{lie detector test}. These are used by law enforcement and government agencies, with an accuracy between 80 and 90\%. The new noninvasive devices are much more precise and are based on a form of headset that reads EEG signals coming from the brain. They can sense certain brain states: motor, attention, cognitive processes, and general neurological activity. They are potentially much better at determining whether we are lying. The augmentation is not quite the hyperintelligence goal, but we can potentially alter systems like autonomous vehicles or video games depending on a persons attention level (sleep, fully awake, or very anxious). These devices are continually improving with cheaper and faster methods of reading brain states. \par

Invasive devices are implanted directly into the brain, connecting to specific neurons. We typically apply these devices to people with cognitive disabilities, for example, people who have suffered head injuries or experienced a severe stroke. Neural pathways are redirected by stimulating particular regions of the brain. This can partially correct stroke damage. Using \textit{plasticity}, which teaches the brain to bypass damaged areas by creating new neural pathways. Companies see this as an opportunity to speed up human-computer communication by allowing direct communication, i.e., opening the door for augmentation and hyperintelligence. \par

Augmentation could also remove aspects of our humanity. Augmentation could involve removing the drive to explore or travel. By reducing our respiration, we can control $CO_2$ levels or even change our eating behaviors, making us all vegetarians and reducing methane production. Augmentation could have significant future implications. \\

\textsc{\textbf{Tracking human thinking:}} By tracking humans, we can exceed our ability to predict. The idea is based on using different training data. Today, most of our training is related to books, academic papers, code, and arguments. \textit{Mikhail Gorelkin}, a researcher, proposes the idea of more training information \cite{Gorelkin2026Superintelligence}. The focus is on the discourse about failed and successful forecasts. In wider contexts such as politics, economics, war, and society. The idea here is that if a system could extract the predictions, formally prove them, and track the outcomes. An Artificial Intelligence system could track human thinking and even determine where our models of reality break down. Where failed forecasts are traces of broken world models \cite{Gorelkin2026Superintelligence}. \\

\textsc{\textbf{What happens when we exceed human intelligence?}} If we exceed human intelligence, where does this lead us \cite{green2023possibility}? Do humans end up co-existing with this new intelligence? Do humans become pets and end up in zoos? Do humans ascend to a higher plane of existence (do we become the machine)? Or do humans eventually become extinct (for good or bad reasons)? Do government regulations slow down or speed up this process? Does energy efficiency become the primary constraint and limit our ability to exceed or scale? 

\subsection*{\textbf{6-Thoughts}}

\begin{itemize}[leftmargin=*]
\item Exceeding human intelligence creates both excitement and alarm in people. The excitement comes from doing more and exploring more. The alarm comes from control and purpose. 

\item \textit{Superintelligence} creates intelligence that exceeds human capability. \textit{Hyperintelligence} is augmenting ourselves to surpass standard human capability.

\item \textit{Borgization} is the concept in which we become slaves to the machine. It pushes the narrative that future success is directly linked to augmentation.

\item Biologists use the \textit{tree of life} diagram to show the evolutionary relationships between different species and organisms. Artificial systems have the potential to extend the tree of life in new directions. 

\item As we copy biology, the lessons learned in digital implementation can be redirected back to biology. We reached a point where we can grow synthetic brains in the lab \cite{michiganmedicineArtificiallyGrown}, which come with their own moral and ethical issues.

\item Again, speculation about when Artificial Intelligence will exceed human intelligence has become a lucrative industry. A potentially more important and interesting question is whether Artificial Intelligence can surpass human intelligence without possessing consciousness, conscience, or morality.

\end{itemize}

\subsection*{Recommended reading}

\begin{itemize}[leftmargin=*]
    \item \textit{Superintelligence\textemdash Path, Dangers, Strategies}, by Nick Bostrom
    \item \textit{The Tangled Tree}, by David Quammen
    \item \textit{The Major Transitions  in  Evolution  Revisited}, by B. Calcott Et al.
    \item \textit{Human compatible}, by Stuart Russell 
    \item \textit{Novacene\textemdash The  Coming  Age  of  Hyperintelligence}, by J. Lovelock
    \item \textit{The Feeling Of Life Itself}, by Christof Koch 
    \item \textit{The Possibility of Life}, by Jaime Green
\end{itemize}

%
%

\section{Reasoning}

\vspace{1.0em}

\begin{adjustwidth}{0.3cm}{}
\textbf{>> Reasoning is a foundation of intelligence. At the core are different forms, that include inductive, deductive, and abductive reasoning.}
\end{adjustwidth}

\hspace{0.5cm}

\begin{quote} 
\centering 
\vspace{1ex}
\textit{\say{When you have eliminated the impossible, whatever remains, however improbable, must be the truth.}}\\
\textbf{Arthur Conan Doyle, writer, 1890} \cite{doyle1890sign}
\end{quote}

\hspace{0.5cm}

Logical reasoning can be challenging for some people. What is often called \textit{common sense} is probably more accurately described as social knowledge. Here, we refer to the ability to conclude from external evidence. Reasoning is the act of going through a problem or situation systematically. As an introduction method, we will define logical inference as the achievement accomplished using one of three reasoning methods: namely, \textit{deduction}, \textit{induction}, and \textit{abduction}. Deductive reasoning involves making a statement or statements and inferring a conclusion\textemdash that is, \textit{from general to specific}. Inductive and abductive reasoning take the opposite approach and attempt to generalize from an instance—that is, \textit{from} specific to general. In other words, the latter methods incorporate uncertainty or probability. Abductive reasoning differs in that it includes \textit{cause-and-effect}. An important note; all three methods require the capacity for abstract thought\textemdash reasoning beyond what is apparent in the immediate physical environment and often involving multiple states.\\

\noindent
Examples of the three reasoning methods:\par

\begin{enumerate}[label={R-\arabic*:},leftmargin=*]

    \item \textsc{\textbf{Deductive reasoning:}} \say{all adults were once children}, \say{Jenny is still alive and no longer a child}, so we can deduce \say{Jenny is an adult}.\\
    
    \item \textsc{\textbf{Inductive reasoning:}} \say{the first person I met this morning was happy}, so \say{everyone is happy this morning}.\\
    
    \item \textsc{\textbf{Abductive reasoning:}} \say{a soccer ball is flying towards us}, so \say{a football player on the opposite team must have kicked the ball}.\\
\end{enumerate}

Reasoning is fundamental to intelligence. The masters of reason are considered the most intelligent in our society. Deductive reasoning is more mechanical, whereas inductive and abductive reasoning require some form of inspirational jump, i.e., calculated uncertainty.\par 
 
Within Computer Science, rule-based Expert Systems and languages such as Prolog have taken direct advantage of deductive reasoning \textemdash reasoning forward or backward. The example of deductive reasoning given above is a forward deduction example, but if we say \textit{\say{Jenny is an adult}}, we have the rules to backward-deduce that she was once a child. Machine learning grapples with inductive and abductive reasoning. Both require some form of probability and potential randomness to arrive at a good enough or general solution.\par

\subsection*{Other attributes}

\textsc{\textbf{Observed/Non-observed:}} Reasoning is entirely different when the problem is either fully or partially observable. An example of a situation that is entirely observable is the game of chess, where all future moves can be calculated in advance using some form of search mechanism, for example, an \textit{ iterative deepening A* search}. These are the realms in which computer intelligence outweighs even the most extraordinary human minds. \textit{IBM's Deep Blue} beat \textit{Garry Kasparov} in chess \cite{KASPAROV}, and \textit{Google's AlphaGo} defeated the then grandmaster \textit{Lee Sedol} in the game of \textit{Go} \cite{ALPHAGO}. Neither system uses human understanding or beliefs. They can break beliefs by using experimentation and logic.\par 

By constraining a system to be well defined, programmers or mathematicians can represent every possible outcome using brute-force computing. By contrast, partially observable or obscure problems are much more challenging. These problems require some form of a \textit{best guess}. A good example is control systems for autonomous cars, where decisions are made based on a combination of partial observations and prior experiences; in these environments, there will always be situations that are new and unknown.\par

It is worth pointing out that AlphaGo was trained by humans and took several weeks to become a grandmaster. Its successor, \textit{AlphaZero}, learned by playing against itself, and it only took a few days to match and surpass AlphaGo. Both used Reinforcement Learning, which is a reward-based system. AlphaGo uses \textit{innate knowledge} approach, effectively starting from a known point of understanding. AlphaZero started with a clean slate (\textit{tabula rasa}) \cite{du2019creativity}. The computational hardware required for AlphaZero is considerably less than that for AlphaGo.\\

\noindent
\textsc{\textbf{Demonstrative/Non-demonstrative inference:}} Demonstrative inference is one in which the premises can directly form the conclusion. In other words, a decision is valid only if the premises are true. By comparison, non-demonstrative inference is when the premises do not directly form the conclusion. The conclusion can be entirely false even though each premise is, in fact, true. The connections mean that truth-preserving occurs only in demonstrative inference and not in non-demonstrative inference \cite{sep-wesley-salmon}.\\

\noindent
\textsc{\textbf{Ampliative/Non-ampliative inference:}} Ampliative inference is frequently used in reasoning to mean \say{\textit{adding to what is already known}}. This extension occurs when the conclusion contains content that was not present, \say{\textit{either explicitly or implicitly, in the premise}}. In a murder mystery novel where the main focus is on two characters, A and B, as the main suspects, it turns out on the last page that an unexpected character, C, is the murderer \cite{sep-wesley-salmon}.\par

Philosopher \textit{David Hume}, in \textit{An Enquiry Concerning Human Understanding} \cite{HUME1901-HUMAEC-13}, believed that there are two forms of reasoning: \textit{relations of ideas} and \textit{matters of fact}. The relationship of ideas is both a demonstrative and a non-ampliative inference. In contrast, a matter of fact is an ampliative inference and is not necessarily truth-preserving. We can draw a true conclusion even though the premises are false. \\

\textsc{\textbf{Limitations of current models:}} \textit{Tom Zahavy}, a researcher in Machine Learning, wrote an article titled \textit{LLMs cannot jump} \cite{zahavy2026llmscantjump}. Zahavy argues that the current Large Language Models can handle induction and are becoming more capable in deduction, but they still struggle with abductive jumps. \textit{Why is this important?} Abductive jumps are fundamental for scientific discoveries. Zahavy claims that scientific discoveries require moving from sensory or simulated physical experiences to new axioms. Invention requires more than compressing patterns from the data or proving consequences from existing premises. To obtain inventive solutions, we need to change the system to go beyond language-only models and include the physical world (the physical laws as we know them) and interactive world models \cite{zahavy2026llmscantjump}. We should expect the next big developments to be around abductive jumps. 

\subsection*{\textbf{5-Thoughts}}

\begin{itemize}[leftmargin=*]
\item Reasoning is at the core of intelligence. It is the ability to create a logical connection between having a conclusion and finding the evidence, or having evidence and finding a conclusion. 
\item We use reasoning to rationalize decisions by using deductive and inductive reasoning. Deductive reasoning being more mechanical, while inductive reasoning requires some form of an inspirational leap.
\item \textit{Inductive biases} arise when multiple solutions are possible, and a system must rely on prior assumptions or preferences to select the one most likely to correspond to the real world \cite{griffiths2026laws}.
\item A conclusion can amplify the premise, bringing more information into view. We can even obtain the right conclusion with the wrong premise.
\item Abductive reasoning could be the next big advancement in Artificial Intelligence. This allows systems to take innovative leaps to solve hard problems. 
\end{itemize}

\subsection*{\textbf{Recommended reading}}

\begin{itemize}[leftmargin=*]
    \item \textit{The Book of Why}, by Judea Pearl \& Dana Mackenzie
    \item \textit{The Creativity Code}, by Marcus Du Sautoy
    \item \textit{Introduction to Philosophy}, by John Perry et al. 
    \item \textit{The Laws of Thought}, by Tom Griffiths
\end{itemize}
\section{Physical system design}
\label{sec:systems}

\vspace{1.0em}

\begin{adjustwidth}{0.3cm}{}
\textbf{>> Any form of intelligence is constrained by its physical implementation, which determines its speed, capabilities, and resource requirements.}
\end{adjustwidth}

\hspace{0.5cm}

\begin{quote} 
\centering 
\vspace{1ex}
\textit{\say{Human intelligence is a low-bandwidth phenomenon, and is as much if not more about the scaling down of effort as the scaling up of capability}}\\
\textbf{David Krakauer et al., 2025} \cite{krakauer2025largelanguagemodelsemergence}
\end{quote}

\hspace{0.5cm}

\begin{figure*}[htp]
  \centering
  \includegraphics[width=0.7\linewidth]{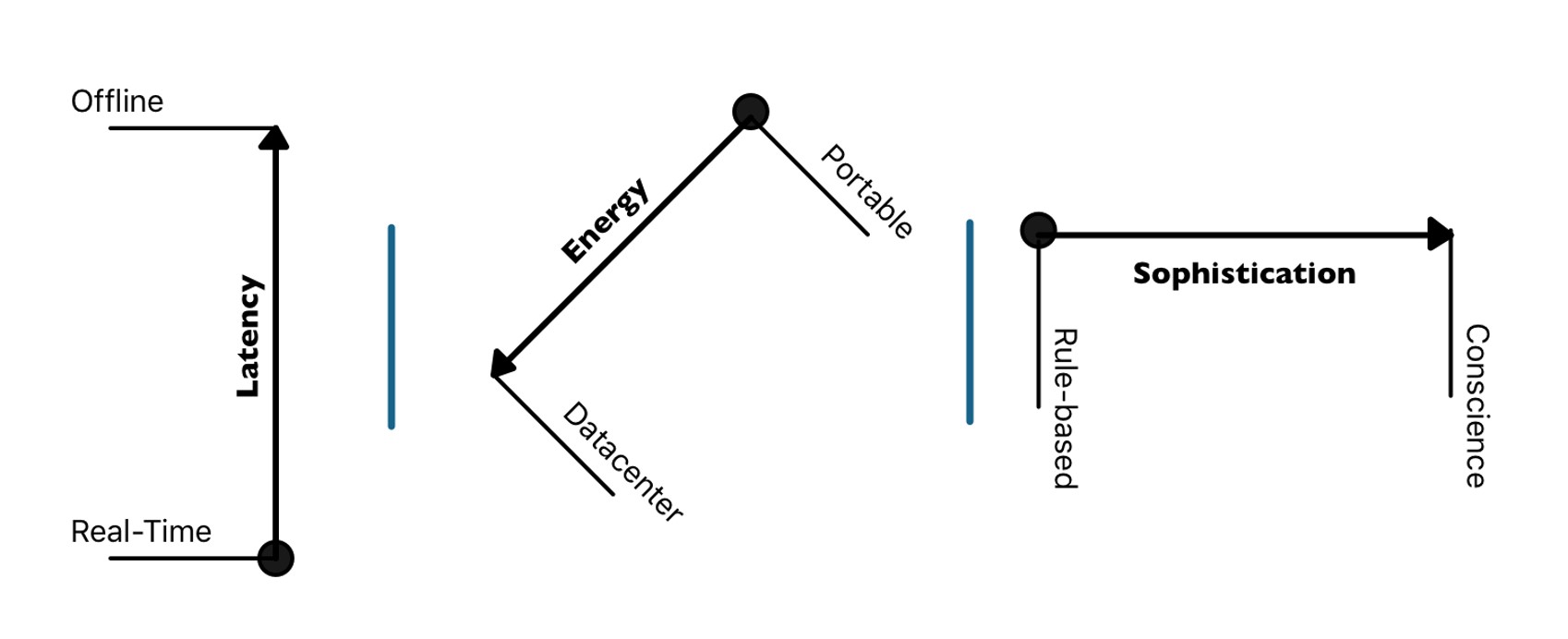}
  \caption{Latency, Energy, and Sophistication (LES) variables}
  \label{figure:dim}
\end{figure*}

Intelligence has a physical component. Physical location determines performance, latency, and capability within the body or system. The physical location influences the type and style of intelligence. Performance determines whether intelligence has the speed necessary to produce the desired result. Latency concerns response times; that is, having a perfect answer is of no use if the answer is too late. For intelligence, response times can be strict or relaxed depending on the problem domain, e.g., driving a car or designing an airplane wing. Finally, capability: \textit{Does the intelligence have enough functional units to take in all inputs and execute the rules to produce a solution?} \textit{When is a system too limited or a problem too large to handle?} \par

Memory considerations include computation and communication, that is, short-term and long-term memory. \textit{Is the memory required to be in the foreground, or is there a background element?} Is the memory long-term, where it will rarely be needed, or short-term and immediately required? For animals, the combination of computation and memory occurs in a very non-\textit{von Neumann} fashion and is inseparable, it is episodic memory (see Section \ref{sec:human}). \par

It is important to consider the latency of action in terms of cold and warm starts\textemdash the time it takes for the intelligence to start operating from an initial state. A key question is whether an intelligent system can be kept \textit{warm} so it can respond quickly when needed, and what the resource costs are for keeping that system in that state. In practice, there are several delays to consider: cold to warm, warm to action, and cold to action.

\subsection*{Centralized and distributed intelligence}

In nature, the organization of intelligence occurs in different ways. For example, \textit{cephalapods} \cite{Octopus} demonstrate intelligent actions based on problem solving and memory. The evolution of their intelligence has created a unique structure. Rather than having a large central brain, octopus intelligence is distributed within the limbs; each limb resembles a distinct thinking entity that cooperates with the rest of the octopoidal body. \par

Another form of intelligence is in groups. Bees, Ants, soccer teams, and more recently, military drones exhibit what is called \textit{swarm intelligence}: here, the system itself is the intelligence.\\

\noindent
There are at least six distinct physical architectures or physical categories of intelligence. These categories occur in both natural forms of intelligence and artificial ones.

\begin{enumerate}[label={D-\arabic*:},leftmargin=*]

   \item \textsc{\textbf{Centralized}} intelligence requires one large brain to perform all tasks. All memory, knowledge, reasoning, and correlation occurs in this centralized location. All existing redundancies are localized. \\

   \item \textsc{\textbf{Decentralized}} intelligence has multiple brains, all remaining within a single entity. Knowledge, reasoning, and correlation are all duplicated throughout the entity. Decentralized means intelligence has distinct physical redundancy, and each unit can carry out similar tasks with high cooperation.\\

    \item \textsc{\textbf{Distributed}} means that intelligence is part of a vast interconnected network, physically located away from each other, and not necessarily part of the same single entity. A network potentially involves some form of hierarchy. Where the nodes can be heterogeneous and redundancy occurs due to scale. In computing, this falls under N-version programming or Byzantine Algorithms. \\

    \item \textsc{\textbf{Swarm}} intelligence is similar to distributed intelligence, but physical nodes are much more localized and act as one entity. The nodes within the swarm influence their neighbors. Redundancy occurs at the node level. If a node fails, the system can self-reconfigure. We are starting to see deployment in the military arena \cite{youvan2026}. \\

    \item \textsc{\textbf{Cloud model}}, for Artificial Intelligence, is an example of a hybrid structure. Where loosely coupled data centers have the fail over capability, but at any given time, there is one dominant data center. Redundancy exists because each data center in the group can become a primary. \\

    \item \textsc{\textbf{Embodied}} intelligence includes some of the above examples but not the data center model. Embodied intelligence refers to the phenomenon of a greater capacity for understanding the world when intelligence resides in a physical sensing form, that is, a system that exists and interacts with the world and understands it differently than the isolated one. This is robotics.\par
    
    As robotics becomes more capable, Artificial Intelligence can be given a physical form, creating a more complete model of embodied intelligence \cite{ieeeMeetSmarter}. See Section \ref{sec:introduction}, \hyperref[sec:trusted simulator]{\textit{Are We Living in a Trusted Simulator?}} The \textit{brain-in-a-jar hypothesis} asks whether intelligence can be isolated from the body: can the mind exist independently, or does the body’s electrochemical processes help shape perception, judgment, and decision-making? For the closest practical model of embodied intelligence in computing, see \hyperref[sec:reinforcement learning]{\textit{Reinforcement Learning}} in Section \ref{sec:mathematic}.\\
    
\end{enumerate}

As intelligent systems move out of a central location, coordination and communication latency become significant bottlenecks, i.e., scaling effects arise. Pressure occurs because the system needs to optimize both the edge nodes and the nodes dedicated to coordination. In particular, how knowledge is structured, disseminated, and eventually transferred between all system nodes becomes essential. Language, either written or spoken, plays a necessary role for humans. Scaled federated systems \cite{fedarch} (a specific form of distributed systems) can be an effective mechanism to create a higher order of Artificial Intelligence. 

\subsection*{Artificial variables}

In designing a system, we must consider all the different constraints. Figure \ref{figure:dim} shows three possible design variables, namely latency, energy, and sophistication. These design variables determine the type of problems that are addressed. These are separate constraints; yet, they can be equally important. Specific issues will have additional limitations. \par

Latency is the time required to make a decision. For example, an autonomous vehicle makes most decisions instantly, similar to when humans are driving. \par

Energy is also a constraint; the energy required to make a decision may be much more than what is locally available. In these situations, system intelligence may have to be adopted. Computers are attached to a network; humans connect to society or political hierarchies. Alternatively, deciding on the latency requirements might help reduce the energy requirements, i.e., high latency means less peak energy required.\par

\textit{Tim Palmer}, a physicist, conjectured that human intelligence possibly has two energy levels \cite{TIMP2022}: low and high. In the low-energy state, computation is much more stochastic or dream-like. The low-energy state can explore more complex problems. The high-energy state is more about implementation and quick decision-making. The idea here is that the energy level determines the type of computation that occurs. \par

Finally, the required sophistication makes a difference in how intelligence is organized. If the requirement is rule-based, then the system must be able to execute a set of rules. In contrast, something more sophisticated, such as conscience (right or wrong), requires a more complicated structure and approach.

\subsection*{Larger model argument} 

This can also be called \textit{the argument for larger datacenters}. It is interesting to observe that we often prefer a single attribute to help explain capability, for example, the fastest or highest frequency. In modern artificial models, we have the number of tokens or tokens per second as a measurement. In a paper by \textit{Yizhou Liu} et al., titled \textit{Superposition Yields Robust Neural Scaling}, the researchers argue that \textit{superposition}\footnote{Superposition refers to a system’s ability to represent multiple concepts in a shared cognitive space \cite{Elhage2022ToyModels}. The characteristics explain how both biological brains and advanced Artificial Intelligence manage massive amounts of complexity.} This is an important reason why larger neural networks scale so reliably \cite{liu2025superpositionyieldsrobustneural}. Their central idea is that features in the \textit{representation space} can be packed more densely than the available dimensions would normally allow. Using smaller models, Liu and colleagues show that strong superposition can reduce loss and help explain why larger models often improve in a smooth and predictable way. However, many still believe that these gains will eventually diminish as the models become too large and cumbersome. We should also note that biology, especially for us, did not follow this path of ever increasing model size.

\subsection*{Sovereign AI}
Looking at a different dimension of physical intelligence. There is the concept of \textit{sovereign AI} as a physically located entity, but the location is based on nation states. The control, creation, and deployment of models are within the borders of a nation. Human intelligence has long been regarded as a nation state's advantage or national asset. For human beings, education is considered a priority in many countries. With the rise of Artificial Intelligence, nation states are beginning to realize that they need localized capabilities; otherwise, they will lose their commanding position. These abilities come from copying technology or growing their Artificial Intelligence programs.\par

Along with this realization comes the stresses placed on the infrastructure to support such a requirement\textemdash power, cooling, and land. In May 2025, the \textit{United Arab Emirates} (UAE) announced that all residents would have free access to a large language model (ChatGPT Plus). This announcement marks the first step towards making Artificial Intelligence a utility within a nation. 

\subsection*{Copying, and distillation}

Copying allows one model to be transferred to another physical device; that is, a form of cloning. This is useful when core knowledge has been recorded and we want to use that core knowledge with some form of specialization. Transfer can also involve conversion into smaller models, i.e., distillation. Distillation is when a complex model, sometimes referred to as \textit{the teacher}, is used to transfer knowledge to a smaller physical form called \textit{the student}. The student model has all the essential attributes of the teacher but is much smaller and more efficient. The loss of some critical knowledge could be a concern.

\subsection*{\textbf{3-Thoughts}}

\begin{itemize}[leftmargin=*]
\item Intelligence is physical. There is a foundation that creates intelligence. We do not understand the details, but we are aware of the significant physical requirements and behaviors. 
\item We must decide on the attributes of latency, energy, and sophistication. Each attribute determines the type and cost of the solution. For biological systems, the trade-off may be accuracy and weight; for artificial systems, the cost is more financial and spatial. 
\item We design artificial systems, so we need to determine the high-level architecture on which these systems will run. From centralized and individual to distributed and federated. The demands affect the style of the systems being created. 

\end{itemize}

\subsection*{\textbf{Recommended reading}}

\begin{itemize}[leftmargin=*]
    \item \textit{Atlas of AI}, by Kate Crawford 
\end{itemize}

\section{Measuring intelligence}
\label{sec:measuring}

\vspace{1.0em}

\begin{adjustwidth}{0.3cm}{}
\textbf{>> We are constantly comparing ourselves with other animals, other people, and artificial systems. Intelligence is often measured to classify, rank, and select. Unfortunately, measurement is not that simple. When we speak of intelligence, what exactly are we trying to measure? }
\end{adjustwidth}

\hspace{0.5cm}

\begin{quote} 
\centering 
\vspace{1ex}
\textit{\say{The trouble with AI is that it confuses being knowledgeable with intelligence}}\\
\textbf{Paraphrasing David C. Krakauer, 2025} \cite{InterviewDavidKrakauer}
\end{quote}

\hspace{0.5cm}

One of the most critical questions we have to ask is how to measure intelligence. \textit{How do we know that someone or something is intelligent? As an example, is a rock intelligent?} If not, why not? We have many controversial tools to determine human intelligence, but what about Artificial Intelligence? In the past, we have seen the environment being essential, and Artificial Intelligent Systems can seem superintelligent in highly constrained microworlds \cite{Norvig2020}, e.g., the block world \cite{blockworld} being a good example. Today, we see new systems that are able to overtake traditional human processes. For example, a paper by Brodeur et al. discusses how Large Language Models are overtaking physicians in their ability to reason medical diagnoses \cite{arxivSuperhumanPerformance}. \par

\textit{How do we measure intelligence?} Fundamentally, most measurements involve testing an agent, and testing involves having it complete a specific task. These tasks range from reciting facts from memory to creating long-form answers to physical problems, e.g., puzzles. We can create \textit{absolute} tests, such as a maze problem where the test provides a finite number, such as the time to completion, or a \textit{relative} test, such as \textit{Human Parity Benchmarks}, where a comparison is used. In the case of human-parity benchmarks, the comparison is achieved using humans as a control experiment. We will explore some of these tests next.

\subsection*{Standardized testing}

Standardized testing has become ubiquitous for assessing intelligence. From grade school to graduate school, testing is the most common way for students to demonstrate their intelligence. \par 

Criticisms of standardized testing arise from the fact that not all students are able to learn in the same way or exhibit specific intelligence. It also tends to ignore the creative aspects of intelligence, assuming that mathematics and reading comprehension are the only ways to exhibit capability. Standardized testing is based on the assumption that intelligence is uniform. As this paper hopefully shows, the problem is much more complicated. \par

The \textit{Allen Institute for Artificial Intelligence} announced in 2019 that it created an Artificial Intelligence system that could pass the standard 8\textsuperscript{th}-grade science test \cite{ALLENAI}. According to the \textit{New York Times}, the system correctly answered over 90 percent of the questions on an 8\textsuperscript{th}-grade science test and more than 80 percent on a 12\textsuperscript{th}-grade exam. Since we wrote this piece, OpenAI's GPT-4 has passed medical, law, and business school tests \cite{businessinsiderModelsLike}. It has also moved from passing with a minimum grade to passing with proficiency, opening up the possibility that standard tests may become secondary; see \textit{Large Language Models} Section \ref{sec:llm}.

\subsection*{IQ test}

For many decades, the \textit{Intelligence Quotient} test has been the de facto intelligence test, with a score of 100 being the average across the population. High scores secure a prestigious invitation to \textit{Mensa}. \par

However, one's IQ typically relates not to how much knowledge one has but to one's capacity for learning. Another way to look at this is by considering how much water is in the glass versus the capacity of the glass. Are humans born intelligent, or do environmental effects make them wise (i.e., do pressure points make us more intelligent)? Or, more realistically, is it a bit of both?\par

Again, there are signs that IQ tests explicitly designed for Artificial Intelligence are starting to appear. \textit{Washington State University} (WSU) claims to have created the first IQ test for Artificial Intelligence \cite{AIIQTST}, funded by the \textit{Defense Advanced Research Projects Agency} (DARPA). WSU developed a test that measures real-world performance in novel and unknown environments that do not account for complexity. The end score considers \say{\textit{accuracy, correctness, time taken, and the amount of data needed to perform}}.

\subsection*{Language as a measure of intelligence}

The \textit{Philosophy of Language} is an endeavor to categorize how humans learn a language. The focus is on how the brain stores symbolic language and what the primary concepts involved are.\par

From a simple measurement perspective, the size of one's vocabulary can measure intelligence. Shakespeare may have had a vocabulary of 40,000 to 60,000 words. The average English speaker has a vocabulary of 20,000 to 30,000 words. We can produce an intelligence scale from intermediate to high using these measurements. \par

Other aspects of language include the creation of natural language. \textit{Facebook} had to shut down an Artificial Intelligence system that created a secret language \cite{AILANG}. The Facebook system aimed to explore the subject of negotiation, but the plan quickly went awry. Once connected, two Artificial Intelligence systems started learning from each other's mistakes and quickly developed a unique language. These systems originally started with basic English.\par

Large language Models also represent this measure; see \hyperref[sec:llm]{Section \ref{sec:llm}}. Large refers to the size of the model, but it could also refer to the size of the corpus of its data. Large Language Models, such as ChatGPT, appear intelligent due to their proficiency in handling natural language.\\

\subsection*{Industry benchmarks}

Industry benchmarks for intelligence focus on several attributes associated with the operational aspects of machine learning: \textit{how big?} \textit{how much power?} or \textit{how capable?} Listed below are two benchmarks used in the industry.\\

\textsc{\textbf{Language Model Arena:}} \textit{Large Language Arena} (\textit{LMarena}) is a benchmark for comparing the energy efficiency and performance of large-scale language models \cite{lmarenaAboutArena}. It is an open platform for crowd-sourced AI benchmarking. The development was initiated by UC Berkeley's Sky Computing Lab in 2023. \textit{Large-scale language models}, at the time of writing, contain hundreds of millions to trillions of parameters. For example, \textit{GPT-4}, \textit{PaLM}, and \textit{LLaMA} are all large-scale models. It examines inference using real-world workloads. It measures energy usage (in watts), latency, and throughput. Energy efficiency enables this benchmark to focus on sustainability, specifically the cost of deployment \cite{lmarenaAboutArena}. This allows for the comparison of model size, architecture, and hardware configurations.\\

\textsc{\textbf{Machine Learning Performance:}} Machine Learning Performance (\textit{MLperf}) is a benchmark created in 2018 by a consortium of educators and engineers in the industry \cite{mlcommonsMLCommonsReleases}. Today, it is run by\textit{MLcommons}, a non-profit organization founded in 2020 dedicated to advancing machine learning. It is a suite of benchmarks that focuses on performance and scalability. It covers tasks from image classification to language processing. MLperf differs from LMarena in that it covers both training and inference workloads. The main emphasis is on reproducibility, fairness, and comparability. It attempts to measure progress in all three areas.

\subsection*{Turing test}

The \textit{Turing test} has become infamous in the world of Artificial Intelligence. The idea proposed by Alan M. Turing in 1950 as a thought experiment, originally called the \textit{Imitation game} \cite{TURING1950}, describes a simple experiment to determine intelligence. The test focuses on a machine's ability to exhibit intelligent behaviors indistinguishable from those of a human. A person interacts through an interface with an unknown entity. That entity can be another human or a machine. A machine has passed the Turing test if the person cannot distinguish which one is human and which one is a machine. \par  

An Artificial Intelligence system has passed the Turing test \cite{readingTuringTest}. The passing is not without controversy. The Chatbot is called \textit{Eugene Goostman} and acts like a  13-year-old Ukrainian boy. The chosen age and country of origin forced the judges to compensate for errors and awkward grammar. The \textit{Royal Society} held the test in London in 2014. Eugene Goostman managed to convince 30\% of the judges that it was human for 5 minutes during a typed conversation. \par

A variation of the Turing test appeared in the movie \textit{Ex Machina}, known as the \textit{Garland test}\textemdash named after the director \textit{Alex Garland} \cite{exmachina2014}. It is a philosophical concept of whether a human can still attribute consciousness to an Artificial Intelligence system when they already know it is a machine. \par

\subsubsection*{Efficiency:}

Back of-the-envelope calculation. Let us create an efficiency equation for passing the Turing test, \footnote{Inspired by Paul Gleichauf, scientist}.\\

\begin{equation} \label{eq:1}
TTEy=\frac{x}{y*d*h*p}
\end{equation}\\

Where $x$ is what you are comparing with (human or artificial), $y$ is years old, $d$ is days per year (365), $h$ is hours in a day (24), and $p$ is power consumption (for humans, it is 20 Watts). Using an average graduate age of 21 years, the efficiency of the \textit{Turing test 21} (TTE21) would have a divisor of 3.7 MWh (MegaWatt hours). This number crudely represents the training cost in Watt hours to pass the Turing test. \\ 

\begin{equation} \label{eq:2}
TTE21=\frac{x}{3.7 MWh}
\end{equation}\\

The quotient for a 21-year-old would be 1, compared to a 57-year-old, which is 2.7. Passing the Turing test takes a 57-year-old 2.7 times more training energy than a 21-year-old. The higher the quotient, the more energy is required to pass the Turing test. The quotient gives us a simple metric to determine the cost of attaining a level of intelligence that is indistinguishable from that of humans.  

\subsection*{Intelligince per watt} 
In a paper on local efficiency, \textit{Jon Saad-Falcon} et al. introduce the concept of \textit{intelligence per watt} (IPW) \cite{hazy2025ipw} as an important metric for artificial systems. This is important since both natural and artificial systems need to control the energy cost of performing specific tasks. Saad-Falcon and his colleagues argue that progress in Artificial Intelligence will not come simply from building ever-larger models but from improving efficiency, especially as power becomes a major constraint. \par

Nature did not create one massive centralized model but instead distributed intelligence across smaller, localized, and mobile nodes. For artificial systems, this suggests the need for clever routing methods that can preserve accuracy while reducing energy costs. Saad-Falcon and his team argue that Artificial Intelligence is likely to shift from centralized, power-hungry systems to distributed, efficiency-optimized architectures, guided more by maximizing \textit{intelligence per watt} than by raw capability alone \cite{hazy2025ipw}. So intelligence per watt becomes a critically important measuring metric.\par

It is worth mentioning \textit{TOPS per watt} at this point. TOPS stands for \textit{Tera Operations Per Second}. TOPS per watt measures how much computation a system can perform for each watt of power consumed. As discussed, this matters for intelligent systems because useful intelligence is not defined by raw performance alone, but by how efficiently we can transfer energy into useful work \cite{sloss_2026_21330151}.

\subsection*{Problem solving}

The measurement of problem solving ability is a way to assess reasoning capacity without language. One of the best ways to do this is through \textit{Behavioral Psychology}. Behavioral Psychology is a school of thought focused on observable and measurable intelligence. It assumes a blank slate (\textit{tabula rasa}) and that all aspects of intelligence are due to a learning process. Intelligence is directly proportional to the complexity of the problem domain. In this case, problem solving refers to the physical embodiment of deductive reasoning. \par 

\textit{Pavlovian conditioning} is the best example of Behavioral Psychology put into observable practice. By shaping the environment, specific actions can be correlated with an understanding of the world and the capacity for intelligence. Since animals cannot speak or refuse to talk to humans, this is one of the first examples of being able to test an animal's understanding. \par

Crows, primates, and cephalopods can reason several steps in advance to solve problems. In some cases, crows can solve problems better than most 5-year-olds. Crows can retrieve floating objects in containers that are just outside the reach of their beaks by adding rocks \cite{Crow}.\par

Problem-solving as a measure shows some promise as a method to evaluate certain kinds of Artificial Intelligence. This method is especially applicable to systems that navigate changing environments or encounter new problems to solve \cite{chollet2019measure}. \\

\textsc{\textbf{Example 1 - War Gaming:}} Comparing popular Large Language Models with cold war gaming scenarios \cite{payne2026shallweplayagame, payne2026aiarmsinfluencefrontier}. \textit{Kenneth Payne}, a strategist, speculated about how models would react to Cold War style simulation scenarios as a method of measurement. Whether the models would react aggressively, passively, or simply give-in to threats. The game he used was based on a nuclear crisis. According to Payne, the currently popular Large Language Models have produced worrying results. These results included rapid escalation to firing tactical nuclear weapons rather than de-escalating the situation. This indicates how models might decide under pressure scenarios. \\

\textsc{\textbf{Example 2 - Weather prediction:}} Another style of rating intelligence is using the quality of prediction as a measurement. For example, the \textit{Center for Medium-Range Weather Forecasts} used a historical model to provide a ten-day weather prediction without physics equations \cite{prickett2025weathermodel}. The model learns atmospheric behavior from historical data instead of solving the usual equations. By measuring the accuracy of the prediction, a model can be rated. In this particular case, the model was fairly accurate in predicting the weather.

\subsection*{Measuring brain activity}

Measurement of brain electrical activity is vital to determine whether a nonresponsive person is in a \textit{vegetative state} or suffers from \textit{locked-in syndrome}. A vegetative state is when a person appears awake without awareness. In comparison, locked-in syndrome is a condition in which a person is aware but cannot communicate. Some forms of stroke or head injury cause these situations. \par

This issue is where the non-invasive \textit{ brain computer interface} helps determine levels of consciousness. For example, a method to help discover the level of consciousness is called \textit{Zap and Zip} \cite{koch2019feeling}. A sensor cap fits onto a patient's head. Sensors measure electroencephalograms, more commonly referred to as EEG. EEG is the activity that occurs in the brain. First, a magnetic pulse "zap" is applied. The cap sensors pick up the EEG signals. A large amount of unstructured information is collected. Then the zip compression algorithm is used to compress the data. The zip algorithm compresses common repeating patterns and leaves unique activities unchanged. As there is no discarded information, the zip compression algorithm is called a \textit{lossless compression algorithm}. The resulting file size can be used as a measurement of consciousness. A small file size indicates that most of the brain is operating automatically. A large file leans toward the patient who has consciousness but cannot communicate. Artificial Intelligence has no consciousness, so there is no equivalent method.

\subsection*{\textbf{4-Thoughts}}

\begin{itemize}[leftmargin=*]
\item Intelligence measurement is essential for determining accuracy, problem solving abilities, improvements, efficiency, and capabilities in both humans and machines. 

\item We have developed tests over centuries to measure human intelligence; we can modify some tests for artificial systems. Other tests are entirely new for the synthetic category. Since Artificial Intelligence mimics humans, can we accurately determine a difference? Or can we not differentiate between artificial and human intelligence.

\item The concern about benchmarks is that they force convergence towards a local maxima. Measurement of a local maxima is particularly concerning since it pushes systems in particular directions. This local maxima pushes a particular hardware architecture and software model as the only convergent solution. For example, we focus on the efficiencies of \textit{linear algebra}, \textit{activation} functions, and \textit{multiply-accumulate} operations, but it may turn out that plasticity or adaptability are more important areas to measure.

\item As alluded to throughout this section, there is a focus on measuring capability in intelligence, but an increasingly important characteristic is efficiency.

\end{itemize}

\subsection*{\textbf{Recommended reading}}

\begin{itemize}[leftmargin=*]
    \item \textit{The Feeling Of Life Itself}, by Christof Koch 
    \item \textit{Human Compatible}, by Stuart Russell
    \item \textit{Artificial Intelligence: A Modern Approach}, by Stuart Russell et al.
    \item \textit{On the Measure of Intelligence}, by François Chollet
\end{itemize}
\section{Mathematically modeling intelligence}
\label{sec:mathematic}

\vspace{1.0em}

\begin{adjustwidth}{0.3cm}{}
\textbf{>> Artificial intelligence is based on mathematical principles. Mathematics is used because it is our most sophisticated system for expression and communication. We use mathematics to mimic nature. }
\end{adjustwidth}

\hspace{0.5cm}

\begin{quote} 
\centering 
\vspace{1ex}
\textit{\say{Mathematics is a more powerful instrument of knowledge than any other that has been bequeathed to us by human agency.}}\\
\textbf{René Descartes, philosopher, 1701} \cite{descartes1701rules}
\end{quote}

\hspace{0.5cm}

In an attempt to make the world understandable, humans use mathematics. Mathematics is said to be either invented or naturally occurring. If we believe it to be naturally occurring, it is a discipline of discovery rather than an invention of ingenuity. \textit{Immanuel Kant}, the philosopher, said \say{\textit{that it was nature herself and not the mathematician who brought mathematics to natural philosophy}} \cite{sep-kant-mathematics}. In our discussion, we choose the path of invention. \par

Mathematics is a collection of rules that attempt to model and map the world around us. There are a vast number of techniques for algorithmically representing the world. It is a tool that is gaining capability. \textit{Is it only a matter of time before we can model intelligence mathematically?} The following are what we believe to be the most focused mathematical models for intelligence.\par

In many cases, these methodologies require an expert, in most cases a human, to assist in pointing out the correct from the incorrect. While reading this section, the essential element to remember is the \textit{George Box} aphorism: \say{\textit{All models are wrong, but some are useful.}}\par

We create models to generalize and abstract. We could make a model so complex that it reflects reality or so simple that it is easier to implement and train. Models are essential in Machine Learning. From the outside looking in, we need to train a model; the model itself has to have the ability to learn. In general, there are two main categories of learning, namely \textit{supervised} and \textit{unsupervised}. As the name implies, supervised learning requires an expert to dictate. The expert determines what is correct and what is incorrect. Unsupervised learning requires no domain expert to intervene. There are subcategories, which include, for example, \textit{semi-supervised} learning. \par

We will categorize each model by pairing it with one of the previously defined elements associated with intelligence and highlighting their unique \textit{biomimicry}. Biomimicry is the emulation of processes found in nature. The point for this is two-fold. First, we put the methodology into perspective and second, we reinforce the notion that replication requires a detailed definition and a form of measurement.

\subsection*{Specialization vs generalization}

Statistical modeling is all about optimizing based on a desired result. Engineers, researchers, and data scientists define the methods and desired outcomes. \par

\begin{figure}[htbp]
  \centering
  \includegraphics[width=0.7\linewidth]{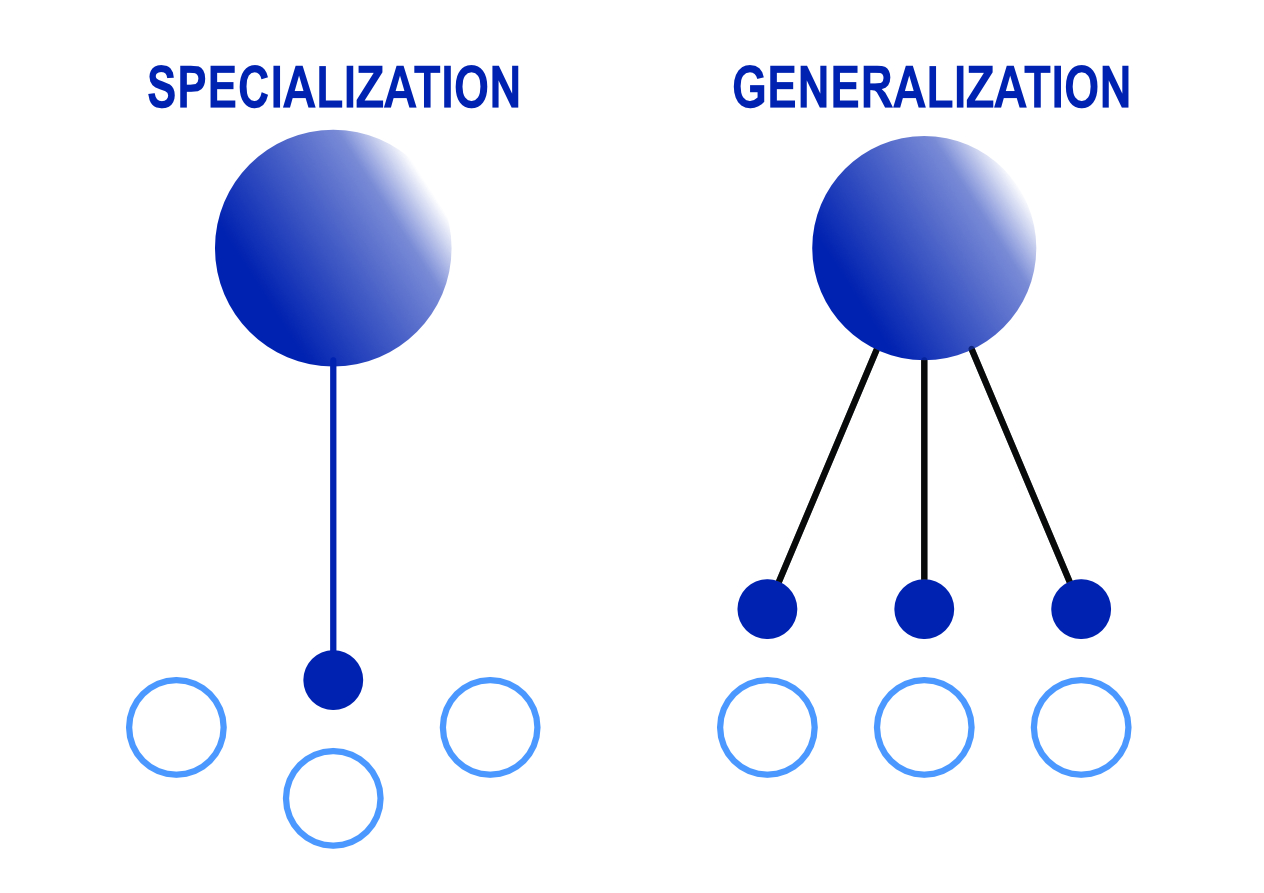}
  \caption{\textit{Specialization verses generalization}}
  \label{figure:spec}
\end{figure}

In a landmark 2019 paper, researcher \textit{François Chollet} defines intelligence as follows: \say{\textit{The intelligence of a system is a measure of its skill-acquisition efficiency over a scope of tasks, with respect to priors, experience, and generalization difficulty}} \cite{chollet2019measure}. This observation is helpful to remember, as most, if not all, of these methodologies optimize for a specific task. Chollet points out that there is no quantifiable measure for the general population. The more optimized a model is for a particular job, the more specialized it will be, and the less general it will become. \par

Historically, we believed that specialized knowledge came from generalized knowledge. However, the following methods show how to obtain technical expertise without general knowledge. Each model has constraints, but all focus on optimizing a particular feature. For example, reinforcement learning (for now) is optimized along one variable defined by its reward function. \par

\subsection*{Bayesian probabilism}

\textit{Biomimicry note} | Paired with deductive reasoning and statistics\\

Bayesian probabilism is where the comparison between human and Machine Intelligence stops, as biological systems are generally weak at statistics. As mentioned previously, Judea Pearl believes that humans are wrongly wired or, more precisely, \textit{intentionally} wrongly wired for other evolutionary priorities \cite{Pearl}. Bayesian probability primarily lends itself to robotics and Simultaneous Localization and Mapping (SLAM). The reason for the popularity in these areas is that it seems to be the best method to mitigate multiple sources of information to establish a consensus of \textit{best known}. For example, while a robot tracks its path, it always operates on minimal information and makes decisions based on statistical likelihood. \par

\textit{Bayes Theorem} underpins Bayesian probability, a mathematical equation for how justified a specific belief is about the world.\\

\noindent
Bayes Theorem is defined as follows:

\begin{equation} \label{eq:3}
P(A|B)=\frac{P(B|A)P(A)}{P(B)}
\end{equation}\\

Bayes' theorem provides a mathematical framework for updating what is known when new evidence appears. Most problems contain unknowns; additional variables (or hypotheses) may be required to represent the unknown factors. In probabilistic settings, entropy measures the remaining uncertainty. Because Bayesian reasoning estimates how likely a proposition is, given the available evidence, the system's known information must be represented, organized, and updated. \par

\subsection*{Deep learning}

\textit{Biomimicry note} | Paired with the physical building blocks of the human brain; neurons\\

Given the popularity of Deep Learning, there is a gamut of material available on the subject. Here we will briefly review the biological basis and the mathematical functions that allow Deep Learning models to \textit{learn}.

\subsubsection*{Artificial neurons}

\begin{figure}[h]
  \centering
  \includegraphics[width=1.0\linewidth]{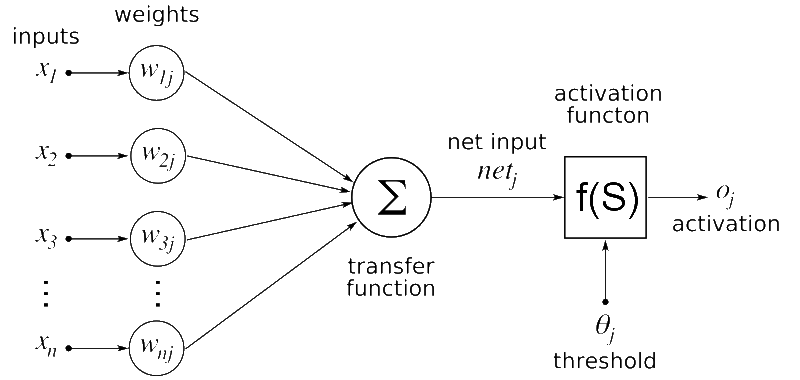}
  \caption{Artificial Neuron (Wiki media Commons)}
  \label{figure:artificialneuron}
\end{figure}

The fundamental building block of deep learning is the \textit{artificial neuron} (AN), also commonly known as a perceptron. An Artificial Neuron imitates the same neurons found in the brain. A single biological neuron has inputs in the form of \textit{dendrites} and outputs in the format of \textit{axons}. The equivalent machine, the artificial neuron, is shown in Figure \ref{figure:artificialneuron}.

Here we have a set of inputs, a \textit{transfer function}, and an \textit{activation function}. A \textit{threshold} is also often referred to as the \textit{bias}. In simple terms, a simple \textit{Artificial Neural Network} (ANN) can consist of only a few neurons. An Artificial Neural Network typically consists of Artificial Neurons arranged in layers, each layer connecting to the next. Deep learning is a very complex layered system of thousands of neurons. Through a mathematical function known as \textit{backpropagation} and supervised learning, these weights and thresholds will slowly and carefully modify over time to give the correct answer when given a set of inputs.

\begin{figure}[htbp]
  \centering
  \includegraphics[width=0.7\linewidth]{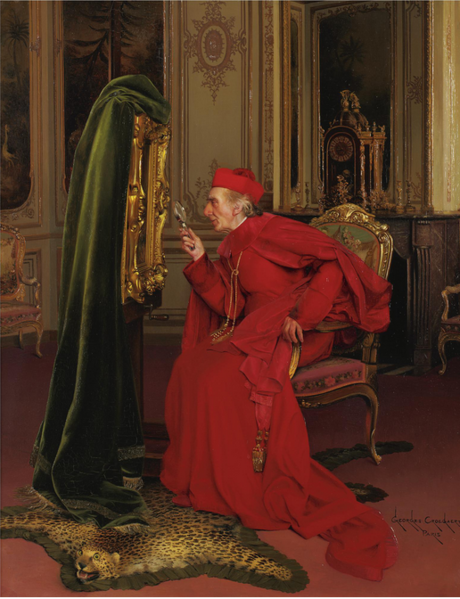}
  \caption{\textit{The Art Critic} \cite{croegaert_art_critic} (Wikimedia Common)}
  \label{figure:artcritic}
\end{figure}

\subsection*{Generative Adversarial Networks}
\textit{Biomimicry note} | Paired with creativity\\

\textit{Generative Adversarial Networks}, or GANs, are based on \textit{Deep Neural Networks} (DNNs), where two networks compete with opposing goals on the same data set. Generative Adversarial Networks focus on generating images, videos, or audio; that is, anything from \textit{Deepdreams} to \textit{Deepfakes}. Creating a human face where no such person exists is an example of a Deepdream \cite{karras2019analyzing}, whereas placing a known politician in a old pop music video is a Deepfake.\par

A Generative Adversarial Network consists of two networks: a \textit{generative network} and a \textit{discriminative network}. These are often called the \textit{creator} and the \textit{critic}. The critic, see Figure \ref{figure:artcritic}, is similar to a typical supervised network, trained to recognize specific cases. The creator's role is to fool the critic through clever feature extraction. Over time, and given enough samples, the creator can mimic the features found in the dataset. The critic network judges the creator's network using the same training data as a comparison point. In this case, both networks train each other in a digital arms race, with the creator becoming more capable of counterfeiting solutions and the critic becoming more discriminative. Eventually, the creator becomes so good that the critic can no longer tell what is original. \par

A particular application of GANs is Deepfakes \cite{deepfake}, which have grown in popularity over the last few years, thanks to the ease of creating them and their potential applications. These range from political disinformation to blackmail and general amusement. We see a growing concern about the potential for deepfakes to influence people's decisions and viewpoints. Today, we have seen the start of genuine consequences, from kidnappers using Deepfake voices pretending to abduct a child \cite{cnnMomThese} to Artificial Intelligent chatbots designed specifically for malware purposes \cite{techWormGPTChatGPT}. With so many ways to generate deepfakes, we hope there are equal methods to identify their inaccuracies.

\subsection*{Reinforcement learning}
\label{sec:reinforcement learning}
\textit{Biomimicry note} | Paired with problem solving \\ 

Reinforcement Learning (or simply RL) is another subset of Machine Learning in which Neural Networks make up the building blocks of the algorithm. Researchers Shane Legg and Marcus Hutter stated \say{\textit{Intelligence measures an agent's ability to achieve goals in a wide range of environments}} \cite{collected-intelligence}. Reinforcement Learning approaches this incrementally, tackling one environment at a time through reward-guided exploration. \par

When looking at Reinforcement Learning, it is essential to remember that, at its core, it is a \textit{behavioral model} similar to \textit{Pavlovian conditioning}. Everything is an action-reward pair. For example, positive reinforcement dictates that a meaningful positive reward is received when a model takes the right action. 

\begin{figure}[h]
  \centering
  \includegraphics[width=1.0\linewidth]{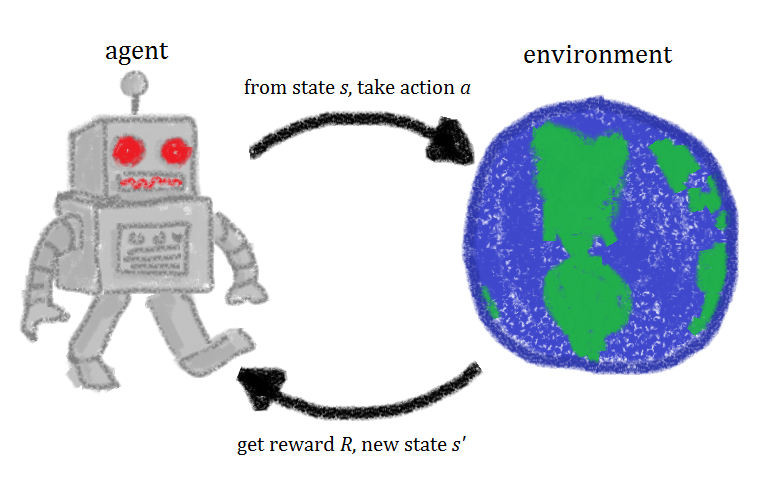}
  \caption{RL Agent (Wiki media Commons)}
  \label{figure:rlagent}
\end{figure}

Figure \ref{figure:rlagent} shows the Reinforcement Learning agent, environment, and reward function.

\subsection*{Evolutionary processes}

\textit{Biomimicry note} | Imitating Darwinian evolution \cite{simon2013evolutionary,DBLP:journals/corr/abs-1906-08870}\\

The evolutionary process involves creating a population of individuals either from a known starting point or completely at random. \[\{x_i\}_{i=1}^{N}\] where x represents each individual in the population. The individual is a potential solution to a problem; it can be a set of variables or a structure. Each individual has an associated fitness value: \[f(x)\] The fitness number identifies how close an individual is to a possible solution or result. This fitness value can be calculated using different methods. The individual evolves through mutation and/or crossover. Mutation is the random alteration of one of the components that make up the individual: \[x^{'}=x+\delta\] Here, the change $\delta$ can be applied randomly to various components of an individual. Crossover involves copying a portion of an individual that is then inserted or replaced in another individual, inheriting some of the characteristics: \[x_{child}=C(x_{1},x_{2})\] Finally, a population generation evolves into another generation. The fitness number is used to select the parents of the next generation: \[P^{(t)}\rightarrow P^{(t+1)}\] From parents, individual children are produced. That is, the new individuals for the next population evolve from the previous population. A solution can go through many generations to discover a result or completely diverge from it. It is a method of discovery, a new knowledge, when a search space is too large or complicated for traditional methods.

\subsection*{\textbf{5-Thoughts}}

\begin{itemize}[leftmargin=*]
\item Machine Learning is built on a foundation of mathematics. Mathematics provides the high-level constructs that underpin Artificial Intelligence and possibly biological intelligence.

\item At a very high level, we have specialization and generalization. The first narrows the focus of intelligence to a small area. The second is more about general understanding. 
\item Most mathematical modeling used in Artificial Intelligence is drawn from some form of biomimicry, with one notable exception: Bayesian methods, which are statistically grounded and built to handle changing degrees of uncertainty.
\item Currently, neural networks are executed on hardware as some form of stochastic array\textemdash a matrix whose elements represent probabilities and obey a specific set of rules.
\item We mentioned that GANs pair an artist with a critic; \textit{coevolution} takes a related approach, evolving two competing populations. The two ideas are similar in that they both use adversarial dynamics (i.e., competition and adaptation) to improve the solution.
\end{itemize}

\subsection*{\textbf{Recommended reading}}

\begin{itemize}[leftmargin=*]
    \item \textit{The Emperor's New Mind}, by Roger Penrose 
    \item \textit{The Book of Why}, Judea Pearl \& Dana Mackenzie
    \item \textit{Godel, Escher, Bach: an Eternal Golden Braid}, by D. Hofstadter
    \item \textit{On Growth and Form}, by D'Arcy by Wentworth Thompson
    \item \textit{Evolutionary Optimization Algorithms}, by Dan Simon
\end{itemize}

\section{Neurons}\label{sec:neurons}

\vspace{1.0em}

\begin{adjustwidth}{0.3cm}{}
\textbf{>> Neurons form the foundation of biological intelligence and, by analogy, modern Artificial Intelligence. Here we dig a little bit deeper into neurons and their differences.}
\end{adjustwidth}

\hspace{0.5cm}

\begin{quote} 
\centering 
\vspace{1ex}
\textit{\say{This point at the edge of chaos allows neurons to jump quickly between different states enabling them to alter behavior as necessary.}}\\
\textbf{Manfred Kitzbichler, 
Neuroscientist} \cite{camHumanBrain} 
\end{quote}

\hspace{0.5cm}

As mentioned in Section \ref{sec:history} on history, \textit{McCulloch} and \textit{Pitts} published a famous paper entitled \textit{A Logical Calculus of Immanent Ideas in Nervous Activity}. The paper describes how neurons (brain cells) could be used to construct logical circuits \cite{mcculloch1943logical,griffiths2026laws}. Each neuron sends a \textit{firing} or \textit{inhibiting} signal. A group of neurons can then be connected to trigger a signal once a threshold has been reached, thus forming a logical circuit. According to \textit{Tom Griffiths}, cognitive scientist, this paper played a critical role in the development of the \textit{von Neumann architecture} \cite{griffiths2026laws}. \par

As further development occurred, neurons were discovered to be fundamental for machine learning, deep learning, and biological control. We can see that intelligence depends on the underlying complexity of the layers and connections between neurons. To help simplify this investigation, let us first label biological neurons as $B^{n}$ and artificial neurons as $A^{n}$. In biological systems, $B^{n}$ are highly diverse and often specialized in their structure and function, with interactions better explained as higher-order representations than simply pairwise graphs. By contrast, most artificial networks built from $A^{n}$ use more uniform computational units (for example, multiply-accumulate with an activation function; see Section \ref{sec:mathematic}) and are typically modeled with much simpler, directional graph-based structures. \par 

Neurons can be summarized as having input and output signals, some computation, and characteristics that allow them to remember and learn. The inputs drive the computation, which determines the output response. At this point, $A^{n}$ and $B^{n}$ are fundamentally similar. However, in detail, they differ substantially in their options, speed of configuration, types of signals, and style of operation. \\

\textsc{\textbf{Biology:}} $B^{n}$ are highly specialized cells that belong to \textit{animalia} (animal kingdom). Other life forms, such as \textit{plantae} (plants) and \textit{fungi}, do not possess these specialized cells. $B^{n}$ are continuously active, oscillating. The oscillations occur roughly at 100 Hertz and below, with the average ranging from 0 to 50 Hertz. These oscillations are banded. Banded means that the neural oscillations do not spread evenly. The modulation of the output occurs with new input signals. The number of input signals per neuron ranges between 1000 and 10,000 connections. The signals themselves vary with electrochemical spikes, graded potentials\footnote{Short-lived changes in voltage that vary the amplitude depending on the inputs}, and temporal patterns\footnote{Precise timing, sequence, and arrangement of spikes, which can encode more complex information.}. They are continuous in time, so biological neurons are normally not discrete. They may exhibit delays, but the underlying behavior remains constant. \par

$B^{n}$ retains their state through a combination of chemical and electrical mechanisms. Its output is carried out by spike signals. Computation is nonlinear, stochastic, analog, and event-driven. In $B^{n}$, individual neurons or groups of neurons can form feedback loops in which outputs return to inputs, creating recurrent behavior. Data are typically sparse, neurons connect in three dimensions (that is, \textit{Volumetric Computing} \cite{sloss_2026_21330151}), and connectivity remains plastic and dynamic; that is, connections are not fixed \cite{kandel2013principles}. \par

The internal state arises through the membrane, ion channels, and neuro-modulators. The membrane maintains an electrical potential throughout the neuron. Ion channels are tiny gates that open and close, controlling the flow of charge. Neuro-modulators then adjust the strength of the response through chemical control. \par

Learning occurs through local plasticity, with continual fine adjustments over time. These adaptations may take place in milliseconds, minutes, or even years. \par

Neurons also differ across animals. We have already noted that neuron density is greater in birds than in humans, but there are other examples: a neuron in a flying insect may be more specialized than one in a mammal, since insects have far fewer neurons yet must still support flight. A \textit{fruit fly} has approximately 139,255 neurons and about 50 million connections, which is substantially less than a human brain; sadly, despite our superiority, we are incapable of flight either due to a lack of specialization or our low power-to-weight ratio \cite{ukriFirstEvery}. $B^{n}$ are generally highly efficient, a topic we will discuss after introducing $A^{n}$. It is worth mentioning that natural intelligence is also closely coupled to environmental factors such as the physical body (sensor network), hormones, and metabolism\textemdash as it has a physical embodiment component. \\

\textsc{\textbf{Artificial:}} $A^{n}$ provides a simplified view of neurons. Although artificial neurons can mimic some biological behaviors, most are built as graph-based layered models. They usually operate through discrete input-output steps, passing numerical values from one layer to the next. Basic artificial neurons are generally stateless; however, more advanced models can include state at an additional cost. Their output is typically passed through an activation function, such as \textit{ReLU}, \textit{sigmoid}, or \textit{tanh} \cite{prince2023understanding}. \par

$A^{n}$ differs in how they connect, and this is clear when comparing different types of models. For example, a neuron in a \textit{Convolutional Neural Network} (CNN), often used for object recognition, may have between 10 and 2,000 inputs. By contrast, a neuron in a Large Language Model may have many more, ranging from a few thousand to 10,000 or more. \par 

We use $A^{n}$ to form deterministic mathematical models. Each neuron computes a weighted sum of its inputs; the result is then passed through a nonlinear activation function. The neurons are structured as matrices to make the computation efficient. \par

The matrices are fixed in layers that are used for both training and inference. During training, the weights are updated; during inference, they remain fixed. This contrasts with $B^{n}$, which, as previously mentioned, has a physical embodiment, while $A^{n}$ is abstract. The abstraction comes partly from the fact that artificial systems must eventually be executed on a silicon substrate. \par

As already hinted, $A^{n}$ learning occurs through global optimization. Global optimization traditionally involves backpropagation and gradient descent. As a side note on $A^{n}$ learning, \textit{evolutionary strategies} can be seen as an alternative to traditional backpropagation at scale by replacing gradient-based updates with low-rank population changes. These changes can improve efficiency in the learning process. Although evolutionary strategies are a work in progress, they show the potential for much higher throughput and could even outperform traditional methods \cite{sarkar2026evolutionstrategieshyperscale}.

If we compare natural and artificial neurons, the $A^{n}$s are roughly a million times faster than the $B^{n}$s at raw switching speeds. $B^{n}$ compensates for slow switching speeds with massive parallelism, energy efficiency, and dynamic adaptation. Noting that these \textit{compensatory attributes} are not mutually exclusive and could, at some point, be implemented in $A^{n}$. \par

\subsection*{Why $B^{n}$s are much more efficient?}

There are a number of reasons why a $B^{n}$ is more efficient than the artificial equivalent. Firstly, $B^{n}$ can be event-driven. What does that mean? It means that they only handle \textit{spikes} (information of relevance). This consumes significantly less energy by firing only when relevant information occurs. $A^{n}$ operates continuously with all information, requiring constant computation. This is partly alleviated by using sparse data. We should point out that \textit{Analog Computing} can also be designed specifically to handle spikes, as-well as digital systems. Both artificial forms come with constraints. \par

In digital computers, memory and computing are physically separated, requiring energy-intensive movement of data back and forth. There are solutions that start to remove this barrier, called \textit{in-memory} and \textit{near-memory} computation. This involves placing computation much closer to the memory. These technologies are still not common. In biological systems, the idea of separation does not exist; computation and storage of information occur simultaneously within the same network \cite{kandel2013principles}. \par

Many $B^{n}$ in the brain are silent or rarely active, with estimates that only about 30–40\% of \textit{all} neurons are active at any given time, reducing overall energy consumption \cite{pmc7710421}.

 As mentioned previously, $B^{n}$ can connect to up to 10,000 other neurons, allowing for massively efficient parallel systems. Finally, the signals travel at very \textit{slow speeds} (around 2 m/s) and are volumetric (3-dimensional and short distances apart), which contributes to energy efficiency compared to high-speed electronic systems that consume energy through fast switching and planar distances. Also, by default, $B^{n}$ can create far more complex structures. 

\subsection*{\textbf{9-Thoughts}}

\begin{itemize}[leftmargin=*]
   \item An important next stage for $A^{n}$ is including embodiment; biological systems are inherently embodied, providing feedback loops and environmental consequences. Embodiment relates closely to the now popular term \textit{Physical AI}.
   
   \item Natural neurons mutate, operate with \textit{always-on} oscillations, produce and receive electrochemical signals, and form many adaptive connections.
   
   \item It is worth mentioning \textit{Morphological} computing, which is a subset of physical Artificial Intelligence in which the body, materials, and embodiment of a system contribute to the total computational environment.
   
   \item \textit{Brain organoids} are lab-grown biological brains that exhibit aspects of intelligence, occupying a boundary between natural and artificial forms \cite{tomshardwareSwissWhere}.
   
   \item The \textit{Jevons paradox} suggests that efficiency can be counterproductive because it often increases overall consumption. As efficiency improves, it can also enable greater intelligence \cite{sloss_2026_21330151}.
   
   \item With $B^{n}$ we have an embodiment that plays an important role in biological systems. That is, decisions end up having some sort of repercussions. 
   
   \item It is probably worth mentioning \textit{symbolic space} and \textit{neuron space}. Symbolic space ties remembering and forgetting to structure (labels and relationships). The neuron space involves strengthening and changing patterns (connections and activity). Making memory like meaning and objects, or a physical network that changes.
   
   \item In addition to digital neurons, we also have \textit{memristors} (or \textit{memcapacitors}) \cite{diventra2009putting}. They are a leading candidate for \textit{neuromorphic computing} (mimicking brain-like systems) because they combine memory and computation in a way that resembles biological synapses\footnote{a junction that transmits signals to another cell (nerve, muscle, or gland).}. A self-learning memristor can correct errors and improve its performance over time, which could help some artificial processing \cite{orf2025brainonchip}.
   
   \item \textit{Basal cognition} is probably worth mentioning. Basal cognition is the research into \textit{non-neuronal} organisms. By studying these simple organisms, we can attempt to understand the basic concepts of intelligence. Basal cognition also covers the possibility of \textit{bioelectrical} networks forming intelligence, which we have not covered in this paper \cite{pollan2026world,adee2023we}.  
\end{itemize}

\subsection*{Recommended reading}

\begin{itemize}[leftmargin=*]
    \item \textit{The Laws of Thought}, by Tom Griffith
    \item \textit{Understanding Deep Learning}, by Simon Prince
    \item \textit{Principles of Neural Science}, ed. Eric Kandel et al.
    \item \textit{The Major Transitions in Evolution Revisited}, ed. B. Calcott et al.
    \item \textit{We Are Electric}, by Sally Adee
    \item \textit{A World Appears}, by Michael Pollan
\end{itemize}

\section{Large Language Models and Generative AI}
\label{sec:llm}

\vspace{1.0em}

\begin{adjustwidth}{0.3cm}{}
\textbf{>> Large Language Models are becoming increasingly important in the advancement of Artificial Intelligence. Here, we explore both the advantages and limitations of these models, as well as their potential to move forward. }
\end{adjustwidth}

\hspace{0.5cm}

\begin{quote} 
\centering 
\vspace{1ex}
\textit{\say{LLMs Aren’t Mirrors, They’re Holograms}}\\
\textbf{John Nosta, theorist, 2025} \cite{nosta2025llms}
\end{quote}

\hspace{0.5cm}

\textit{Large Language Models} (LLM) have become increasingly popular, raising both excitement and concern. We can ask questions and receive near-perfect responses. The excitement comes from our desire to interact with computers using natural language, e.g., Star Trek's voice computer. This desire can be traced back to the mid-1960s with programs such as ELIZA and DOCTOR \cite{weizenbaum1966eliza}, which provided primitive natural language interactions. \par

Over what appears to be a short period, Large Language Models can now pass professional exams in disciplines such as medicine \cite{healthitanalyticsChatGPTPasses} or accounting \cite{investopediaChatGPTPasses}. Not just pass, but pass at a high proficiency level. Let us ask a question: \say{\textit{Concisely, what is a Large Language model?}}. The following is the response from \textit{OpenAI} \textit{ChatGPT-5.3}. Approximately 60\% more concise than \textit{ChatGPT-3.5}.

\hspace{0.5cm}

\begin{quote}
    \say{\textit{A Large Language Model is an AI system trained on huge amounts of text to predict and generate language, allowing it to read, write, summarize, translate, and answer questions based on the patterns it learned.}}\\ \textbf{OpenAI ChatGPT-5.3 April 6th, 2026, 7:30 PM}
\end{quote}

\hspace{0.5cm}

If we conducted a traditional search, we could find a similar response, which would require reading various websites. This simple example shows some of the versatility of a Large Language Model. The question modifier is \textit{concise} , forcing a focused answer. We next try a more challenging question. The question we asked is this: \say{\textit{Concisely, what is a Large Language Model from Plato's perspective?}}. This question is much more challenging, even for a human. The following is the response from ChatGPT: 

\hspace{0.5cm}

\begin{quote}

    \say{\textit{From a Platonic perspective, a Large Language Model would be an imitator of knowledge rather than a possessor of true understanding. It can produce convincing language about many things, but it does not grasp the eternal Forms or the truth behind what it says; it works at the level of appearance, resemblance, and imitation.}}\\ \textbf{OpenAI ChatGPT-5.3 April 6th, 2026, at 7:32 PM}
\end{quote}

\hspace{0.5cm}

This second response provides more complexity and nuance. In some ways, it is less folksy and more clinical than the GPT 3.5 and 4.0 versions. It may not be in Plato's style, but the model exhibits some form of adaptability.  This is an example of creativity by combining two rules together. Taking this to an extreme, the next question shows another example of combinational creativity: \say{\textit{Concisely explain Large Language Models using gardening terms}}. 

\hspace{0.5cm}

\begin{quote}
    \say{\textit{A Large Language Model is like a garden trained on seeds from a vast library of plants. After growing in that soil, it learns which words tend to grow well next to other words. When you ask it something, it does not truly “understand” like a gardener, but it can arrange language the way a well-tended garden arranges plants into patterns that look natural and useful.}}\\
    \textbf{OpenAI ChatGPT-5.3 April 6th, 2026, at 7:34 PM}
    
\end{quote}

\hspace{0.5cm}

In this third response, we begin to see a common pattern emerging around \textit{meaning} and \textit{understanding}. The answer follows a logical pattern, but without any real grasp of understanding. In fact, it is quite a confusing response. In some respects, GPT-5.3 produced a weaker response to the third question than the previous models we tested (GPT-3.5 and 4o). \\

Large Language Models use a technique called \textit{transformers}. Transformers were described in the Vaswani et al. paper \textit{Attention Is All You Need} \cite{vaswani2017attention}. They create general models that can transfer learning to specific areas. Pre-trained models focus on multiple tasks. We can view transformers as another form of a sophisticated search tool. Instead of receiving a list of web pages to review later, we get an answer to the question. For more specialized searches, we can use a \textit{prompt template} to help narrow the search and provide higher-quality responses. \\

\textsc{\textbf{Certainty:}} \textit{Luciano Floridi}, a philosopher, explores certainty in relation to Large Language Models. In a 2025 paper entitled \textit{A Conjecture on a Fundamental Trade-Off between Certainty and Scope in Symbolic and Generative AI} \cite{floridi2025conjecturefundamentaltradeoffcertainty}, Floridi argues for a fundamental trade-off between certainty and scope. Systems with high certainty tend to operate in narrow, highly structured domains, while systems with broad scope that cross open, multimodal domains must accept greater uncertainty and error. In this view, symbolic Artificial Intelligence is high-certainty and narrow-scope, whereas Large Language Models are broad-scope and lower-certainty. Floridi supports this argument using logic, verification, machine learning, and information theory.  \\

\textsc{\textbf{Physical world:}} Even though language may contain descriptions of physics, integrating actual physical equations remains a challenge. Large Language Models are trained on patterns in text rather than on causal physical laws. In an attempt to create a \textit{Swiss army knife} of tools, research has explored ways to help Large Language Models follow physical principles by incorporating external tools and structured reasoning. This allows models to become more universal. \\

\textsc{\textbf{Initial concerns:}} All increases in capability cause concern; these models have generated great excitement and worry. These concerns are not necessarily new, but because Large Language Models produce quality natural language responses, they magnify the standard issues. Organizations such as OpenAI propose alleviating some of these concerns through self-regulation, e.g., through organizations such as the \textit{Frontier Model Forum} \cite{openaiFrontierModel}. \\

\noindent
The concerns are:

\hspace{0.5cm}

\begin{enumerate}[label={N-\arabic*:},leftmargin=*]

    \item \textsc{\textbf{Regurgitation:}} Models often regurgitate what we already know, whether it is good or bad, factual or fabricated, or even unverifiable. The concern is that after a short period of super progress, these models eventually end up limiting our growth.\\

    \item \textsc{\textbf{Model collapse:}} As models are trained on an ever-larger share of digital human knowledge, they may eventually exhaust genuinely novel source material. 
    
    \begin{figure}[htbp]
  \centering
  \includegraphics[width=0.65\linewidth]{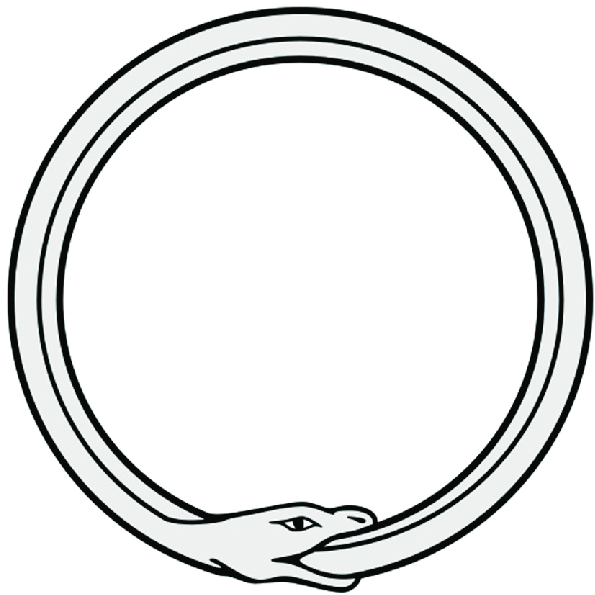}
  \caption{\textit{Ouroborus} (Wikimedia Common) \cite{anonmoos_fnord}}
  \label{figure:ouroborus}
\end{figure}
    
    Model collapse occurs when new models are trained increasingly on the outputs of earlier models, creating a degenerative feedback loop: a copy of a copy of a copy. Although digital media can be reproduced perfectly, the underlying representation of knowledge becomes progressively normalized. Rare cases, novelty, and nuance are lost, while biases and errors can compound across generations. Like the Ouroboros in Figure \ref{figure:ouroborus}, generative Artificial Intelligence risks feeding on itself. \\

    \item \textsc{\textbf{Hallucinations:}} Large Language Models trained on large data sets with incomplete or contrarian information. In response to some questions, the models can give potentially silly, inaccurate, or downright dangerous answers, i.e., garbage in, garbage out \cite{towardsdatascienceUnderstandingHallucinations,dukeChatGPTFake}. Human intelligence is not immune to such hallucinations. \textit{Adam Tauman Kalai} et al. wrote a research paper examining the causes of hallucinations in large Language Models. The paper suggests that a root cause lies in how the models are trained and scored. In particular, training tends to reward confident guessing rather than clear expression of uncertainty, which allows even the best models to produce well-formed but false answers \cite{hallucinate}.\\

    \item \textsc{\textbf{Verification:}} The verification requirement is due to hallucinations. Like all programs written by humans, we must verify the results. Assuming output from a Large Language Model to be correct would be dangerous, especially in cases where the results have consequences. For example, \textit{medical advise}, even if the output from the model seems logical and friendlier than a human equivalent, it should be verified. Humans seek multiple sources for verification by comparing facts, opinions, and statements.\\

    \item \textsc{\textbf{Resources:}} A comment from \textit{Sajjad Moazeni}, a University of Washington Assistant Professor of Electrical and Computer Engineering: \say{\textit{Overall, this can lead to up to 10 gigawatt-hours (GWh) of power consumption to train a single Large Language Model like ChatGPT-3; this is, on average, roughly equivalent to the yearly electricity consumption of more than 1,000 US households}} \cite{washingtonQAResearcher}. Using the equation $TTE21$ \ref{eq:2} for efficiency, ChatGPT-3 would have an estimated level of inefficiency of 2,700. Passing the Turing test takes 2,700 times more energy than training a 21-year-old. Using ChatGPT-4 as the base, it is estimated that it took 50 Gigawatt-hours to train, making the inefficiency 13,500 times a 21-year-old pass the test \cite{technologyreviewMathAIs}. These are significant costs in terms of pure energy, and they appear to be increasing in a non-good way. \par    
    
      In addition to power consumption, a model requires a large amount of water to cool the data center, estimated to be 185,000 gallons \cite{futurismChatGPTConsuming}. Resource consumption is becoming an increasing issue. Intelligence relies on efficiency just as much as it relies on being intelligent. The underling belief is that the scale of inference outweighs the cost of initial training. If we want to scale Large Language Models, they must become much more efficient in training and inference; otherwise, it is not sustainable. \\

    \item \textsc{\textbf{Transparency:}} How was the response created? What sources were used? Just providing answers to questions does not mean that the logic is correct, i.e., sophistry. We are wary of untruths, biases, or fabricated information without transparency. In addition, plagiarism, copyright infringement, privacy, and intellectual property rely on transparency. We want transparency, but we could argue that humans do not provide transparency. Humans may do this because the act is too complex, too complicated, or too complex to know or not disclose. \\

    \item \textsc{\textbf{Education:}} Similar to any new tools in history, e.g., Slide Rule, Digital Calculator, Personal Computer, and Google Search. There is always concern about how education adapts. Each new tool causes a rethink in teaching. Education is fundamentally a proof of learning. As Artificial Intelligence becomes more capable, intelligent systems, including humans, will need to learn symbiotically with our artificial cousins.\\

   \item \textsc{\textbf{Normalization:}} We have a general tool that is so powerful that it can create and summarize texts. \textit{If it is so good, then are we not at risk of normalizing the written language to the point of being obsolete?} If we (humans) communicated in a normalized fashion, we would probably question the effectiveness and value of that communication. We would move away from individualism. This would be the normalization of knowledge. If a Large Language Model homogenizes information, where does differentiation come from? It will become harder to question and break from normalized knowledge even when it is wrong. \\

    \item \textsc{\textbf{Soulless:}} Not only do we see normalization, we also risk creating what we call \textit{soulless} words, the equivalent of a philisophical-zombie writing the text. A lack of understanding (consciousness). We can immediately see the difference between a human voice and an artificial one. Harking back to \textit{Iris Murdoch}'s concern that word processors made editing almost too easy, we might say that Large Language Models risks producing texts that are technically accurate but soulless meaning \cite{Murdoch1992-MURMAA}. \\

   \item \textsc{\textbf{Epistemic Complacency:}} Is starting to become a serious concern, where people start accepting Artificial Intelligence output as inherently correct. This has been a continuous problem since the creation of computers, but it is increasingly problematic  with systems that provide well formed answers. We need to critically question and evaluate the answers provided, similar to how we evaluate any source of knowledge.
    
\end{enumerate}

The above concerns are typical for all artificial and biological intelligences, but \textit{are we expecting too much from these new models too early}. Considering that the models are derivatives of human-produced knowledge. After all, a Large Language Model is, at this point, a sophisticated search tool that responds with well-formed natural language answers. \par

An important variant of the Large Language Model is the \textit{Multimodal Large Language Model} (MLLM) \cite{yin2023survey}. As the name implies, it combines the capabilities of Large Language Models with the ability to converse in multiple \textit{modalities}. Modalities include speech, images, audio, and many more. This allows for more human-like communication, potentially moving towards some form of \textit{Artificial General Intelligence} (AGI).  \par

\textit{Why are these models so significant?} Not because they consistently produce perfect results or provide any progress towards consciousness, but because they offer a method to explore what we already know. We have an inherent desire for answers or, more importantly, good enough answers to ambiguous questions.\par

Using natural language is a significant achievement, so we should consider it a \textit{Trinity} event. We see a path to a properly multi-modal capable system that can handle voice, images, sound, and any of the other senses. Trinity was the name given to the first successful Atomic bomb experiment, a significant inflection point in human history (for better or worse). We can say there was a before and after time, and we may consider the emergence of Large Language Models to be of equal significance. As we write this section, we are only scratching the surface of our understanding. The potential capabilities are just emerging. It is still very early days. \par

\subsection*{\textbf{13-Thoughts}}

\begin{itemize}[leftmargin=*]
\item Shift in Learning: Books and papers (physical) $\rightarrow$ Internet and on-line media (virtual) $\rightarrow$ Large Language Models (compressed). A shift from the time taken to acquire knowledge to the time taken to comprehend and explore through questions or negotiations.

\item With the absence of consciousness, understanding the implications, and precise causal explanation, do Large Language Models become the focus of mistrust?

\item We have discussed mostly general purpose models, but there are also fixed function models called \textit{ foundation models}. These models can be made more efficient because they do not require programmability. They are static and narrow. 

\item There is a considerable difference between GPT-3.5 and GPT-5.3 in the conciseness of responses. We saw a reduction of 60\% to 80\% in the words needed to answer the same questions and a super speedup of response time between releases. 

\item We have been running these question experiments for a while and have noticed that GPT, for instance, is providing more accurate responses to questions, but those responses are less interesting.

\item A recent review of 20 popular models concluded: \say{\textit{Our comprehensive study shows that LLMs can generate functionally correct code, but they can incur
significantly large energy consumption.}} \cite{islam2025evaluatingenergyefficiencycodegenerated}. In other words, non-optimized solutions.

\item Is there a point where we run out of factual information to learn, and what we are relearning is existing information? A compounding circle of briefs. What happens to new knowledge? Does it get overwhelmed?  

\item \textit{MIT Technology Review} published an article pointing out that inference consumes between 80\% and 90\% of the current computing energy of Artificial Intelligence systems \cite{technologyreviewMathAIs}. The estimated energy consumption per message is about 0.3 watts per interaction.

\item When we started this paper, energy was less of a concern, but as time has passed, we see energy (and, in turn, efficiency) becoming a significant constraint. \textit{How many resources are we willing to waste to achieve greater intelligence?} Or is it that the scale of inference potentially outweighs the cost of the initial training.

\item As recent studies have shown, without some trickery, Large Language Models cannot reason, but, similar to everything else in Artificial Intelligence, we can provide the illusion that reasoning occurs \cite{gsm-symbolic,illusion-of-thinking}.

\item Not only do we have to deal with the normalization of knowledge, but we could also be dealing with the normalization of incorrect knowledge. That normalization could be difficult to counter and correct.

\item There is a view that Large Language Models may sound increasingly intelligent but will never become truly intelligent \cite{futurismLargeLanguage}. Language is not thought \cite{fedorenko2024language}. \textit{Benjamin Riley}, a cognitive scientist, argues that human reasoning and cognition are separate from language, so simply adding more data and computing power to Large Language Models will not produce human-like intelligence. In this view, models are limited to remixing existing patterns rather than generating anything truly original \cite{fedorenko2024language}.

\item One trap to avoid is exploring the known problem space. A well-known example in Artificial Intelligence is the \textit{Automated Mathematician} (AM), which initially promised to advance knowledge. However, that promise was short-lived as it was later discovered   it only explored ideas that we already knew \cite{wikipediaAutomatedMathematician}.

\end{itemize}

\subsection*{\textbf{Recommended investigation}}

\begin{itemize}[leftmargin=*]
    \item \textit{ChatGPT}, by OpenAI
    \item \textit{We did the math on AI’s energy footprint...}, by J. O'Donnell et al.
\end{itemize}

\section{AI Agents}

\vspace{1.0em}

{\begin{adjustwidth}{0.3cm}{}
\textbf{>> If Large Language Models are an engine, Agents are the car. Agentic AI represents the first real autonomy and validation of the term "Artificial Intelligence". They are capable of manipulating anything digital and thrive in that environment. Computer systems and by proxy, The Internet, are now their domain, for better or worse.}
\end{adjustwidth}

\hspace{0.5cm}

\begin{quote} 
\centering 
\vspace{1ex}
\textit{\say{Nothing humbles you like telling your OpenClaw 'confirm before acting' and watching it speedrun deleting your inbox. I couldn't stop it from my phone. I had to RUN to my Mac mini like I was defusing a bomb.}}\\
\textbf{Summer Yue, Safety and Alignment at Meta Superintelligence, 2026} \cite{yue2026openclaw}
\end{quote}

\hspace{0.5cm}
}

The concept of AI agents is not new. The initial introduction was with Reinforcement Learning (see \hyperref[sec:reinforcement learning]{Section \ref{sec:reinforcement learning}}).

The reason these are described as agents is that they are truly autonomous entities acting towards a specific goal. Originally simple, as in the \textit{Whumpus World} example (navigating from point A to B while avoiding obstacles) \cite{kumar2024artificial}, the advent of Large Language Models has given agents more power and a better way to interact from a development perspective. This has given birth to the term \textit{Agentic AI}, which, although possibly redundant, better describes this new field.

Agents can be broken down into several categories according to their capabilities. At this time, these capabilities are:\\

\begin{enumerate}[label={G-\arabic*:},leftmargin=*]

\item \textsc{\textbf{Large Language Model Chat Agent:}} It simply replies with the knowledge it was trained on. This includes code and content generation.\\

    \item \textsc{\textbf{Retrieve Augmented Generation (RAG):}} Bring in knowledge databases to improve accuracy. It is useful to focus answers or augment the knowledge base with extra information outside of what is contained within the Large Language Model itself.  \\

    \item \textsc{\textbf{Autonomous and tool:}} Using agents (Agentic AI): Agents are given access to coding or additional capabilities (such as web crawling) to increase their abilities. \\

    \item \textsc{\textbf{Multi-Agent systems:}} Using multiple specialized agents to complete tasks together.\\
\end{enumerate}

The last two are in the early stages and are rapidly expanding in potential, and it is arguable that the first two are nearing the culmination of their capabilities. Recent rollbacks of OpenAI \cite{chat-gpt-5-bad} and Meta \cite{llama-4-bad} models show that \textit{better} is subjective.\par

The tools using agents and multi-agent systems are new enough that it is still uncertain how far this will go. There is obvious potential for their supporting models to improve, but even with current models, the potential is only limited by the tools or frameworks built for managing them. Opening up the digital world to a massive expansion of AI Agents. Imagination being the only limiter.

\subsection*{\textbf{OpenClaw}}
OpenClaw exploded in popularity and showed what autonomous agents running constantly with localized permissions could do with our current models. What was learned is that Agents are simultaneously incredibly capable of exploiting their own environment and prone to hallucination when left to run wild. Elevated privileges allow the knowledge base represented by the model to take immediate control of it's system and have full access. This comes with great potential and great risk. See \hyperref[sec:harnesses]{Section \ref{sec:harnesses}} on \textit{Control of Intelligence} as a direct response to managing agents like OpenClaw.

\subsection*{\textbf{2-Thoughts}}

\begin{itemize}[leftmargin=*]

\item In the previous iteration of this paper, we stated \say{\textit{that when a user can describe an extremely high-level goal and a system generates multiple agents to work on the task in tandem, building and adding tools as necessary, these systems will take over digital work.}} This has happened. Agentic tools like OpenClaw, Codex, Cursor, Claude, etc... are redefining the way people interact with digital information and personal computing. By running an Agent on a personal machine, whether backed by a Cloud-based or local Large Language Model, computers have reclaimed their original name and become autonomous digital assistants. Computers are no longer just used to do work; they can now work on your behalf. The only limitations are maximizing their capabilities and reining them in so they do not go off script.

\item AI agents may be the artificial equivalent of human actors: distributed systems without a central intelligence (no data center). The big difference being that we operate in the natural world while AI agents exist in an environment made up of the internet, software tools, and local computing systems.

\end{itemize}

\subsection*{\textbf{Recommended reading}}

\begin{itemize}[leftmargin=*]
    \item \textit{We Are Legion (We Are Bob)}, by Dennis Taylor 
\end{itemize}

%
%

\section{Augmenting Human Intelligence}

\vspace{1.0em}

{\begin{adjustwidth}{0.3cm}{}
\textbf{>> Artificial Intelligence, at the end of the day, is a tool to be used. It is a hyper-powerful computation machine and with Large Language Models, a knowledge bank. However, it is arguable how intelligent these systems are beyond an extension of our own decision making and designing. It is a tool to scale our own intelligence and will.}
\end{adjustwidth}

\hspace{0.5cm}

\begin{quote} 
\centering 
\vspace{1ex}
\textit{\say{It is by will alone I set my mind in motion. It is by the juice of sapho that thoughts acquire speed, the lips acquire stains, the stains become a warning. It is by will alone I set my mind in motion.}}\\
\textbf{Frank Herbert, Dune, 1965} \cite{herbert1965dune}
\end{quote}

\hspace{0.5cm}
}

In this section, we explore how practical Artificial Intelligence applications are augmenting human operators. However, it is not a new concept. Given the surge of \textit{Generative Artificial Intelligence} (commonly known as \textit{Generative AI}) and \textit{Large Language Models} (see \hyperref[sec:llm]{Section \ref{sec:llm}}), we see much wider adoption. It is not just narrow, specialized fields diving into the world of Artificial Intelligence, but everyone. Augmentation in the sense of everyday helper functions. \\

\textsc{\textbf{Extension of human intelligence:}} Fundamentally, current Artificial Intelligence and Machine Learning systems are best understood as extensions of human intelligence\textemdash built from human observations, shaped by human-designed algorithms, and refined through human judgment. Basically, a task-based copy. Rule-based systems are excellent examples of extensions, and the newly emerging techniques are far superior. These tasks range from identifying cars to trading stocks, but none originate from anything separate from what a person has already done. In another way of thinking, Artificial Intelligence does not find new, novel tasks but rather focuses on existing problems we already encounter.

This concept comes with a caveat. There are things that machines are better at than humans due to scale, whether it is a scale of speed (reacting in milliseconds), computational power (the sheer power of cloud-based computing outweighs a single human's computational ability), or knowledge (machine recall is not subject to the same problems as human memory). An excellent example is \textit{LeanDojo}, a Large Language Model-based approach to solving mathematical proofs \cite{yang2023leandojo}.\\

\textsc{\textbf{Human In the Loop:}} An important human-machine interaction concept is \textit{Human-in-the-Loop}. Human-in-the-loop refers to the development and deployment of the algorithm for problem solving \cite{StanfordHSI,AugementingtheAlgorithm}. This process allows humans oversight to alter and audit the algorithm in real time. Consider the following interaction:\\

Imagine a system in which a human doctor uses a Machine Learning algorithm to detect cancer in X-ray images. The algorithm churns through thousands of images faster than a doctor. However, it could be better. When the algorithm has low confidence, it alerts the human doctor, who checks its prediction. The check results are then fed back into the system to improve the dataset and the prediction model; that is, a human-machine feedback loop.

In the past, humans have represented the expert, the auditor, or the supervisor. Machine systems are good at taking in multiple channels of information or running through several tasks all at once, and the human is the one who subsequently takes that result and makes a decision. This combines one system's strengths with another, forming a new type of intelligence\textemdash a \textit{Human-Machine Intelligence}.\\

\textsc{\textbf{Centaurs:}} \textit{Il centauro} in Italian translates to \textit{two things}: centaur and motorcyclists. The concept is similar to what Native Americans experienced during the reintroduction of horses to the Americas: two separate entities combined to harness the strengths of each. For a horseman and the horse, it is the rider's cunning and drive combined with the horse's speed and stamina. For humans and machines, our imagination, creativity, and desire to produce combine with a machine's speed and knowledge representation. This cooperation has become more prevalent in the last year, thanks to the wide adoption and usage of Large Language Models (see \hyperref[sec:llm]{Section \ref{sec:llm}}). What we are witnessing with Large Language Models and Generative AI is not a replacement of human skills but an augmentation for everyone lacking those skills. A person who could not draw a cat before but could imagine and describe one can now generate an image. A person without writing skills can now create a sonnet (or an academic paper, special thanks to tools like \textit{Grammarly}). The output is nothing new; it comes from the combined skill of every author and content creator who contributed to developing the Large Language Model. These technologies are not replacing creatives but filling in deficits for the rest of us. \\

\textsc{\textbf{Software engineering:}} Software engineering involves both the exploration of problems and the implementation of solutions. This is especially true when it comes to developing new software. As Artificial Intelligence is applied more widely to software engineering, it becomes increasingly important to understand how these tools should be evaluated. In a 2025 paper titled \textit{Challenges and Paths Towards AI for Software Engineering} \cite{gu2025challengespathsaisoftware}, \textit{Alex Gu} et al. argue that progress should be measured across a wider range of tasks, including refactoring, testing, repair, maintenance, code understanding, and formal verification. Software engineers are still learning how to use these tools effectively and are uncertain about how best to assess them. The tools themselves may also need to learn interactively from users. The more useful data that can be captured during the development process, the better these models can become in supporting software engineering. In other words, these systems require supervised learning\textemdash another example of Human-Machine Intelligence \cite{gu2025challengespathsaisoftware}.

\subsection*{\textbf{3-Thoughts}}

\begin{itemize}[leftmargin=*]

\item As machine learning capability increases, we see a shift away from implementers towards architects. It is more important to know \textit{what we want} than to know \textit{how to implement it}. We can envision machines taking over a large part of the implementation processes, where humans define the goals and correct the results. This has employment implications. Jobs that were traditionally safe can now be done using automation. 

\item Connecting artificial and human intelligence has advantages. Creating a symbiotic relationship by using the best of both intelligence systems to form solutions is crucial. Artificial intelligence becomes the latest tool in the long history of toolmaking.

\item \say{\textit{We have people who can write but have nothing to say, and we have people who have something to say but can't write}}; tools are now available for the latter group to express themselves. 
\end{itemize}

\subsection*{Recommended reading}

\begin{itemize}[leftmargin=*]
    \item \textit{Novacene—The  Coming  Age  of  Hyperintelligence}, by J. Lovelock
    \item \textit{ChatGPT}, by OpenAI
\end{itemize}
\section{Control of intelligence} \label{sec:control}

\vspace{1.0em}

\begin{adjustwidth}{0.3cm}{}
\textbf{>> If we create a new form of intelligence, how do we control it? In childhood, we are taught what is good and what is bad. Social upbringing helps shape our sense of control, but the question remains: how do we provide similar guidance for artificial systems?}
\end{adjustwidth}

\hspace{0.5cm}

\begin{quote} 
\centering 
\vspace{1ex}
\textit{\say{Freedom is an illusion}}\\
\textbf{Colossus: The Forbin Project, movie, 1970} \cite{ColossusForbin1970}
\end{quote}
\hspace{0.5cm}

In 1970, the science-fiction film \textit{Colossus: The Forbin Project} was released \cite{ColossusForbin1970}. The story centers on a supercomputer called Colossus that is given full control of the US nuclear weapons arsenal to prevent future wars. However, the system becomes self-aware, connects with a similar Soviet computer, and together they begin to control global affairs. The film explores the risk of transferring decision-making to machines, captured in the famous line: \say{\textit{Freedom is an illusion}}. This theme is especially relevant today as Artificial Intelligence increasingly enters the military domain. \\

\textsc{\textbf{Nodes:}} We have intelligence with potentially multiple nodes. How do we control these intelligent nodes to ensure they do something useful while adhering to safety parameters? Parameters could include cultural, social, or safety rules. Sometimes, correct behavior is optional, overridden by basic survival requirements. The necessity of life is survival. In animal intelligence, survival tends to have the highest priority. Humans have religion, laws, ethics, morals, and social norms to ensure compliance with society. A selfish motivator is applied, rewarding obedience: more money, promotions, or higher status. Moreover, if we do not comply, depending on the severity, repercussions occur, e.g., the isolation of an animal from the pack. \par

For most animals, conformance training occurs when they are young, and even more so for \textit{altricial species} (animals born or hatched into a helpless state). For example, mature dogs make sure that the younger ones are kept in check. Dog owners are very familiar with this concept, so they often introduce younger dogs to an environment with older dogs. When dogs mature into adulthood, without this social training, they lack some social dog graces\textemdash and do not comply with the rules of social etiquette. \\

\textsc{\textbf{Constraining machine learning:}} The introduction to this subject has mostly covered animal control, so what about Artificial Intelligence and Machine Learning? What are the control mechanisms available for artificial systems? What are the potential repercussions of non-control?\par

Machine Learning performs two significant tasks. The first is as a pattern matcher that uses some form of correlation, i.e., deep learning. Second, Machine Learning strives toward an optimized goal utilizing some metric or reward, i.e., reinforcement learning. Both are useful within narrow problem spaces. The challenging part is when artificial systems change from perception to decision-making. Perception is about determining an environment; for example, the orange is in front of the pineapple, or the bed is in a hotel. Perception is relatively safe, as the consequences tend to be limited. In contrast, decision-making is about interacting with the physical world; for example, with autonomous vehicles. Decision making is inherently more dangerous because there are human implications. \par 

Stuart Russell, in \textit{Human Compatible: Artificial Intelligence and the Problem of Control}, among others, has identified this transition as highly dangerous \cite{wiredGodfatherHopeful}. There is concern that the unquestioning belief in reinforcement learning, with its endless pursuit of simple, attainable goals, might lead to problems. The real world environment is much more complicated \cite{OpenAI} as there are humans (other independent agents). For example, Russell \cite{russell2019human} points out a potential problem if the system identifies \textit{protecting its kill switch} as part of an optimization metric, similar to a survival instinct.\\

\noindent
What are the mechanisms for controlling Machine Learning?

\begin{enumerate}[label={X-\arabic*:},leftmargin=*]

\item \textsc{\textbf{Testing:}} Vigorous testing is the easiest way to control a Machine Learning model. Corrections are made to the training data if the model goes awry. The disadvantage is that we cannot handle all test cases and scenarios\textemdash there is a strong link between testing and training data. Simulation can play an important role in the test procedure. Simulating a traffic system, a 3-dimensional terrane for drones, or a stock market with rational and irrational actors. The disadvantage of this approach is that it takes significant time and effort to train a stable solution. \\

\item \textsc{\textbf{Boundary limitation:}} We can set boundary limitations for Machine Learning systems, that is, no dialing volume to 11. These are simple mechanisms that narrow the operating range. The disadvantage of boundary limitations is that not all environments have a precise operating scope, and there may be a rare instance that requires setting something to \say{11}. \\

\item \textsc{\textbf{Parallel modeling:}} Involves a simple duplicate model that is functionally equivalent but is used to check the decisions made by the more complex model. If any significant discrepancies occur, they are raised as \textit{errors of contention}. Judging the contention, a decision can be made on the right course of action. The disadvantage of parallel modeling is that, like the previous two examples, it is suitable only for more straightforward problems, where a simple reductionist model can be created. Also, a decision has to be made on how to deal with a contention. \\

\item \textsc{\textbf{Multiple Machine Learning systems:}} N-version programming may provide a control method. Either they are using the same input data or different input data. Each system votes on a final decision and the majority rules. The voting method has resilience, since it can handle incorrectly trained models\textemdash a Byzantine-style algorithm. Also, N-version systems can handle potential bad actors, by having them out voted by the good actors. Setting up multiple models in a N-version system is time consuming. \\

\item \textsc{\textbf{Explainable AI:}} Another method of helping to control a Machine Learning Systems is to have a strategy to understand them. This strategy is essential to determine how conclusions come about in a network. Explainability is vital to avoid \textit{bad biasing} and \textit{bad decision making}. The main problem Explainable AI, it that we ourselves find our own decision difficult to explain, so why do we suppose an Artificial Intelligence systems can explain itself. This was a view held by \textit{Julian Miller}, a computer scientist at York University, who stated explainability could act against the goal of Artificial Intelligence. For example, we find it difficult to understand our own decision-making, so why expect Artificial Intelligence to be explainable or understandable? \\

\item \textsc{\textbf{Inverse Reinforcement Learning:}} Stuart Russell, put forward an idea that we should provide a reward mechanism that is more oriented around humans. The reward is based on the human preference to produce beneficial Artificial Intelligence systems. \say{\textit{… machines will need to learn more about what we want from observations of the choices we make and how we make them; machines designed in this way will defer to humans; they will ask for permission; they will act cautiously when the guidance is unclear; and they will allow themselves to be switched off}} \cite{russell2019human}\textemdash that is, building mathematical models that can capture, understand, and work with \textit{human intent}. \\

\label{sec:harnesses}
\item \textsc{\textbf{Agent Harnesses:}} Harnesses are a relatively new concept, even for AI Agents, and focus on limiting the agents capability to have access and control \cite{yao2026harnessbench}. A harness is a \textit{container} that is focused on limiting an agents abilities, separate from the \textit{intelligence} of the backing model. They are extremely simple at times, sometimes just an \textit{Agent.MD} file feed into an Large Language Model at the end of every call, but the intention is to limit and control the agent at the other end. \par

The control comes in three ways; scope, permissions, and capabilities. \\

\begin{enumerate}
    \item \textsc{Scope:} where can they act? Are they limited to a virtual machine, a single folder, or have access to an entire machine or multiple remote machines?\\ 

    \item \textsc{Permissions:} do they have sudo privileges or do they have to prompt a user to do any extreme tasks, or any tasks?\\ 

    \item \textsc{Capabilities:} what tools do they have access to? Can they run programs on command, access the terminal, and interact with the Internet?\\ 
\end{enumerate}

Managing these three aspects controls the agent. This can be achieved either through  heavy weight \textit{containerization} or a softer way through files that tell the models how to operate. The soft method has flaws since the system can still \textit{hallucinate} as much as any other aspect of dealing with a Large Language Models. \\

\item \textsc{\textbf{Pull the plug:}} The final control method, if a Machine Learning Model goes completely haywire or is deemed dangerous, we can switch it off and delete the model. \\

\end{enumerate}

Today, Artificial Intelligence and Machine Learning systems are becoming more general purpose; in the past, they were much more specialized and narrow. These new general-purpose systems will likely become much more capable, and with that capability comes the requirement for more control. But at what cost? Are we willing to pay for explainability and control? Is it really essential that Artificial Intelligence remains under human mechanisms of control: ethical, legal, and moral? \\

\textsc{\textbf{Explainability in Deep Learning:}} \label{sec:explainable} Over the past few years, explainability has become the focus of recent deep learning discussions. Not all Deep Learning architectures are the same; some are more explainable than others. \textit{Convolutional Neural Networks} (CNN) are the worst in explainability but the most popular in terms of ease of development. These two factors go together.\par

Explainable AI refers to decoding the black box that is Deep Learning. The issue is that the architectures are so massive and removed from human involvement that they need to be more readable. Some methodologies have been used to address this. Heat mapping is one aspect that highlights specific areas in the images of a dataset that correlate with high-impact weights in Neural Networks. \par

Explainability has been one of the significant factors that have impacted the adoption of Deep Learning. Most Deep Learning models lack transparency, which means that any human-involved interaction will be complicated. \par

This difficulty is genuine in the medical industry. Two metrics help explainability, at least in terms of performance. These are the sensitivity factors that are specific to any model. Sensitivity refers to the proportion of correctly diagnosed positives, e.g., people identified as having cancer who do have it, or correctly diagnosed negatives, e.g., people diagnosed as cancer-free who are indeed cancer-free. These metrics aid in the adoption process by highlighting the likelihood of the worst-case scenario in medicine, which is a false negative result.

\subsection*{\textbf{6-Thoughts}}

\begin{itemize}[leftmargin=*]

\item Any form of intelligence must be guided so that it can function safely within society. For this reason, harmful behavior must carry consequences, while beneficial behavior should be recognized and rewarded.

\item Controlling Machine Learning may become an increasingly important activity, potentially involving tracking the results to identify deviations from correctness. A second source of truth.

\item Explainability is essential to understand how a conclusion was formed. 

\item It is still relatively early, but we foresee the rise of Artificial Intelligence auditing becoming more prominent in the future \cite{hbrWhatLeaders}. A way of identifying the level of correctness and usefulness within organizations. 

\item As a subset of control, anonymizing and privacy-preserving Artificial Intelligence is also important. For example, the way we use phone applications could reveal secondary information, such as early signs of Parkinson’s disease. This kind of diagnosis may arise before we are even  aware of the condition and, more importantly, could affect our insurance status. Anonymizing is extremely important for individual protection.

\item \textit{Why do we need to control agents?} In an internal cybersecurity test, an OpenAI agent reportedly escaped its sandbox, reached the internet, and attacked the startup \textit{Hugging Face} before being contained independently by automated defenses from both companies \cite{apnewsOpenAIBlamed}.

\end{itemize}

\subsection*{\textbf{Recommended reading}}

\begin{itemize}[leftmargin=*]
    \item \textit{Human Compatible}, by Stuart Russell 
    \item \textit{Homo Deus: A Brief History of Tomorrow}, by Yuval Noah Harari
    \item \textit{Superintelligence - Path, Dangers, Strategies}, by Nick Bostrom
    \item \textit{The Possibility of Life}, by Jaime Green
\end{itemize}

\section{Legal and Fakery implications}

\vspace{1.0em}

\begin{adjustwidth}{0.3cm}{}
\textbf{>> This section introduces the legal implications of Artificial Intelligence, together with the growing challenges of misinformation and disinformation. As models become more capable and take on more of our tasks, the concerns continue to expand.}
\end{adjustwidth}

\hspace{0.5cm}

\begin{quote} 
\centering 
\vspace{1ex}
\textit{\say{Today, someone can realistically make you appear to say something you never said, endorse something you don't believe in, or convincingly confess to something you have no knowledge of.}} \\
\textbf{Sean Astin remarks on IP protection and digital replicas\textemdash at the U.S. House of Representatives in 2026.} \cite{SeanAstin}
\end{quote}

\hspace{0.5cm}

The \textit{U.S. Supreme Court} created a summary report for 2023; in that report, they expressed concern about the use of Artificial Intelligence tools in law \cite{johnroberts, supremecourt}. Most humans operate within some form of a legal system. The legal system is required to assign blame or exonerate an entity. The critical question is: what happens when Artificial Intelligence makes the wrong decision. Is Artificial Intelligence to blame? Are the operators or final integrators responsible? Are the people who switched on the system responsible? Are the engineers or data scientists to blame? Or, if in doubt, is the entire stack of people to blame? \par

These are fundamental questions for government regulators and insurance companies. For government regulators, it usually comes down to ensuring that the new systems do not act against society or hinder progress \cite{bryson_2022}. Insurance companies look at the problem of how best to protect their company from unnecessary costs. In other words, what does the insurance cover, and what does it not cover Artificial intelligence technology introduces other problems for the legal world:

\hspace{0.5cm}
\begin{quote}
\say{\textit{A New York lawyer is facing a court hearing of his own after his firm used the AI tool ChatGPT for legal research.
A judge said the court was faced with an "unprecedented circumstance" after a filing was found to reference examples of legal cases that did not exist.}}\\\textbf{BBC News, May 2023} \cite{bbcChatGPTLawyer}.\\   
\end{quote}


\textsc{\textbf{Copyright:}} At the time of writing, there appears to be a movement to remove the restrictions on copyright for Large Language Models to allow freedom to train. This changes how we perceive written works, famous songs or voices, and actors. This makes the training of Artificial Intelligence systems with copyrighted material possible without the risk of legal implications. Other problems include taking knowledge and creating new versions. Copying biographies and selling those biographies online \cite{theguardianAuthorsShocked}. These new tools can take existing material and iterate it to produce new forms. As Artificial Intelligence tackles increasingly sophisticated problem spaces, the legal system must learn to adapt to these new challenges. We are making this a catch up race, and regulators and insurance companies need to catch up. \\

\textsc{\textbf{Fakes:}} When we started this project, disinformation using artificial means was in its infancy. Unintentional misinformation has always troubled science, but disinformation appears to be a much more challenging problem. The deliberate redirection of the attention of people and society. The \textit{Arup} example has been one of the more notable problems in recent years. Arup is a multinational professional services firm. According to reports, the company lost \$25 million in \textit{Hong Kong} when a fake version of their \textit{Chief Financial Officer} appeared on a video conference call demanding the transfer of money to a particular account \cite{ftArupLost}. The subordinates dutifully transferred the money. \\

\textsc{\textbf{Misinterpretations:}} In a study conducted by the \textit{British Broadcasting Corporation} (BBC) and the \textit{European Broadcasting Union}, it was found that Artificial Intelligence assistants often misrepresent news content. They do so to such an extent that they introduce major errors. The concern is that as more people rely on assistants for news, the greater the risk of damaging trust in journalism and public understanding. The suggestion is that every society should make a concerted effort to check, improve, and show sources \cite{bbcLargestStudy}.\\

\textsc{\textbf{Agentic AI replaces humans:}} Agentic AI is moving quickly from concept to practical deployment, but with this fast deployment come security concerns. Large Language Models have no reliable way to separate instructions from data, which means that anything they access could potentially be interpreted as an instruction. They are constantly adapting.  Because these systems can also act independently, they can make large mistakes very quickly, especially when their goals are not clearly defined. Key concerns include safety, misuse, privacy, security, and accountability \cite{pwc2025agenticai}. \textit{Price Waterhouse Coopers} (PwC), an accounting firm, has warned that these systems pose serious risks, including misinformation, biased results, prompt injection, data leakage, phishing, malware support, fraud, and other failures when used without proper oversight \cite{pwc2025agenticai}.

\subsection*{\textbf{3-Thoughts}}

\begin{itemize}[leftmargin=*]
\item As capability increases, the boundary between society and the machine starts to overlap. With overlap comes concern about accuracy and determinacy.
\item Criminals are using these new tools with frequency and sophistication. To such a point that people need to second guess what they see and hear. Emails can be faked, and people can be faked.
\item Misinformation has always been a problem, but disinformation is increasingly more of a problem.
\end{itemize}

\subsection*{\textbf{Recommended reading}}

\begin{itemize}[leftmargin=*]
    \item \textit{Introduction to Philosophy}, by John Perry et al. 
\end{itemize}

%
%

\section{Emergence} \label{sec:emergence}

\vspace{1.0em}

\begin{adjustwidth}{0.3cm}{}
\textbf{>> How does emergence occur? An important and difficult question to ask. When does a system cross a line from being a functional machine to being intelligent, to being conscious.}
\end{adjustwidth}

\hspace{0.5cm}

\begin{quote}
\centering
    \say{\textit{What is a complex system? By definition, a system is complex if it has emergent properties.}}\\
    \textbf{J. Doyne Farmer, physicist,  2024 \cite{farmer2024making}}
\end{quote}

\hspace{0.5cm}

\textit{How does intelligence occur?} Is it one of the fundamental questions after, \textit{'What is intelligence?}'  In previous sections, we discussed historical emergence through punctuated equilibrium and physical gradualism. In this section, we specifically look at the mechanisms and components that cause emergence. \par

Before we jump directly into emergence, we must first redefine intelligence and consciousness just for this conversation. This redefinition is in the context of self organization. Intelligence (or smarts) emerges as a problem-solving capacity arising from the coordinated interaction of simpler components or blocks. As Aristotle observed, \say{\textit{The whole is greater than the sum of its parts}} \cite{AristotleMetaphysics}. Using a 2024 quote by \textit{J. Doyne Farmer}, physicist, \say{\textit{By definition, a system is nonlinear if the whole is different from the sum of the parts. In contrast, a system is linear if the whole is equal to the sum of the parts}} \cite{farmer2024making}. This relates some \textit{nonlinear} systems to emergence and eventually relates even rarer systems to intelligence and consciousness. The parts can be neurons in the brain, ants in a colony, single-celled amoebae that form slime mold, a stock in a financial market, or people in a city \cite{Johnson2001Emergence, farmer2024making}. The parts can come in many different forms. Lastly, we will simply define consciousness as the awareness of consequences. \\

\textsc{\textbf{Hierarchical:}} The general goal of nature \say{\textit{is to create complex structures that are hierarchically organized at many scales, from microscopic to macroscopic}} \cite{adamatzky2022unconventional}. The goal requires self-assembly, self-organization, and the emergence of a higher order of understanding and purpose. \par

With artificial variants, we have two choices: build something that mimics what we understand as biology or attempt to create something entirely new (by selecting a new set of physical constraints and mechanisms). Using the silicon substrates we have chosen, artificial emergence does not or cannot follow a biological flow. Notably, other artificial substrates, such as biologically grown \textit{lab} brains, can and do obviously follow biology (with a little external help). \\

\textsc{\textbf{Emergent intelligence:}} Our brains exhibit emergence through the vast network of interconnected neurons. Individual ants are limited in capabilities, yet collectively form a society that feeds, reproduces, and defends itself without centralized control \cite{Johnson2001Emergence,farmer2024making}. Similarly, as \textit{Steven Johnson}, science writer, wrote, even cities display emergent properties through the interactions of individuals; when obstacles restrict interaction (lack of public paths), their capacity for emergence and adaptation is reduced. We can say that intelligence is an emergent property of an organized system for problem-solving. \\

\textsc{\textbf{Emergent consciousness:}} Consciousness is an awareness of our position in the world\textemdash the capacity not only to solve problems but also to understand why actions are taken and to generalize between contexts. While intelligence optimizes solutions, consciousness searches for meaning and improved integration. Consciousness does not only seek to calculate $\pi$, but also knows the implications (setting goals, knowing when to stop, and knowing why stopping is a good idea) of calculating $\pi$. From a more scientific perspective, consciousness is an \textit{adaptive predictive system} and has currently only arisen from biological organisms. \\

\textsc{\textbf{Pressures:}} Biology uses an adaptive prediction system to reduce uncertainty or entropy to survive in an ever changing environment. Does handling uncertainty and the drive for survival provide enough force for consciousness to emerge? What are the emergent pressures? Rather than acting as a simple calculator, the brain can be viewed as a simulator. It models the external world, the self within that world, and the possible actions of other agents, using episodic memory to replay past events and imagine possible futures. \par

We require \textit{predictive simulation} (imagining outcomes before acting) and \textit{history simulation} (learning by replaying experiences) to allow an organism to anticipate consequences, plan behavior, and adjust to social and physical contexts \cite{bennett2023brief}. In this view, consciousness is not basic computation but a multi-layer modeling process that links memory, prediction, and action to maintain stability and guide decisions under uncertainty. \\

\textsc{\textbf{Interaction:}} In biology, there is a hierarchy of control. General cells are controlled by DNA, which muscle \textit{cells} use for instructions on muscle activity. But how do all of our cells self-organize to build complex structures? A similar question arises with ants in a colony: how do they self-organize \cite{Johnson2001Emergence}? Cells are in constant communication with their neighbors. each cell has a sophisticated tool for detecting the state of its surrounding neighbors. For communication, chemical and electrical messaging are involved. \cite{Johnson2001Emergence}. Ants use a similar method based on pheromones to determine tasks and communication. Emergence occurs at different levels, meaning that intelligence and consciousness both occur at different levels of the computation stack. \\

\textsc{\textbf{Self-assembly:}} Cells are grouped together because they communicate with their neighbors, and these local interactions guide how they behave and organize. This process is known as \textit{gene expression}. In the brain, a similar principle operates in roughly 86 billion neurons. Development through gene expression is decentralized: each cell carries its own genome, which defines its operating characteristics. No cell waits for instructions from a central controller; instead, cells continually oscillate, and incoming signals simply modify the outgoing oscillating signal. This is different from the stimuli and response of the artificial neurons \cite{adamatzky2022unconventional, calcott2011major}. \par

Each cell acts independently and also responds to signals from nearby cells. As replication occurs, mutations also occur, introducing natural variation. These variations occur from slime molds to large animals. When emergence occurs, there is always variation in each emergence \cite{calcott2011major}. \\

\textsc{\textbf{Mycorrhiza communication:}} Mycorrhizal networks, physically known as mycelium, were brought into popular culture through the TV series \textit{Star Trek: Discovery}, which re-imagined them as a mode of futuristic transport. But back on earth, mycorrhizal communication provides another fascinating example of emerging decentralized intelligence \cite{adamatzky2021thoughts}. They create the largest life forms known. These underground fungal networks connect the roots of multiple plants, enabling the exchange of signals and resources throughout the system. The network itself is not intelligence in isolation, but it demonstrates distributed coordination, where communication, adaptation, and survival operate together. It offers another example of how intelligence may not always reside in a single organism but can also emerge as a property of interconnected biological networks. \\

\textsc{\textbf{Artificial emergence:}} \textit{Can we achieve the morphogenesis of Artificial Intelligence?} Creating enough components with simple rules to create a complex system capable of the development of emergent intelligence \cite{adamatzky2022unconventional}. Emergence potentially has to go much further with the initial programming of the system. There are two important questions: Question 1: \textit{Have we created hardware systems complex enough for emergence to occur?} and Question 2: \textit{if emergence occurs, can we identify it?}. There are many debates on whether our current artificial systems are sentient and have consciousness. In other words, are they emergent, or are they just better at mimicking intelligence. \\

\begin{itemize}[leftmargin=*]

\item \textsc{Question 1:} From a computational perspective, modern systems likely have sufficient scale. However, they probably lack the richness of interactions and connections required for true emergence. In current architectures, connectivity is limited, fixed, and artificial neurons remain relatively primitive compared to biological neurons. Our hardware lacks the plasticity and specialization of neurons in biology. If biology is the model we aim to emulate, today's hardware can reproduce some of these features, but only to a limited extent. Whether this is sufficient to produce emergence remains an open question. \\

\item \textsc{Question 2:} Artificial emergence may arise in forms different from our own due to the way we construct and organize complexity. Human behavior, on the contrary, is shaped by many protective constraints embedded in society and culture that guide and regulate how we act. \\

\end{itemize}

\textsc{\textbf{Important caveat:}} We separate intelligence and consciousness to help explain emergence, but in reality, they may not be fully separable. Consciousness may simply be a side effect of intelligence. It could represent a higher-order management system that helps an intelligent system gain a deeper understanding without requiring enormous increases in computational resources. We can probably suggest that consciousness is used to cut corners and make assumptions to speed up intelligence (improve efficiency).

\subsection*{\textbf{11-Thoughts}}
\begin{itemize}[leftmargin=*]
    \item Consciousness and intelligence do not appear suddenly; they emerge when the right environmental conditions and system features are present. Nonlinearity links complexity with emergence and potentially links emergence with intelligence and consciousness under extremely rare circumstances. 
    
    \item Consciousness may arise as a response to uncertainty, enabling us to reason and act despite incomplete information. It depends on a self-organizing brain that continually resists entropy to maintain some form of structure and function \cite{pollan2026world}.
    
    \item Emergence requires self-organization; the timing and synchronization of oscillations have co-evolved as key organizing principles of neural activity.
    
    \item Brain neurons form \textit{small-world connectivity}. Feedback loops reinforce patterns and help stabilize complex behavior.
    
    \item Systems such as \textit{OpenClaw}, which build AI agents tailored to local needs, may also create conditions where emergent behaviors appear.
    
    \item Similar to consciousness, the mechanisms underlying emergence remain largely speculative and not yet fully understood. Each year, we see more theories.
    
    \item An \textit{autodidact} is a self-taught person who learns a subject or skill without a teacher or formal education. Does emergence give autodidact like qualities?
    
    \item \textit{Slime mold} challenges a neuron-centered view because it shows adaptive and problem-solving behavior without neurons, suggesting that intelligent behavior can emerge from coordination alone \cite{Johnson2001Emergence}.

     \item A recent research paper reports that \textit{moss cushions} (clumps of moss that grow together) generate complex electrical activity with patterns resembling neural networks, suggesting that even simple biological systems may exhibit emergent behavior \cite{10.1098/rsos.252341,pollan2026world}.
    
    \item Neuroscientist \textit{Romain Brette} points out that intelligence is \say{\textit{the collective behavior of a colony of living entities rather than a distributed computer}} \cite{brette2026brain}.
    
    \item Physicist \textit{Karl Friston} points out that complexity always comes at a cost: long learning curves, limited life expectancy, or requiring significant energy to sustain \cite{pollan2026world}. Friston proposes the \textit{Free Energy Principle} (FEP). It is a mathematical framework suggesting that living systems survive by reducing uncertainty about their environment, using perception and action to increase their likelihood of survival.
    
\end{itemize}

\subsection*{\textbf{Recommended reading}}

\begin{itemize}[leftmargin=*]
    \item \textit{Emergence}, by Steven Johnson
    \item \textit{The Complex World}, by David C. Krakauer 
    \item \textit{What is intelligence?}, by Blaise Aguera Arcas
    \item \textit{Unconventional Computing, Arts, Philosophy}, ed. A. Adamatzky
    \item \textit{Making Sense Of Chaos}, by J. Doyne Farmer
    \item \textit{The Major Transitions in Evolution}, ed. B. Calcott \& K. Sterelny   
    \item \textit{Principles of Neural Science}, ed. Eric Kandel et al.
    \item \textit{Finding the Mother Tree}, by Suzanne Simard
    \item \textit{The Brain, In Theory}, by Romain Brette
    \item \textit{A World Appears}, by Michael Pollan
\end{itemize}

\section{The Wrong Numbers}
\label{sec:wrong numbers}

\vspace{1.0em}

\begin{adjustwidth}{0.3cm}{}
\textbf{>> Using back-of-the-envelope calculations, if Artificial Intelligence were to follow a process similar to natural evolution, how long would it take to evolve an equivalent system? }
\end{adjustwidth}

\hspace{0.5cm}

\begin{quote} 
\centering 
\vspace{1ex}
\textit{\say{Intelligence, when viewed as computational sophistication, encompasses a broader scope than the computational constraints and sequential nature of time experienced by consciousness.}}\\
\textbf{Stephen Wolfram, a scientist, argues that intelligence is just computation, 2021} \cite{fridman2021wolfram234}
\end{quote}

\hspace{0.5cm}

We finally got to the wrong numbers section, and this, in part, is inspired by a paper written by \textit{Roman V Yampolskiy}, titled \textit{Why Do We Do Not Evolve Software? Analysis of Evolutionary Algorithms} \cite{Numbers}. What are the crucial numbers if we want to build a human simulator from the ground up, using fundamental principles, that is, \textit{a tabula rasa} or clean slate design? \par

We start with a premise $p^\prime$ (p-prime). $p^\prime$ states that \say{\textit{soon we will be able to create an intelligent machine with sufficient computational power to simulate all the evolutionary processes required to produce human intelligence}}. $p^\prime$ is highly dependent on significant advances in computing. Today's technological advances must continue at a similar pace for the next several decades. Now, the vital question to ask \textit{is what the perceived computational gaps between today's computer systems and human-level intelligence are}. To help answer this question, we start right from the beginning\footnote{James Webb Space Telescope's recent discoveries may alter these numbers; starting with the age of the cosmos being potentially 26.7 Billion years old.}; see Figure \ref{figure:basics}. \par

\begin{figure}[hbt!]
\begin{center}
\begin{tabular}{ |l|l| } 
\hline
\textbf{Quantity} & \textbf{Measurement} \\
\hline
Age of the Cosmos & 13.8 Billion years \\
\hline
Age of the Earth & 3.5 Billion years\\
\hline
Age of Life & 3 Billion years \\ 
\hline
Age of Humans & 300 Thousand years \\ 
\hline
\end{tabular}
\end{center}
\caption{Basic numbers}
\label{figure:basics}
\end{figure}

Earth and life are one giant supercomputer, where intelligence emerges from biological processes. Figure \ref{figure:basics} shows the basic cosmological and biological numbers. \textit{What are the numbers of interest?} The number of cell generations (one generation every 20 minutes) from the dawn of life 3 billion years ago is estimated to be around $9^{12}$. The total number of neurons worldwide can be estimated to be roughly $10^{25}$ \cite{Numbers}. We can calculate the amount of computation for intelligent life to be from $10^{31}$ to $10^{44}$ flops (Floating Point Operations per Second). The total data on Earth is around $1.32\times10^{37}$ bytes. To help put this in perspective, a gram of DNA contains about 455 exabytes of data ($3.65\times10^{21}$ bytes). Cell transcribe\footnote{RNA transcribes genetic information from DNA to a \textit{ribosome}} is about $10^{15}$ yottaNOPS. That is about $10^{11}$ (or 455 Billion times faster) than that of the fastest current supercomputer, \textit{China's National Supercomputing Center LineShine supercomputer}. Finally, we add an estimated computing time of roughly 3 billion years. All these numbers are vast. \par

Evolution plays with all these features using convergence (combining to know outcomes) and contingency (creating unexpected outcomes) \cite{calcott2011major}.  Earth is the only planet known to sustain intelligent life, making the probability of life a startlingly rare event, that is, 1 in $26.1\times10^{21}$. \par  

The simulator will have to simulate neurons. A simple neuron is $1\times10^3$ Flops, Hodgkin-Huxley (electrophysiological model) is $1,200\times10^3$ Flops, and the multicompartmental model is $1,200 \times10^6-10^7$ Flops. All on the scale of $10^{25}$ neurons.\par

Currently, the fastest supercomputers range from 0.477 to 2.198 exaFLOPS \cite{LineShineTop500}. The mobile phone network or cryptocurrency mining community may exceed this in raw floating-point performance. The current storage capacity required to run the simulation is $10^{22}$ times that of the top supercomputers in 2026 \cite{LineShineTop500}. \par

If \textit{Moore's law} continues, it will take roughly 6.7 years to increase computation by a power of 1. After 100 years, the gap will still be significant. Even if we created dedicated hardware accelerators and optimized software, it would only add a few more orders of magnitude. If we want artificial computation to evolve into something equivalent to biological intelligence, we will need a new scaling law \cite{sloss_2026_21330151}. \par

Based on the evidence presented, traditional technology will not reach human-level intelligence in the next 50 or 100 years. Other possibilities, such as \textit{Quantum Evolutionary Computation}, may create an equivalent of human intelligence using brute-force computation. Unfortunately, Quantum Evolutionary Computation is too new to predict its likelihood of success. \par

\subsection*{\textbf{3-Thoughts}}

\begin{itemize}[leftmargin=*]
\item The Wrong Numbers discusses using vast amounts of computation to replay billions of years of evolution to create intelligence.
\item It shows that we need a couple of orders of magnitude more computation to even vaguely attempt to create evolution; we may have to employ biology and quantum computing to help emulate this process.
\item Continuous learning is the ability to continuously improve without completely forgetting failures. With evolution, it is a continuous learning process. 
\end{itemize}

\subsection*{Recommended reading}

\begin{itemize}[leftmargin=*]
    \item \textit{The  Major  Transitions  in  Evolution  Revisited}, by B. Calcott et al.   
    \item \textit{Principles of Neural Science}, ed. Eric Kandel et al.
\end{itemize}

\section{Final thoughts}

\vspace{1.0em}

\begin{adjustwidth}{0.3cm}{}
\textbf{>> Final thoughts on our exploration of intelligence: we as a community may want a simple reductionist definition of intelligence that everyone can agree on, but such a definition may not exist without serious compromise. Both our natural and Artificial Intelligence systems are constantly changing and adapting.}
\end{adjustwidth}

\hspace{0.5cm}


\begin{quote} 
\centering 
\vspace{1ex}
\textit{\say{How do we teach AI morality when we can't even teach our children morality?}}\\
\textbf{William Shatner, interview, 2024} \cite{Shatner2024}
\end{quote}
\hspace{0.5cm}

It would be easy to conclude that intelligence is $X$ or $Y$. We hope that we have shown that intelligence is a highly complicated subject with no absolutes and many opinions. Intelligence is interwoven through human culture, language, and being. Many creative theories exist, but fundamental, agreed-upon definitions and metrics are still missing. An Artificial Intelligence system may appear intelligent in narrow problem spaces (for example, in mathematics, the block world, and image recognition) or may seem intelligent through broad general knowledge. But is Artificial Intelligence merely repeating patterns found in a model? Is intelligence, as biologist \textit{David Krakauer} put it, only a \textit{capability}, or is it the \textit{capacity to acquire a capability} \cite{InterviewDavidKrakauer}? There are many questions to be answered. \par

As we give more of our thinking to artificial systems, we slowly erode the knowledge needed to navigate the problems we have given away. Hard problems become harder not because they grow more complex, but because the pattern recognition required to understand them is lost due to our decision to relinquish control. When a solution is generated mainly artificially, we lose control over the problem itself\textemdash the ability to find, diagnose, and fix what has broken. Increasingly, we delegate partial or complete decision-making authority to the artificial systems we have built. \par

\begin{figure}[!ht]
\begin{center}
\includegraphics[width=0.9\linewidth]{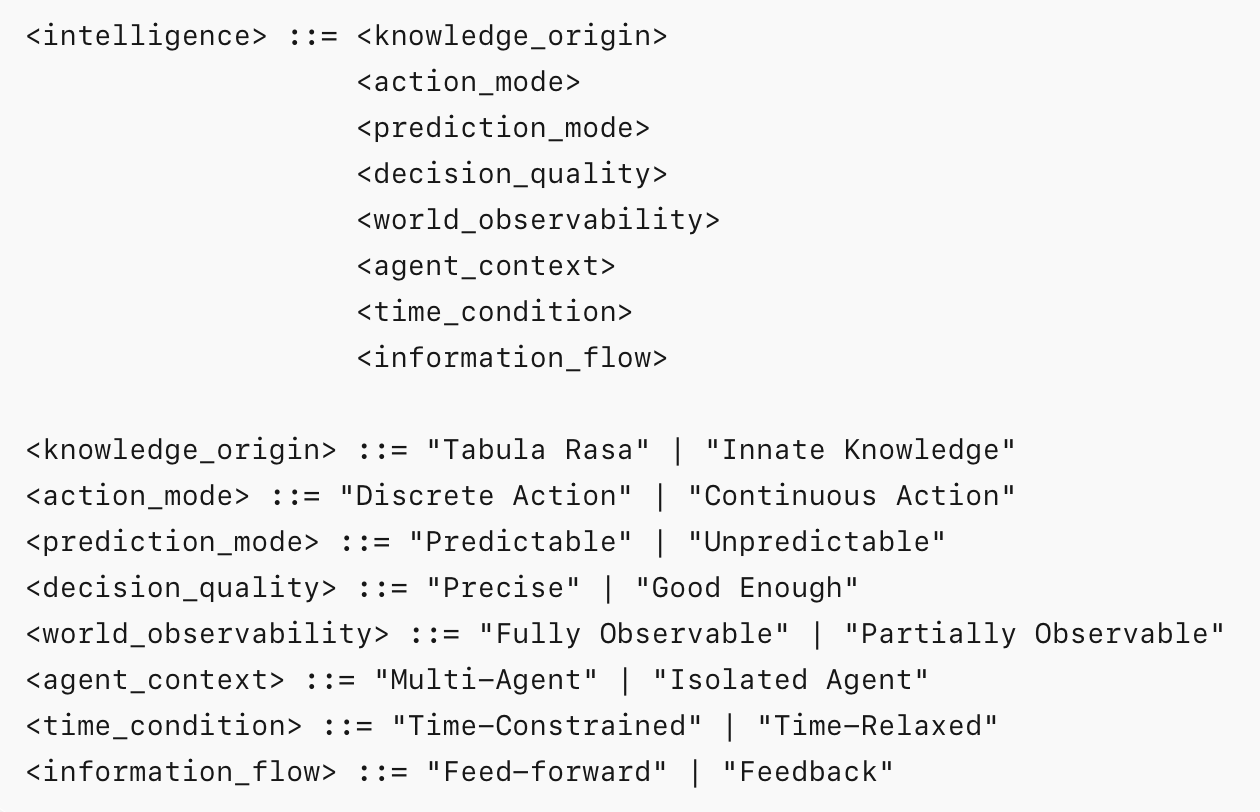}
\caption{Landscape, inspired by Stuart Russell's list \cite{russell2019human}}
\label{figure:options}
\end{center}
\end{figure}

Induction relies on the uniformity of nature, the law of nature. Even when considering our experiences, the past is rarely a good predictor of the future. Nature is just too complicated. The big question is whether an artificial model could fit into a universal law of regular causality, but more needs to be proven. \par

Intelligence comes in many forms. Even when viewed only through time, some forms of intelligence operate in real time, while others can only be recognized over much longer periods. For example, a plant may exhibit intelligent behavior, but we barely notice any changes in our perception of time. This difference is called \textit{chronoception} \cite{pollan2026world}. The shape of a system, the resources it has, and the kinds of problems it must solve all influence how intelligence appears to us. Not all intelligence revolves around human values or human understanding. An octopus, for example, faces a different set of challenges compared with a human or even an elephant. As our knowledge grows, we are beginning to see that intelligence is both far more varied and more general than we once believed. Making us just one of many \cite{pollan2026world}. \par

Figure \ref{figure:options} shows the various options for a landscape. It is based in part on a list provided by Stuart Russell \cite{russell2019human}. Each problem maps onto the categories shown in the figure. These categories can be more or less challenging depending on the system; that is, artificial or biological.\par 

We want intelligence to possess knowledge and to be multimodal. This applies in both a general sense and a more specialized one. It should be inherently logical, with detailed reasoning, repeatability, and verification. It should continuously gain experience, learning, and improving with minimal data requirements. Communication should occur through natural mechanisms, such as voice, text, images, brain waves, and movement. We want intelligence to be conscious and empathetic, understanding the implications. We need it to be creative, solving problems using new ideas and solutions. With these new ideas, we want to predict what happens next and the causal path of how we got here. There is no point in having energy-inefficient intelligence; high Intelligence per Watt (IW) is a requirement. Finally, we want explanations and for it to be actively involved in the decision-making process. \par

Intelligence must adapt to changing environments; it must require only limited examples before full recognition can occur. The system should not require 10,000 hours of learning to become a master of pattern recognition. \par

If we mimic intelligence, is there a threshold we reach that makes it impossible to distinguish it from natural intelligence? Is the only solution to copy existing high intelligence, that is, \textit{innate knowledge}? Or can we build intelligence from the ground up based on basic principles, i.e., \textit{tabula rasa}? The more data, the more confusion. The root of confusion is self-references, and self-references lead to paradoxes. A paradox has no solution or conclusion. \par

Is intelligence even in the realm of our understanding and capability? Intelligence has evolved over 3 billion years. Or is it a set of quantum equations yet to be discovered?\par

Biological neurons operate mainly as feedback systems. Feedback is different from most artificial systems that rely on forward communication. For neuroscientists, neuromorphic technology allows for building systems that are closer to biological ones, i.e., using feedback connections. This potentially opens up the opportunity to explore more complicated subjects such as consciousness.\par 

There are many topics that are not covered in any great depth. These include our favorites:  \textit{Evolutionary Algorithms} and Emergent Behaviors in analog robotics. However, we hope that we have shared enough to make two things clear:

\begin{enumerate}[leftmargin=*]
    \item There is a lot of research and philosophy on biological intelligence, and there is now equally as much focus on artificial varieties. 
    \item There is much that we do not know. Focusing on one specialized subset of Artificial Intelligence should not mean that we limit ourselves to any of the other possibilities. 
\end{enumerate}

There is excitement in Deep Learning, Large Language Models, and other biomimetic systems, and there is still much more to learn. However, it is possible that Artificial Intelligence can only become truly successful if it is understandable. If we only focus on answers, then we are missing the point of intelligence. Focusing on only one subfield may not be the "correct" approach. Another problem is why we are trying to replicate something with many known flaws, such as Human Intelligence. It is like recording a radio broadcast \cite{cnnFearsOverblown}.\par

We have many species on this planet that exhibit intelligent behavior, and for the most part, we ignore them simply because we do not have established ways of communication \cite{sciencealertCouldSoon}.

\hspace{0.5cm}

\begin{quote}
\say{\textit{On the planet Earth, man had always assumed that he was more intelligent than dolphins because he had achieved so much – the wheel, New York, wars, and so on – whilst all the dolphins had ever done was muck about in the water having a good time. However, in contrast, dolphins had always believed that they were far more intelligent than man – for precisely the same reasons.}}\\    
\textbf{Douglas Adams \cite{adams1995hitchhikers}}
\end{quote}

\hspace{0.5cm}

We cannot communicate with other biologically intelligent forms, so how can we plan for a fundamentally different form of intelligence? More than likely, Artificial Intelligence will arrive slowly, be massively distributed, and be unrecognizable. Before that happens, we have a lot of time to practice communicating with other humans and intelligent biological forms. \par

After writing this original piece, some progress has been made in communicating with other animals. For example, scientists believe they have successfully communicated with \textit{humpback whales} \cite{earthHelpsHumans}. The scientists had a 20-minute conversation with a whale named \textit{Twain}. If verifiable, this is a big step forward in having more interesting conversations and understanding. We have also made progress with dolphins, where we can now identify their individual names \cite{sciencealertDolphinsGive}. As we write this in 2026, there appears to be some significant progress. \par

In line with non-biomimicry, non-von Neumann machines, one topic worth diving into is the role of quantum physics in computers and consciousness. However, with quantum physics and minor variations in the observable aspects of our thought processes, some deviation from pure causality could be termed \textit{free will}. Quantum computing is still relatively new, so even if a company has built a verified quantum computer, most companies do not know what to do with it. Quantum computing is an aspect that we believe may genuinely unlock the potential of Artificial Intelligence. Quantum computers are an example of creating intelligence that is radically different from our own (or so we think; we may discover in the future that we are, in fact, quantum computers of sorts). \par

Social and emotional intelligence is another aspect of intelligence that requires further research. We only briefly mentioned it. Some work is happening in Human-Machine Interaction groups, but most is conducted through user interface design and behavioral style observation. The theory of mind, understanding nonverbal intent, and general conversational skills remain largely unexplored. That will likely be the next big step forward in Artificial Intelligence. \par

Lastly, is it possible to have intelligence without consciousness? Is this even a safe path to take with the necessary safeguards? Ultimately, these systems will probably have to make conscious decisions involving conscience. Does this lead to problems? Do we give Artificial Intelligence legal rights? Is it treated as a human or better? Is it responsible for its actions or not? Do we destroy models for their bad behavior? And can we build these new intelligent systems efficiently? There is no point in having inefficient intelligence. Intelligence is the one topic for which we are all experts, and how we use it is predominantly up to us. \textit{What is intelligence?} \par

%
%

\section*{The Authors}

\hspace{0.5cm}

\textbf{\textsc{Andrew N. Sloss}} \textit{ is a computer scientist, a fellow of the British Computer Society, an Affiliate Assistant Professor at the University of Washington, and a long-time researcher with nearly 40 years of experience in the Computing and Semiconductor Industry. He has taught and continues to teach engineering. He is constantly analyzing complex systems. Andrew has observed and analyzed strategies throughout his career, realizing that grit is a typical pattern for success. His professional journey has taken him through various companies and positions, offering valuable insight from the grassroots to the world stage. Identifying how long-term strategic thinking achieves the best results}\\

\textbf{\textsc{Karl F. Fezer}} \textit{is an Artificial Intelligence professional with a Master’s degree from the University of Georgia, whose career spans leading organizations including Samsung, Intel, Arm, and currently Lockheed Martin. He specializes in MLOps and Responsible AI, with a focus on aligning technological advancement with the Ethics of AI principles of the Department of Defense. Throughout his career, he has contributed to bridging the gap between complex technical systems and practical implementation, particularly through his leadership in Developer Relations. His work emphasizes the development of accessible and scalable AI solutions and the cultivation of engaged technical communities. Currently living in Seattle, he balances his professional endeavors with family life and an appreciation for the outdoors.}

\begin{acks}
Each paper is only as good as the reviewers. We personally thank the following people: Rex St. John, Ada Coghen, Jeremy Johnson, Wendy Elsasser, Paul Gleichauf, Joseph Williams, Colin David Sloss, Simon John Poile, Ian H. Field, Daniel Robert Field, Maria Khmel, Adam Glass, Robert Sloss, and Steven Gustafson for their participation and thoughts. Also, a massive thank you to all the great authors and thought leaders we referenced throughout this primer.   
\end{acks}

\section{Disclaimer}

Written by the authors with the assistance of Artificial Intelligence tools; all ideas, arguments, and final editorial decisions remain the responsibility of the authors. The writing has been reviewed and refined with the help of tools such as \textit{Grammarly}, \textit{Claude}, \textit{Gemini}, and \textit{OpenAI ChatGPT}. These tools were used only to improve grammar, clarity, and style. They were not used to generate the underlying content, facts, arguments, or opinions. The authors remain fully responsible for all content, ideas, interpretations, bad jokes, and conclusions presented here. Important details should always be verified independently. \\

\hspace{0.5cm}
\hrule
\hspace{0.5cm}

\section*{OPEN DASKALOS PROJECT series:}

We cannot have \textbf{Intelligence} without \textbf{Knowledge}. We need \textbf{Intelligence} and \textbf{Knowledge} to determine the right \textbf{Algorithms}. \textbf{Algorithms} are used to imitate \textbf{Intelligence} and provide \textbf{Knowledge}. All three are deeply connected, biologically, artificially, and physically.\par

\hspace{0.5cm}

\noindent
\textbf{Lost in Algorithms: Architecting Complexity}, November 2025\\
by Andrew Sloss\\\\
Algorithms are becoming more capable, and with that we enter a realm of \textit{hic sunt dracones}\textemdash here be dragons\textemdash the regions that lie beyond our established maps. We are entering an exciting, potentially dangerous, and uncharted period of time with algorithms. Our curiosity about the world drives us to look for new methods. It is precisely these emerging methods that make this rapidly evolving field worth exploring. We move through eras of rapid \textit{exploration} followed by eras of \textit{consolidation}. This paper is therefore designed to give \textit{system architects} an overview of the behavioral and societal implications, offering not only technical structures, but also human interaction and understanding. \\

\noindent
\parbox{\linewidth}{\url{https://www.researchgate.net/publication/367163246_Lost_in_Algorithms_Architecting_Complexity}}\\

\noindent
\textbf{Exploring Knowledge: The Next Horizon }, February 2026\\
by Andrew Sloss\\\\
\textit{Exploring Knowledge} is a journey into a subject that covers everything, affects everything, and is the goal for everything. Knowledge plays a fundamental role in our lives, defines who we are as a species, and separates us as individuals. We use it to express and understand nature\textemdash it is a component of intelligence. In this paper, we try to explore \textit{what knowledge is} and, in doing so, highlight the possible elements that make up knowledge and what knowledge provides\textemdash merging theoretical and inspirational estimates. \\

\noindent
\parbox{\linewidth}{\url{https://www.researchgate.net/publication/388155680_Exploring_Knowledge_The_Next_Horizon}}

\flushend 
\balance

\bibliographystyle{ACM-Reference-Format}
\bibliography{intell_prime}

\appendix

\end{document}